\documentclass[lettersize,journal]{IEEEtran}
\usepackage{amsmath,amsfonts}
\usepackage{algorithmic}
\usepackage{algorithm}
\usepackage{array}
\usepackage[caption=false,font=normalsize,labelfont=sf,textfont=sf]{subfig}
\usepackage{textcomp}
\usepackage{stfloats}
\usepackage{url}
\usepackage{verbatim}
\usepackage{graphicx}
\usepackage{cite}
\usepackage{color}
\usepackage{multirow, multicol}
\begin{document}

\title{Threshold-adaptive Unsupervised Focal Loss for \\Domain Adaptation of Semantic Segmentation}

\author{Weihao Yan,
Yeqiang Qian,~\IEEEmembership{Member,~IEEE,} 
Chunxiang Wang,~\IEEEmembership{Member,~IEEE,}
and Ming Yang,~\IEEEmembership{Member,~IEEE}
\thanks{This work is supported by the National Natural Science Foundation of China (62103261/62173228). \emph{(Corresponding author: Ming Yang.)}}
\thanks{Weihao Yan, Chunxiang Wang and Ming Yang are with the Department of Automation, Shanghai Jiao Tong University, Key Laboratory of System Control and Information Processing, Ministry of Education of China, Shanghai, 200240, China (email: mingyang@sjtu.edu.cn)}
\thanks{Yeqiang Qian is with University of Michigan-Shanghai Jiao Tong University Joint Institute, Shanghai Jiao Tong University, Shanghai, 200240, China. (email: qianyeqiang@sjtu.edu.cn)}}


\markboth{Submitted to IEEE Transactions on Intelligent Transportation Systems on April 2, 2022}%
{Yan \MakeLowercase{\textit{et al.}}: Threshold-adaptive Unsupervised Focal Loss for Domain Adaptation of Semantic Segmentation}


\maketitle

\begin{abstract}
	Semantic segmentation is an important task for intelligent vehicles to understand the environment.
	Current deep learning methods require large amounts of labeled data for training. 
	Manual annotation is expensive, while simulators can provide accurate annotations.
	However, the performance of the semantic segmentation model trained with the data of the simulator will significantly decrease when applied in the actual scene.
	Unsupervised domain adaptation (UDA) for semantic segmentation has recently gained increasing research attention, 
	aiming to reduce the domain gap and improve the performance on the target domain.
	In this paper, we propose a novel two-stage entropy-based UDA method for semantic segmentation.
	In stage one, we design a threshold-adaptative unsupervised focal loss to regularize the prediction in the target domain, 
	which has a mild gradient neutralization mechanism and mitigates the problem that hard samples are barely optimized in entropy-based methods.
	In stage two, we introduce a data augmentation method named cross-domain image mixing (CIM) to bridge the semantic knowledge from two domains. 
	Our method achieves state-of-the-art 58.4\% and 59.6\% mIoUs on SYNTHIA-to-Cityscapes and GTA5-to-Cityscapes using DeepLabV2
	and competitive performance using the lightweight BiSeNet. 
\end{abstract}


\section{Introduction}
\IEEEPARstart{S}{emantic} 
segmentation is an important perceptual task for the intelligent transportation system, 
which can provide pixel-level semantic information, e.g., road, traffic sign, and pedestrian. 
Thanks to the development of deep learning and large-scale public datasets, 
semantic segmentation has made remarkable progress\cite{restricted,gated,pass,erfnet} in recent years. 
Most of these supervised deep learning methods have an indispensable requirement for high-quality labeled data, 
which are expensive to obtain. 
It takes about 90 minutes to label an image of Cityscapes\cite{cityscapes}.
So an alternative way is to utilize synthetic datasets like SYNTHIA\cite{synthia} and GTA5\cite{playingfordata} for model training. 
However, the models trained on synthetic datasets (source domain) usually perform poorly in real scenarios (target domain) due to the large domain gap. 
To alleviate this severe problem, researchers have resorted to investigating unsupervised domain adaptation (UDA) methods for semantic segmentation.

\begin{figure}[!t]
	\captionsetup[subfloat]{font=scriptsize,labelfont=scriptsize}
	\centering
	\subfloat[Gradient analysis]{\includegraphics[width=0.8\linewidth]{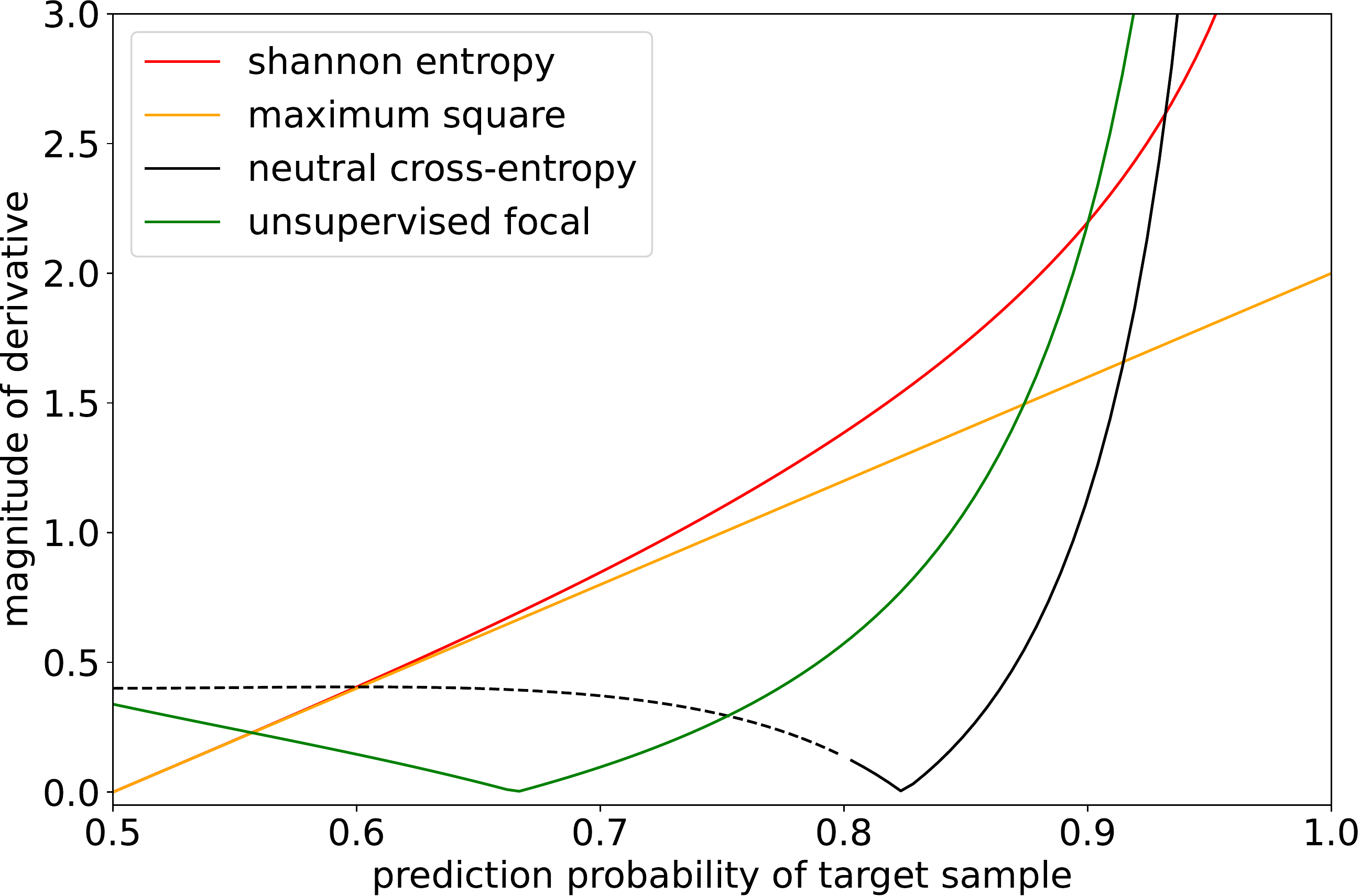}}
	\caption{Gradient analysis of entropy-based unsupervised loss. 
	Shannon entropy sharpens the prediction distribution, 
	and its gradient is strongly biased toward easy samples (prediction probabilities near 1). 
	Maximum square loss reduces the gradient magnitude of easy samples,  
	but the gradient tends to be 0 when probability approaches 0.5.
	Neutral cross-entropy smoothes the over-sharpness of entropy loss. 
	However, it utilizes a high confidence threshold (0.8) to filter the prediction results.
	Unsupervised focal loss has a mild gradient neutralization mechanism and dynamic threshold adjustment strategy, 
	making hard samples with low prediction probabilities optimized.}
	\label{fig:gradient_vis}
\end{figure}

The literature on UDA for semantic segmentation is recently dominated by adversarial-based and self-training methods. 
Adversarial-based methods\cite{advent,cyclegan,leanringoutput,bidirectional} have shown competitive adaptation performance 
by learning domain invariant representations at the input, feature, or output levels.
But the training procedure is complex, and the adversarial loss\cite{gan} frequently converges to local optimal. 
Self-training methods\cite{classbalance,rectifying,iast,prototypical} have state-of-the-art adaptation performance recently.
Pseudo labels for the target domain are generated and used to re-train the segmentation model. 
However, they often require computationally expensive iterative training and act more like post-processing of pre-adapted models. 

Recently, a new line of entropy-based UDA methods\cite{advent, maximum, neutral, confidence} for semantic segmentation has shown progressive development, 
which regards the UDA for semantic segmentation as a regularization process on the target domain.
The work in \cite{advent} proposes to minimize the entropy maps of prediction results on the target domain, 
making the network produce high-confidence predictions. 
Then \cite{maximum} finds the gradient of entropy is biased towards samples that are easy to transfer and proposes the maximum square loss to balance the gradient of the well-classified target samples. 
Besides that, the over-sharpness of entropy minimization is mitigated in \cite{neutral} by introducing the pixel-level consistency regularization and proposing the neutral cross-entropy loss. 
After that, \cite{confidence} designs confidence-aware entropy to help the model focus more on high-confident data points.
Entropy-based methods have good theoretical support and generally train faster \cite{advent,maximum,neutral,confidence}.
Though efficient and effective, 
the performance of entropy-based methods still falls behind the other two types of state-of-the-art methods and needs further improvement.

There are two main problems that hinder the further improvement of these entropy-based methods. 
One is that the hard samples are barely optimized.
In the training process, the model parameters are updated through backpropagation. 
Once that learning rate is selected, the degree of parameters updating is proportional to the magnitude of the gradient of the loss.
The gradient analysis of these entropy-based losses\cite{advent,maximum, neutral} is shown in Fig.~\ref{fig:gradient_vis},
where the binary classification case is used for a clearer presentation.
The gradient of the shannon entropy loss is strongly biased towards samples with probabilities near 1.
Still, it decreases to 0 when probability approaches 0.5, 
making the hard-to-transfer samples barely optimized.
A similar situation happens in the maximum square loss, though it reduces the gradient of these well-classified target samples.
For neutral cross-entropy loss, 
a confidence threshold like 0.8 is selected to filter the prediction results with low confidence for training stability.
It may make the hard samples that are usually with low prediction probabilities not included in calculating the loss, let alone optimized.
Another is that these entropy-based UDA methods only impose a regularization item on the target domain 
while ignoring the explicit connection of semantic knowledge between the source and the target domain.
One way is to utilize adversarial training to learn domain invariant features, 
but these adversarial-based methods are well-known for difficult training. 

In this paper, we propose a novel two-stage entropy-based UDA method to mitigate the above two problems.
In stage one, 
the threshold-adaptive unsupervised focal loss is designed and applied to the target domain. 
It has a gentle gradient neutralization mechanism to smooth the over-sharpness of shannon entropy loss 
and a class-level dynamic threshold adjustment strategy, 
which helps optimize hard samples.
In stage two, we introduce a data enhancement method named cross-domain image mixing (CIM) to bridge the semantic knowledge from two domains.
Three image-label pairs from the source domain, target domain, and CIM are fed into the network for training. 
Experiments are conducted on two synthetic-to-real settings 
(SYNTHIA-to-Cityscapes and GTA5-to-Cityscapes) to verify the effectiveness of our method.
It achieves state-of-the-art 58.4\% and 59.6\% mIoUs using DeepLabV2 and
competitive 52.2\% and 55.4\% mIoUs using lightweight BiSeNet.
Our contributions are summarized as follows:
\begin{itemize}
	\item We design a threshold-adaptive unsupervised focal loss for the UDA problem of semantic segmentation.
	It adjusts the gradient contributions of easy and hard samples 
	and has a class-level dynamic threshold adjustment
	strategy, helping the hard target samples get optimized.
	\item We introduce a data augmentation method named cross-domain image mixing (CIM) into entropy-based methods, 
	which bridges the semantic knowledge of two domains. 
	\item We propose a novel two-stage UDA framework for semantic segmentation, 
	achieving state-of-the-art performance on major cross-domain
	benchmark datasets like SYNTHIA-to-Cityscapes and GTA5-to-Cityscapes. 
\end{itemize}

\section{Related Work}
\subsection{Semantic Segmentation}
Semantic segmentation has made significant advances in recent years since the development of deep learning and the availability of public datasets.
FCN\cite{FCN} is the first segmentation model that achieves pixel-level predictions.
After that, many works on semantic segmentation have emerged.
One stream of methods aims to improve the accuracy of semantic segmentation models\cite{gated,pass,restricted,deeplabv2,hrnet,segformer}.
DeepLabV2\cite{deeplabv2} proposes atrous spatial pyramid pooling (ASPP) with filters at multiple sampling rates, 
improving the segmentation performance on multi-scale objects.
The other trend intends to develop real-time models for applications\cite{bisenet,erfnet,realtimeits,stdcnet}. 
BiSeNet\cite{bisenet} consists of Spatial Path, Context Path, and Feature Fusion Module, 
which makes a balance between speed and segmentation performance.
Both of them have indispensable demand for high-quality labeled datasets, which are expensive to acquire.
One possible method to reduce data labeling costs is adopting synthetic datasets\cite{playingfordata,synthia}.
However, models trained on synthetic datasets often perform poorly when applied in real scenarios due to the large domain gap.
UDA methods are developed for this problem. 
In this paper, lightweight BiSeNet\cite{bisenet} with ResNet18 as the backbone is used for real-time performance and training efficiency. 
Meanwhile, we also adopt the widely used DeeplabV2\cite{deeplabv2} for a fair comparison with state-of-the-art. 

\subsection{UDA for Semantic Segmentation}
UDA methods can alleviate the domain gap between source and target domains, 
improving the performance of the segmentation model on the unlabeled target domain. 
Current advanced UDA approaches for semantic segmentation are dominated by adversarial-based and self-training methods.
On one hand, adversarial-based methods\cite{advent,cyclegan,leanringoutput,bidirectional} usually adopt generative adversarial networks (GAN)\cite{gan} to align the distributions of two domains, 
aiming to learn domain invariant representations at different levels. 
Meanwhile, image style translation (IST) methods\cite{cyclegan,fda,corsetofine} transfer the texture of source domain images to the target domain, reducing the domain gap at the input level.
The work in\cite{corsetofine} proposes a global photometric alignment module that aligns the image in the source domain with the reference image in the target domain.
Although adversarial-based methods usually have competitive performance, they suffer from training instability.
In this paper, a prevalently used IST method CycleGAN\cite{cyclegan} is taken to preprocess the images from the source domain offline.

On the other hand, self-training methods\cite{classbalance,iast,rectifying,prototypical} generate pseudo labels in the target domain 
and employ them for iterative training.
Class balanced self-training\cite{classbalance} is the first to apply pseudo-labeling to UDA.
After that, uncertainty estimation is applied to rectify the pseudo labels in\cite{rectifying}.
The method in \cite{sac} generates co-evolving pseudo labels for their self-supervised framework.
Current state-of-the-art ProDA\cite{prototypical} uses prototypical pseudo label denoising to update the pseudo labels in stage one online 
and adopts knowledge distillation to a self-supervised model in the following two stages.
Nevertheless, producing high-quality pseudo labels remains challenging, 
and iterative training is computationally expensive and time-consuming.
Self-training methods are more like ``post-processing'' of the pre-adapted model, 
which can follow the adversarial-based and entropy-based methods to further improve the adaptation performance. 

Moreover, some UDA methods\cite{bridging,dannet} tend to improve the adaptation performance of the semantic segmentation model from daytime to nighttime.
DANNet\cite{dannet} is the first one-stage adaptation framework for nighttime semantic segmentation via adversarial learning.
In addition, the work in \cite{affinity} investigates the adaptation in affinity space, 
which leverages co-occurring patterns between pairwise pixels.
The guidance from self-supervised depth estimation is leveraged in \cite{corda} to strengthen the target semantic predictions.
Recently, \cite{daformer} explores the possibility of improving the network structure and training strategy for domain adaptative semantic segmentation,
which utilizes the MiT-B5\cite{segformer} encoder and boosts the state-of-the-art.
However, the training of transformer architecture is well-known for being complex and cumbersome.
We still adopt the CNN architecture for experiments, 
and the choice of segmentation model is not the focus of our work.  

Another stream of entropy-based UDA methods follows the cluster assumption\cite{semi}, 
which regards the UDA for semantic segmentation as a regularization process on the target domain\cite{advent,maximum,neutral,confidence}.
Entropy minimization is proposed in \cite{advent}, which minimizes the entropy maps of prediction results on the target domain, 
and encourages the network to produce predictions with high confidence.
Maximum square loss is designed in \cite{maximum} to balance the gradient of well-classified target samples,
which alleviates the problem that the gradient of entropy is biased towards easy-to-transfer samples.
Pixel-level consistency regularization is introduced in \cite{neutral} to form neutral cross-entropy loss, 
which has a gradient neutralization mechanism to smooth the over-sharpness of entropy loss.
The work in\cite{confidence} updates the common entropy to confidence-aware entropy, 
forcing the network to focus on the high-confidence predictions.
Though efficient and effective, 
entropy-based methods only bring little improvement and still lag behind current advanced UDA methods.
As discussed before, they barely optimize hard-to-transfer samples 
and lack the explicit semantic knowledge connection between two domains.
In this paper, we focus on the entropy-based methods, aiming to mitigate the above two problems and boost their adaptation performance to the state-of-the-art. 

\subsection{Image Mixing Strategy}
Image mixing has proven effective in semi-supervised learning (SSL).
The central idea is to mix two images and their labels, forming additional, highly perturbed training samples.
ClassMix\cite{classmix} randomly selects half of the classes in one image and pastes them onto another one to better respect semantic boundaries.
Image Mixing is a preferable approach to bridge the semantic knowledge between two domains, 
and \cite{dacs} has verified its effectiveness in the UDA problem of semantic segmentation.
Different from \cite{dacs}, 
we emphasize the loss contribution of the pixels near the boundaries of mixing mask, 
which have the receptive field of both domains,
and design long-tail class pasting to improve adaptation performance. 

\section{Method}
\begin{figure*}[!t]
	\centering
	\includegraphics[width=0.9\linewidth]{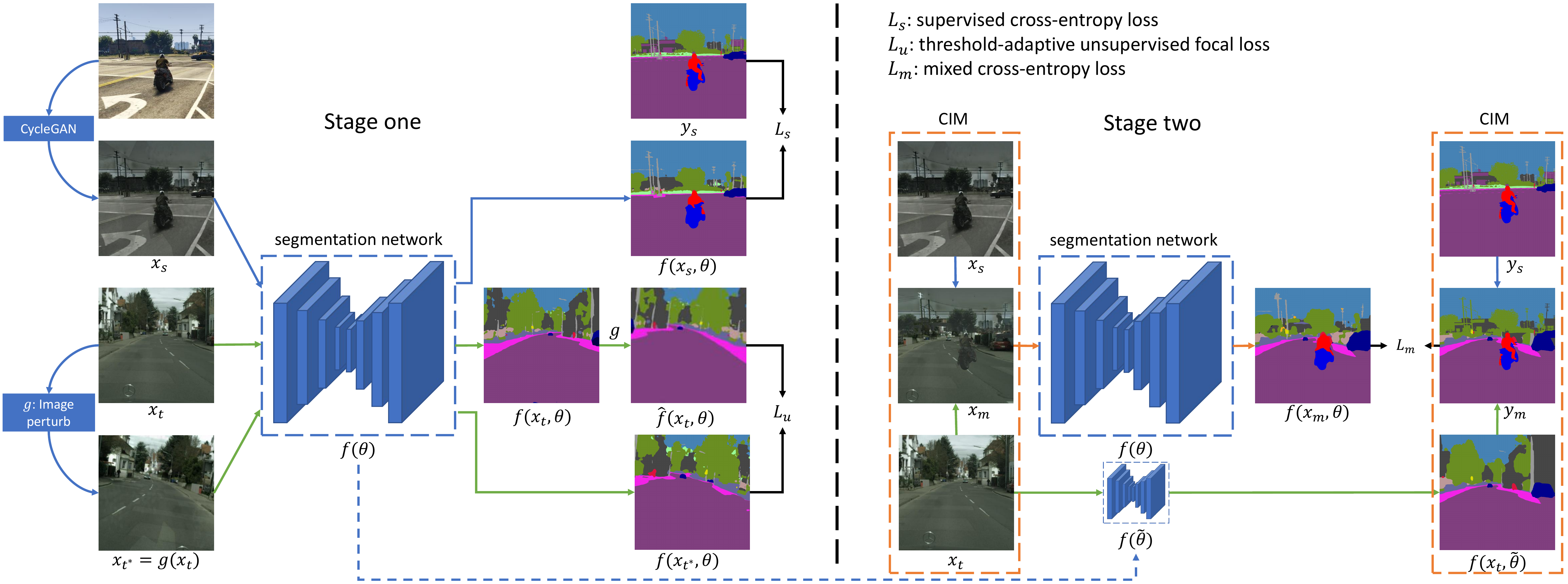}
	\caption{Two-stage UDA framework for semantic segmentation.
	The data flow of the source domain, target domain, and CIM is denoted by blue, green, and orange lines, respectively.
	In stage one, CycleGAN is adopted to preprocess the source domain images. 
	Then the threshold-adaptative unsupervised focal loss is applied in the target domain and the supervised cross-entropy loss in the source domain.
	In stage two, CIM is introduced and helps connect the semantic knowledge between two domains.
	$L_s$, $L_u$, and $L_m$ present supervised cross-entropy loss, threshold-unsupervised focal loss, and mixed cross-entropy loss.
	}
	\label{fig:whole_framework}
\end{figure*}
The framework of our two-stage UDA method is shown in Fig~\ref{fig:whole_framework}. 
In stage one, CycleGAN\cite{cyclegan} is used to transfer the image style of the source domain to the target domain offline when using BiSeNet, 
while DeepLabV2 does not.
The image $x_{t}$ from the target domain goes through perturbation $g$, and the augmented image is denoted as $x_{t^{*}}$.
Then $x_{t}$ and $x_{t^{*}}$ are fed into the segmentation model $f(\theta)$, 
and the predictions $f\left(x_{t},\theta\right)$ and $f\left(x_{t^{*}},\theta\right)$ are obtained.
Apply the same perturbation on $f\left(x_{t}, \theta\right)$ to align with $f\left(x_{t^{*}},\theta\right)$, 
our threshold-adaptive unsupervised focal loss $L_u(\hat{f}\left(x_{t},\theta\right), f\left(x_{t^{*}},\theta\right))$ is calculated.
Meanwhile, the supervised cross-entropy loss $L_s\left(f\left(x_s,\theta\right), y_s\right)$ is computed in the source domain.

In stage two, the pre-adapted model from stage one $f(\tilde{\theta})$ generates pseudo labels for target samples.
Then mixed image-label pairs ($x_m$,$y_m$) are acquired through CIM with boundary enhancement and long-tail class pasting.
Supervised cross-entropy loss $L_m$ is also used in stage two with the $L_s$ and $L_u$. 
The details will be presented in the following.

\subsection{Overview of entropy-based UDA methods}
Use $D_{s}=\{(x_s,y_s)|x_s \in R^{3\times H\times W}, y_s \in R^{H \times W}\}$ 
and $D_{t}=\{x_{t} | x_{t}\in R^{3\times H\times W}\}$
to denote the labeled source domain and unlabeled target domain.
The general loss function of the entropy-based UDA method can be formulated as:

\begin{equation}
	\centering
	Loss = L_{s}\left(x_{s},y_{s}\right) + \lambda_{u} L_{u}\left(x_{t}\right)
	\label{eq:general loss}
\end{equation}
where $L_{s}$ is the supervised cross-entropy loss in the source domain,
and $L_{u}$ is the unsupervised loss applied on the target images with coefficient $\lambda_{u}$.
The training objective is to adjust the parameters of model $\theta$ to minimize the loss function (\ref{eq:general loss}).
Denote $f\left(x_t,\theta\right)$, $\hat{f}\left(x_{t},\theta\right)$, and $f\left(x_{t^{*}},\theta\right)$ as $p_t$, $\hat{p}_t$, and $p_{t^*}$ for convenience. 
The shannon entropy loss ($L_{shan}$) is directly utilized as the $L_{u}$ in \cite{advent}:

\begin{equation}
	\centering
	L_{shan} \left(p_{t}\right) 
	= -\frac{1}{|I_{t}|} \sum_{n\in I_{t}} \sum_{c=1}^{C} p^{n,c}_t \log\left(p^{n,c}_t\right)
	\label{eq:shannon}
\end{equation}	
where $C$ is the number of classes, $c$ represents channel number, $n$ denotes pixel location, and $I_{t}$ is the loss calculation mask.
Suppose $p_t\in R^{C\times H\times W}$, 
and $p_{m}=\max\limits_{dim=C} p_t$ is the prediction confidence map. 
$I_{t}=\{(h,w)|p_{m}^{h,w}>t, t\in [0,1]\}$ means that 
only pixels with prediction confidence greater than the threshold $t$ are included in the loss calculation.

\subsection{Threshold-adaptative unsupervised focal loss}
Inspired by the focal loss\cite{focal}, 
we propose the unsupervised focal loss for domain adaptation of semantic segmentation.
Specifically, it consists of two parts, 
the first part is the shannon entropy for the prediction probability distribution $\hat{p}_t$ of weakly perturbed target image $x_t$,
which makes the model tend to produce high-confidence prediction results:
\begin{equation}
	\centering
	L_{shan} \left(\hat{p}_{t}\right) 
	= -\frac{1}{|I_{t}|} \sum_{n\in I_{t}} \sum_{c=1}^{C} \hat{p}^{n,c}_t \log\left(\hat{p}^{n,c}_t\right)
	\label{eq:shannon_infocal}
\end{equation}

The second part is the KL-divergence $L_{KL^{'}}\left(\hat{p}_t,p_{t^*}\right)$ with the loss adjustment item:
\begin{equation}
	\centering
		\frac{1}{|I_{t}|} \sum_{n\in I_{t}} \sum_{c=1}^{C} {\hat{p}_t}^{n,c} \cdot \Big (\log\left(\hat{p}_t^{n,c}\right) 
		-\left(1-p_{t^*}^{n,c}\right)^{\gamma}\log \left(p_{t^*}^{n,c}\right)\Big ) 
	\label{eq:KL_focal}
\end{equation}
where parameter $\gamma$ controls the degree of regularization, 
the gradient of $\hat{p}_t$ is detached and serves as the soft ``pseudo label.''
The function of KL-divergence is to make the model produce consistent prediction results for perturbed image pairs, 
that is, to align the $p_{t^*}$ with $\hat{p}_t$, making it more robust.
The $\left(1-{p_{t^*}}\right)^{\gamma}$is introduced to balance the loss contribution of easy and hard samples.
As $\hat{p}_t$ is fixed, the adjustment item tends to be 0 when $p_{t^*}$ approaches 1 (easy-to-transfer) and vice versa.
The unsupervised focal loss is obtained through the summation of shannon entropy and adjusted KL-divergence:

\begin{equation}
	\centering
		L_{focal}\left(\hat{p}_t,p_{t^*}\right) =
		L_{shan}\left(\hat{p}_t\right) + L_{KL^{'}}\left(\hat{p}_t,p_{t^*}\right)
\end{equation}

Then we discuss the relationship between our unsupervised focal loss and the supervised focal loss.
The supervised focal loss ($L_{s\_focal}$) is widely used in semantic segmentation for hard samples optimization, which is formulated as:
\begin{equation}
		\centering
			L_{s\_focal}\left(y_t,p_t\right) = - \frac{1}{|I_{t}|} \sum_{n\in I_{t}} \sum_{c=1}^{C}  {y_t}^{n,c}
			\left(1-p_{t}^{n,c}\right)^{\gamma}\log \left(p_{t}^{n,c}\right)
		\label{eq:s_focal}
\end{equation}
We expand the supervised focal loss as:
\begin{equation}
	\centering
	\begin{aligned}
		&L_{s\_focal}\left(y_t,p_t\right) = L_{shan}\left(y_{t}\right) + L_{KL^{'}}\left(y_{t}, p_t\right)\\
		&= -\frac{1}{|I_{t}|} \sum_{n\in I_{t}} \sum_{c=1}^{C} y_{t}^{n,c}\log\left(y_{t}^{n,c}\right) + \\
		&\quad \frac{1}{|I_{t}|} \sum_{n\in I_{t}} \sum_{c=1}^{C} y_{t}^{n,c}(\log\left(y_{t}^{n,c}\right)
		-\left(1-{p_t}^{n,c}\right)^{\gamma}\log \left({p_t}^{n,c}\right)) \\
	\end{aligned}
	\label{eq:supervised focal}
\end{equation}
In the supervised focal loss, the $L_{shan}(y_t)$ is constant 
since the labels are available and fixed. 
It demonstrates that minimizing the supervised focal loss $L_{s\_focal}(y_t, p_t)$ 
is equal to optimizing the adjusted KL-divergence $L_{KL'}(y_t, p_t)$,
which inspires us to use perturbed image pairs to 
establish an unsupervised form of adjusted KL-divergence $L_{KL^{'}}\left(\hat{p}_t,p_{t^*}\right)$.
Compared with the supervised one, 
our unsupervised focal loss replaces the ground truth $y_{t}$ with the estimated prediction $\hat{p}_t$.
Essentially, our unsupervised focal loss is a particular format of supervised focal loss,
with the shannon entropy of soft ``pseudo label'' $L_{shan}(\hat{p}_t)$ as the learnable variable.
It can adjust the contribution of easy and hard target samples to the unsupervised loss $L_u$
so that the hard samples are optimized.

\begin{figure}[!t]
	\centering
	\includegraphics[width=0.995\linewidth]{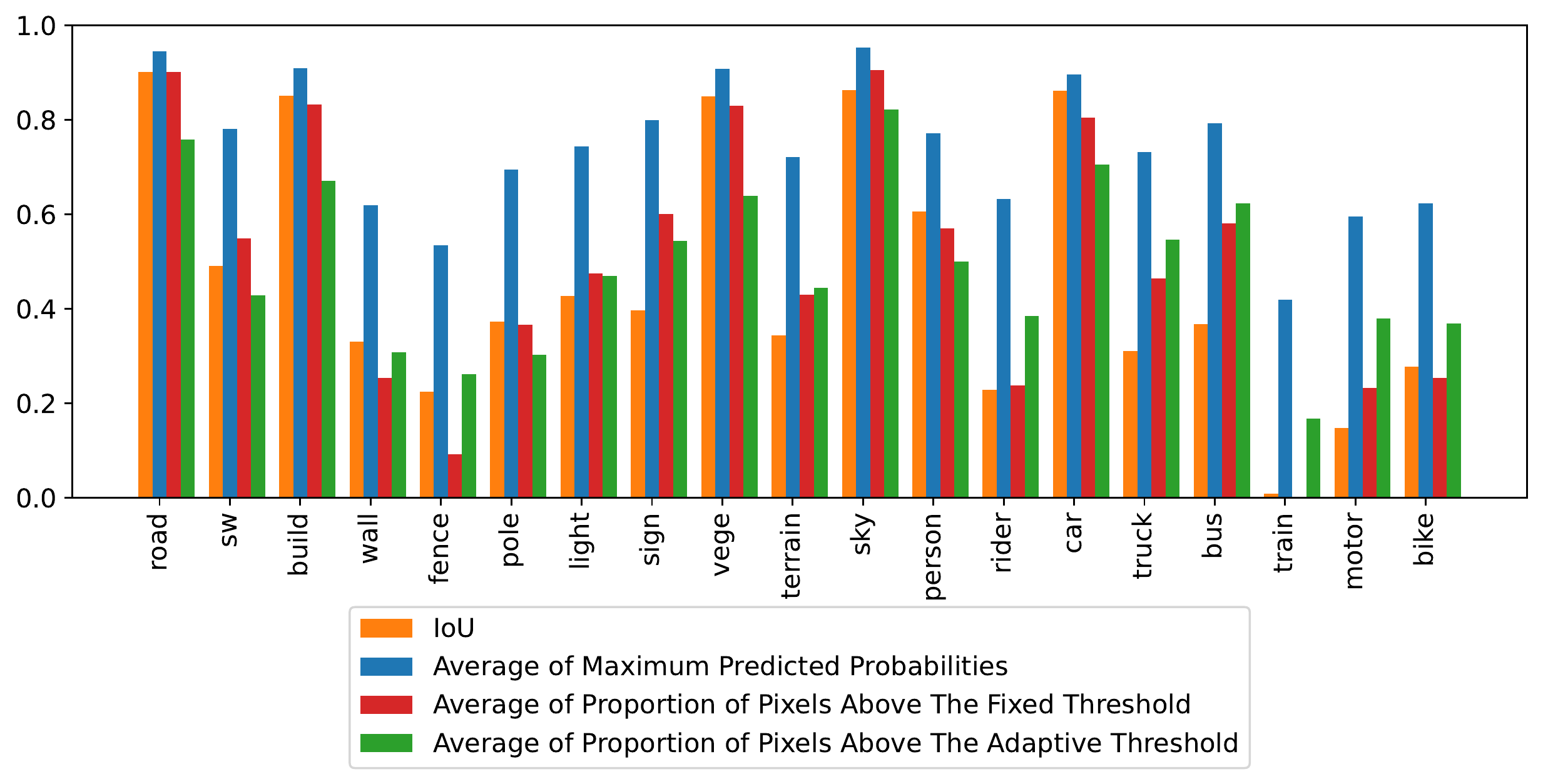}
	\caption{Numerical characteristic analysis of each class on GTA5-to-Cityscapes. 
	When using a fixed high confidence threshold (0.8),
	some classes like fence and train are barely included in the loss calculation.
	Our adaptative threshold strategy helps these classes contribute more to the loss.
	}
	\label{fig:threshold}
\end{figure}

The computational mask $I_{t}$ is another factor that hinders the optimization of hard samples, 
which adopts a confidence threshold like 0.8 to remove pixels with low maximum prediction probabilities to stabilize the unsupervised training process\cite{neutral}.
However, those hard samples are often with low maximum prediction probabilities 
since the model cannot discriminate their classes well.
We visualize the average maximum prediction probability and proportion of pixels above the threshold (0.8) of each class in the GTA5-to-Cityscapes experiment, shown in Fig.~\ref{fig:threshold}.
Classes with higher IoU generally have higher maximum prediction probabilities and vice versa.
Those classes with low maximum prediction probabilities are rarely 
incorporated into the computation of the unsupervised loss, leading to low IoU.

We introduce a class-level dynamic threshold adjustment strategy for $I_t$, 
where the threshold is updated online during the training process.
Denote the threshold for each class as $\alpha\in R^{C}$. 
We use the Exponential Moving Average strategy to update it for the current sample from the target domain:
\begin{equation}
	\centering
	\begin{aligned}
		\alpha_{0} &= [t, ..., t]\in R^{C}\\
		\alpha_{k} &= a \alpha_{k-1} + (1-a) \alpha_{k'}, k\geq 1 \\
	\end{aligned}
	\label{eq:threshold_update_a}
\end{equation}
where $\alpha_{0}$ represents the initial threshold t (set as 0.8), 
$\alpha_{k}$ and $\alpha_{k-1}$ denote the class threshold in the $k$th and $(k-1)$th iteration, 
$a$ is the historical memory parameter, 
and $\alpha_{k'}$ is the class threshold calculated from the current sample:
\begin{equation}
		\centering
		\begin{aligned}
			p_{m}^{c}&=p_{m}[argmax(p_t)=c] \\
			\alpha_{k'}^{c} &= descend\left(p_{m}^{c}\right)[b (e^{\alpha_{k-1}^{c}-1} )^{d}|p_{m}^{c}|] \\
		\end{aligned}
		\label{eq:current_threshold}
\end{equation}
where $p_{m}^{c}$ is the prediction confidence list for class $c$, 
obtained by choosing the pixels with pseudo label $c$.
$\alpha_{k'}^{c}$ represents the threshold for class c computing from the current sample. 
$b$ is the global proportion of pixels used for loss calculation, like pixels with top 80\% prediction confidence.
$\alpha_{k-1}^{c}$ denotes the threshold of class c in $(k-1)$th iteration, 
and $d$ regularizes the proportion of classes with low $\alpha_{k-1}^{c}$.
In other words, after sorting the $p_m^c$ in descending order, 
the $[b (e^{\alpha_{k-1}^{c}-1} )^{d}|p_{m}^{c}|]$th item is utilized as the threshold $\alpha_{k'}^{c}$.

This class-level threshold adjustment strategy has three advantages compared with a fixed threshold:
First, it 
considers the difficulty of different categories 
so that the classes with lower prediction confidence probabilities have lower thresholds.
Second, the threshold is updated on each image, 
which is more suitable for the current segmentation result. 
Third, parameters $b$ and $d$ make each class balance between loss contribution and noise suppression, 
presented in Fig.~\ref{fig:threshold}.

Our threshold-adaptative unsupervised focal loss is finally obtained 
with the class-level threshold adjustment strategy.
The calculation method for $I_t$ is updated to 
$I_{t}=\{(h,w)|argmax(p_t^{h,w}=c), p_{m}^{h,w}>\alpha[c], \alpha[c]\in [0,1]\}$.
Then we analyze the gradient of these entropy-based losses, shown in Fig.~\ref{fig:gradient_vis}. 
For simplicity, we consider the binary classification case. 
Use $p$ and $\hat{p}$ to denote learnable classification probability and estimated prediction. 
Shannon entropy tends to predict at 0 and 1, sharpening the prediction distribution, 
and its gradient is strongly biased toward easy samples.
Maximum square loss reduces the gradient when $p$ approaches 1, but the gradient of hard samples with $p$ near 0.5 tends to be 0.
When set $\hat{p}=0.6$, neutral cross-entropy loss shifts the global minimum towards the middle location (0.82). 
However, it utilizes a high confidence threshold (0.8), 
and the samples with low confidence ($p<0.8$) are not included in the loss calculation.
Our threshold-adaptative unsupervised focal loss also shifts the global minimum to the middle location (0.67), 
with a milder gradient neutralization mechanism and dynamic threshold adjustment strategy. 
Moreover, the gradient increases when $p$ gets to 0.5, making hard samples contribute more to the optimization process and get optimized.







\subsection{CIM}
\begin{figure}[t]
	\centering
	\captionsetup[subfloat]{font=scriptsize,labelfont=scriptsize}
	\subfloat[$x_s$]{\includegraphics[width=0.2\linewidth]{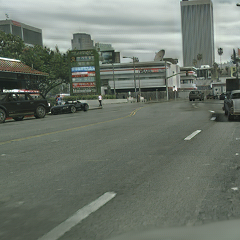}}
	\subfloat[$x_{s'}$]{\includegraphics[width=0.2\linewidth]{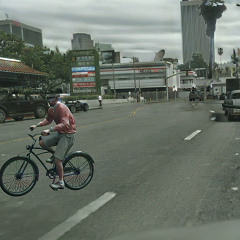}}
	\subfloat[$x_t$]{\includegraphics[width=0.2\linewidth]{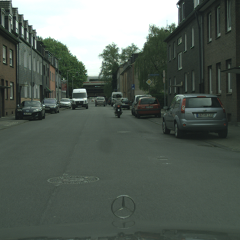}}
	\subfloat[$x_m$]{\includegraphics[width=0.2\linewidth]{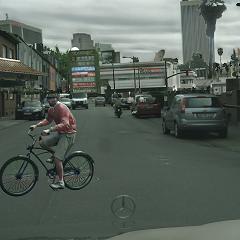}}
	\subfloat[$I_m$]{\includegraphics[width=0.2\linewidth]{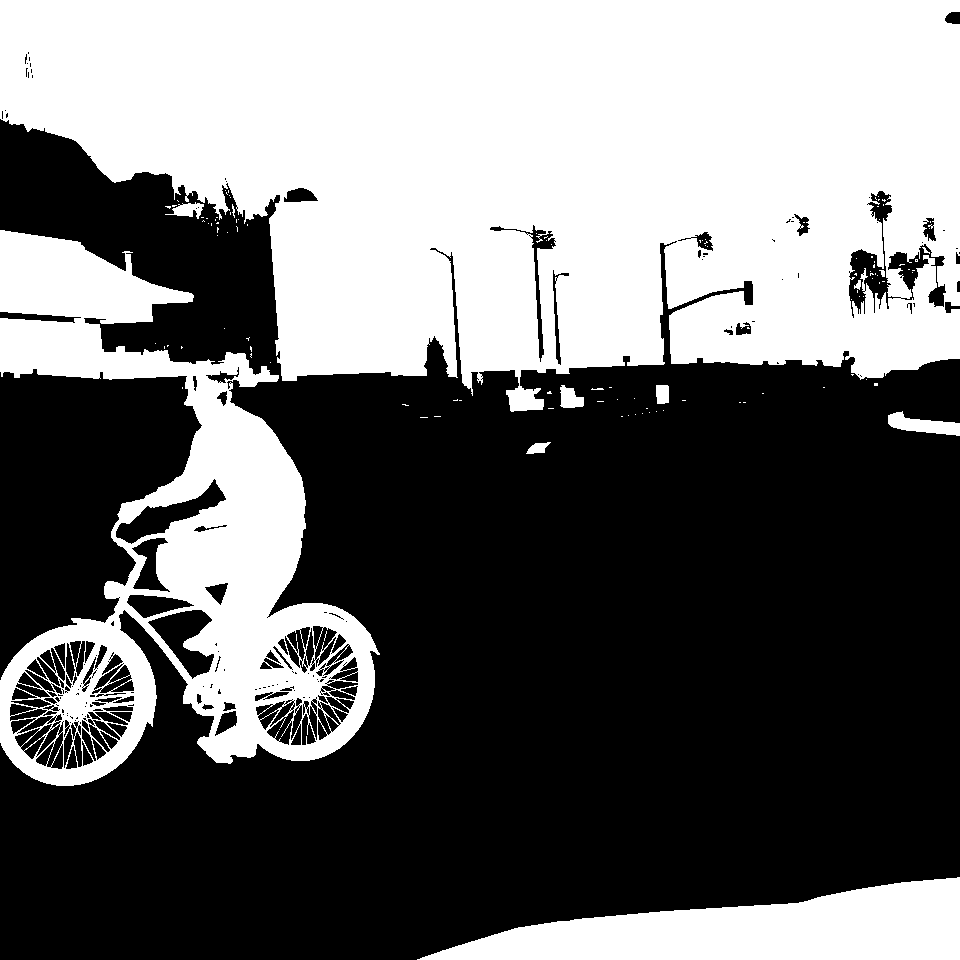}}
	\\ \vspace{-8pt}
	\subfloat[$y_s$]{\includegraphics[width=0.2\linewidth]{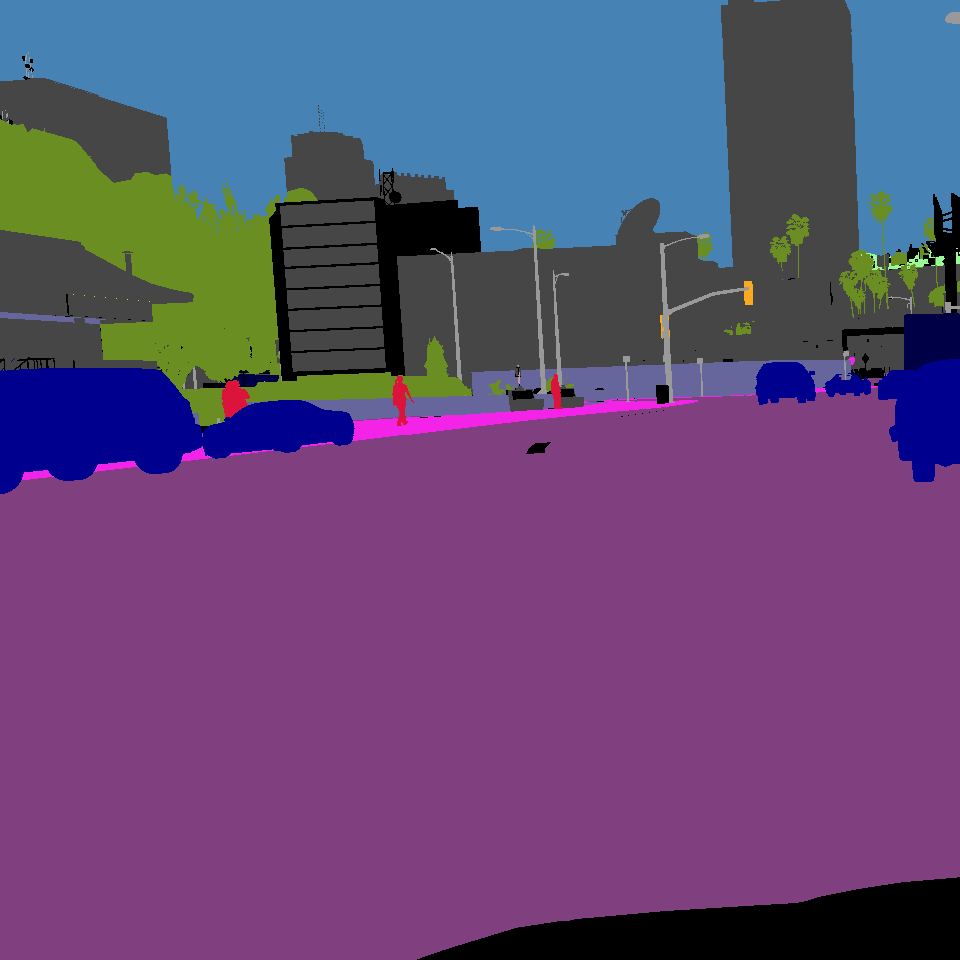}}
	\subfloat[$y_{s'}$]{\includegraphics[width=0.2\linewidth]{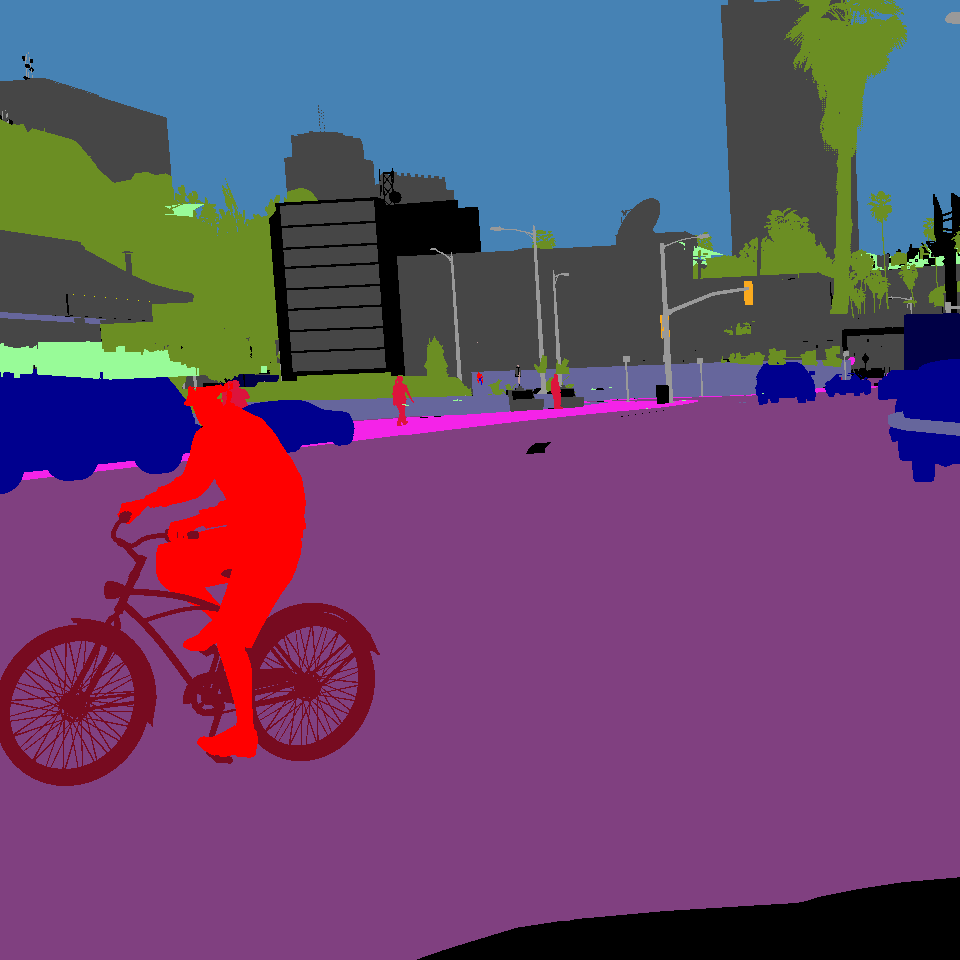}}
	\subfloat[$\tilde{y}_t$]{\includegraphics[width=0.2\linewidth]{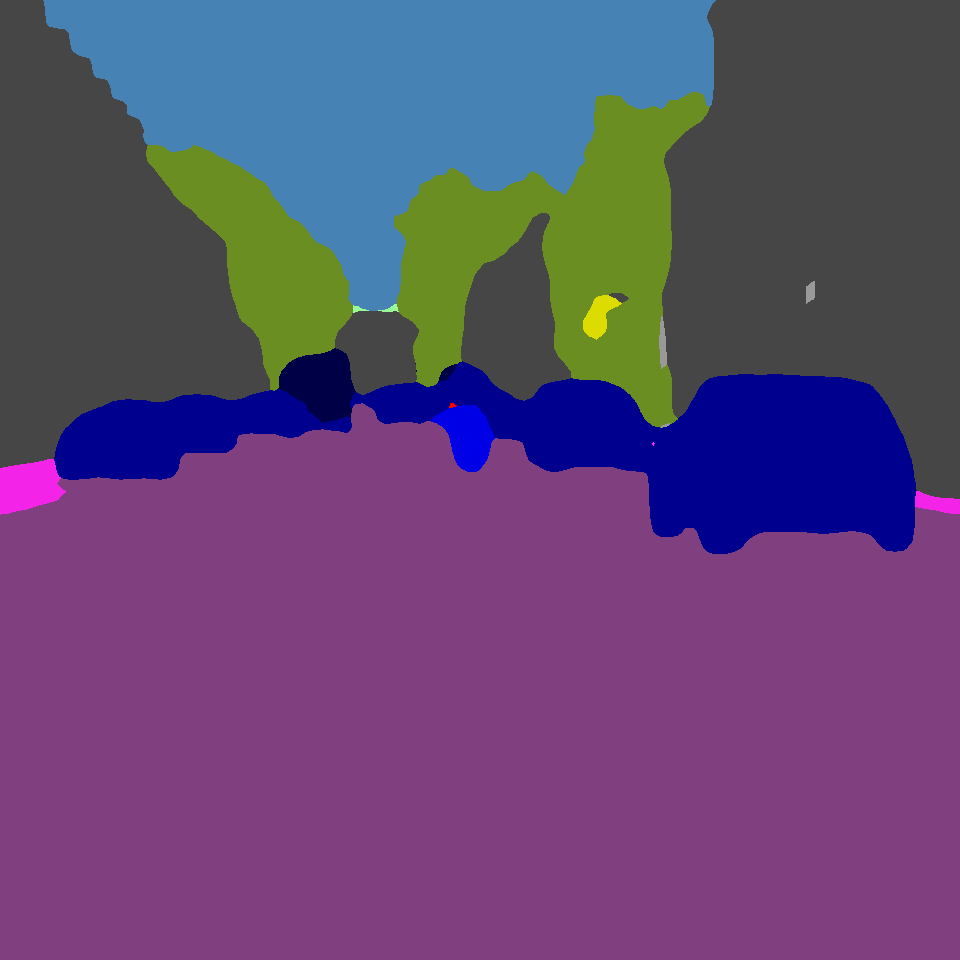}}
	\subfloat[$y_m$]{\includegraphics[width=0.2\linewidth]{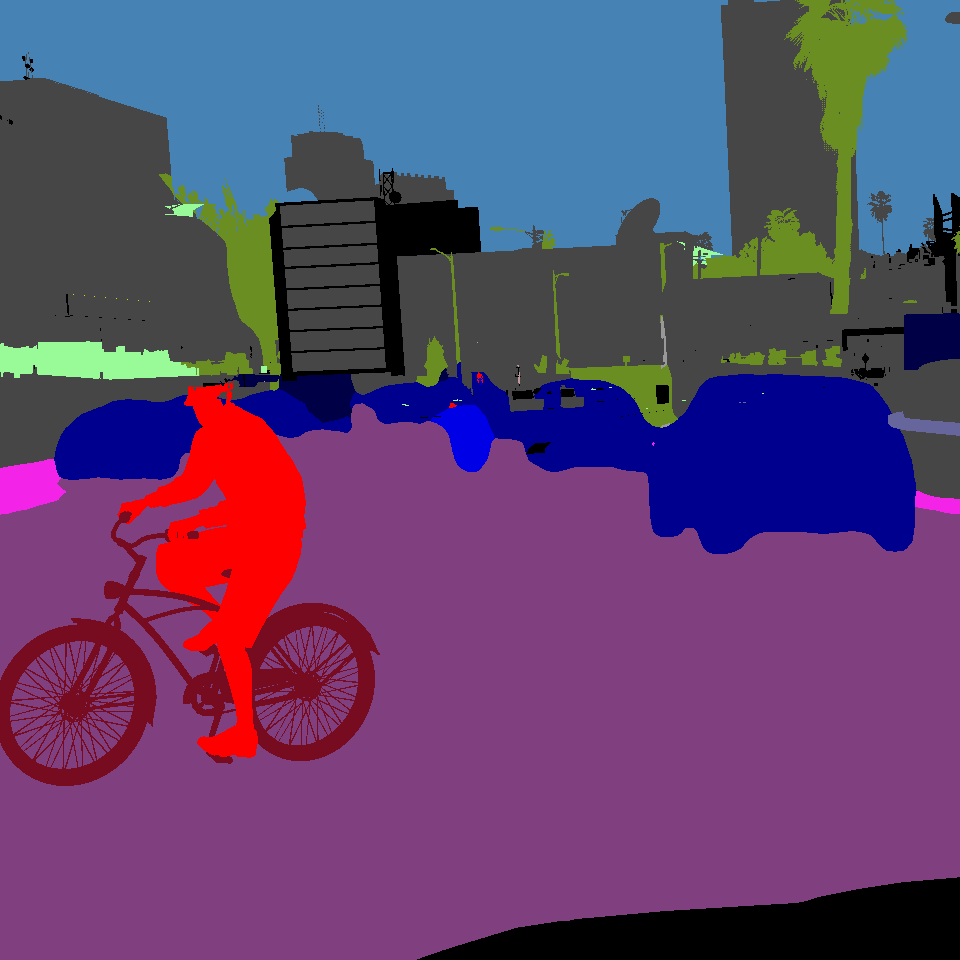}}
	\subfloat[$W_m$]{\includegraphics[width=0.2\linewidth]{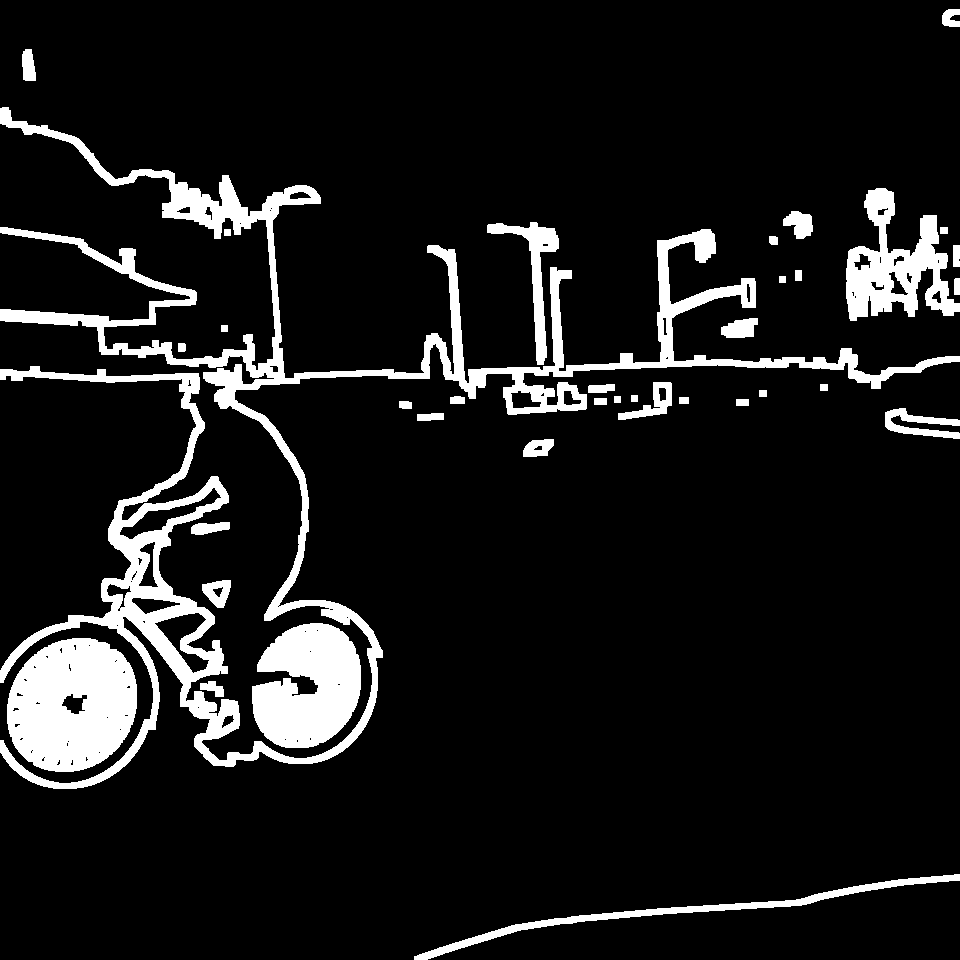}}
	\caption{Qualitative demonstration of Cross-domain Image Mixing. 
	($x_s$,$y_s$), ($x_{s'}$,$y_{s'}$), ($x_t$,$\tilde{y}_t$), and ($x_m$,$y_m$) represent the image-label pairs from the source domain, source domain after long-tail class pasting, target domain, and CIM.
	$I_m$ is the mixing mask and $W_m$ is the coefficient mask.
	}
	\label{fig:mixed_samples}
\end{figure}

To improve the adaptation performance on long-tail classes, 
we use the image-label pairs in the source domain to make a category database and conduct long-tail class pasting on the source image.
Specifically, we build the mapping between classes and source domain image-label pairs offline: $M_{c_i}=\{(x_{s},y_{s})|c_i \in y_{s}\}$, where $c_i$ is the ith class. 
When class $c_i$ is demanded, an image-label pair $(x_s, y_s)$ from $M_{c_i}$ is randomly selected, 
and the corresponding contents are pasted into the current sample.
In the first and second columns in Fig.~\ref{fig:mixed_samples},
the rider and bike are pasted to the source image and label.
The inverse of the dynamic threshold $\alpha$ after softmax is used as the probability of the class selection.
The enhanced image-label pair from the source domain are denoted as ($x_{s'}$, $y_{s'}$).

Denote the adapted model from stage one as $f(\tilde{\theta})$.
The target image $x_t$ goes through $f(\tilde{\theta})$ and gets its pseudo label $\tilde{y}_t$. 
The image-label pairs from the source and target domain are mixed at the class level\cite{classmix}: 
\begin{equation}
	\centering
	\begin{aligned}
		x_m &= I_m \cdot x_{s'} + (1-I_m) \cdot x_t \\
		y_m &= I_m \cdot y_{s'} + (1-I_m) \cdot \tilde{y}_t
	\end{aligned}
	\label{eq:class_mix}
\end{equation}
where $I_m$ is the mixing mask obtained by randomly choosing half of the classes in the $y_{s'}$. 
Qualitative demonstration of Cross-domain Image Mixing is shown in Fig.~\ref{fig:mixed_samples}.
The cross-entropy loss is also used for mixed image-label pairs.
Since the receptive field of pixels near the boundaries of the mixing mask $I_m$ (denoted as $p\in b(I_{m})$) 
after passing through the convolutional network will contain both real and virtual contents, 
we enhance the loss contribution of these pixels. 
Expressly, for the pixels within the 7$\times$7 field of the boundary point, 
their loss coefficient $W_m (p)$ is set to 2 while others remain 1:

\begin{equation}
	W_m(p) = 
	\begin{cases}
		2, & if\quad p \in b(I_m) \\
		1, & otherwise
	\end{cases}
\end{equation}


The training loss of stage two is formulated as follows, 
and $\lambda_{m}$ is the coefficient for the mixed part.
\begin{equation}
	\centering
	Loss = L_s(x_s,y_s) + \lambda_{u}L_{u}(x_t) + \lambda_{m}L_{m}(x_m,y_m)
	\label{eq:loss_stage2}
\end{equation}

\section{Experiments}

\subsection{Datasets}
We conduct experiments on two popular synthetic-to-real settings: SYNTHIA-to-Cityscapes and GTA5-to-Cityscapes.
The synthetic datasets SYNTHIA\cite{synthia} and GTA5\cite{playingfordata} are used as the source domain datasets, 
and the actual driving dataset Cityscapes is the target domain dataset.

SYNTHIA consists of 9400 photo-realistic frames, 
which are 760 $\times$ 1280.
GTA5 has 24966 annotated images with a resolution of 1052 $\times$ 1914. 
The label classes of the two synthetic datasets are consistent with Cityscapes. 
GTA5 contains 19 classes, while SYNTHIA has 16 categories. 
For the target domain, Cityscapes has 2975 and 500 precisely annotated images for training and validation and 19997 roughly annotated images. 
To fully exploit the advantage of our UDA method, 
we use 19997 images for BiSeNet like\cite{neutral},
while 2975 images for DeepLabV2 for a fair comparison.

\subsection{Segmentation Network}
For training efficiency and real-time inference,
BiSeNet\cite{bisenet} with ResNet18\cite{resnet} as the backbone is adopted as the segmentation network in most experiments.
Meanwhile, DeepLabV2 with ResNet101 as the backbone is also utilized for a fair comparison with other state-of-the-art methods.

\subsection{Evaluation Methods and Experiment Settings}
Intersection-over-union (IoU) for each class and mean-intersection-over-union (mIoU) for all classes are the evaluation metrics in our experiments.
For GTA5-to-Cityscapes, 19 classes are evaluated, 
while 16 and 13 are for the SYNTHIA-to-Cityscapes setting.

For BiSeNet, the images in SYNTHIA keep their original size of 760 $\times$ 1280, 
while the ones in GTA5 are randomly cropped into 1000 $\times$ 1000 during the training. 
The model is pretrained on the source domain dataset.
The batch size is 12, and total training iterations are 20000, using two 1080Ti GPUs.   
We use the SGD optimizer with the learning rate of 2.5$\times10^{-4}$, momentum of 0.9, and weight decay of 5$\times10^{-4}$.
For learning rate adjustment, we use warm-up for the first 100 iterations, 
followed by a poly-type tuning strategy, decaying the initial learning rate with $(1-\frac{iter-warm\_iter}{total\_iter-warm\_iter})^{0.9}$.
The cross-entropy loss is adopted as the optimization objective.

In stage one of our UDA framework, unsupervised loss on the target domain is added to the optimization process.
Specifically, shannon entropy loss, maximum square loss, neutral cross-entropy loss, and our threshold-adaptative unsupervised focal loss are adopted as the $L_{u}$.
We first compare their performance with fixed confidence thresholds from 0.2 to 0.8 
and then apply the dynamic threshold adjustment strategy to our unsupervised focal loss. 
The image perturbation methods are like\cite{neutral}, including random flipping, scaling, rotation, Gaussian noise, etc.
The batch size is set as 2 for both source 
and target domain images and the initial learning rate is decayed as 0.5$\times10^{-5}$, 
training for 20000 iterations.

In stage two of our UDA framework, mixed loss item $L_{m}$ is added to the optimization loss.
The main segmentation network for optimization $f(\theta)$ loads weights from the pretrained model on the source domain,
while the segmentation network $f(\tilde{\theta})$ for producing the mixed image-label pairs is 
initialized from the weights of the model obtained in stage one.
The batch size is also 2, with a learning rate of 0.5$\times 10^{-5}$ and 20000 iterations.
The $\lambda_u$ and $\lambda_m$ are set as 0.05 and 1. 

In the testing process, 
we conduct experiments on the Cityscapes validation set to calculate the evaluation metrics of the UDA methods like most previous work.
For DeepLabV2, we follow most of the settings in ProDA\cite{prototypical}, 
and two 3090 GPUs are used for training and inference.

\subsection{Experimental results}
\subsubsection{SYNTHIA-to-Cityscapes}
The overall performance of our UDA method on the SYNTHIA-to-Cityscapes is shown in Table \ref{tab:synthia_comparison}.
The baseline and oracle models are trained on the transferred source domain (IST) and target domain. 
The Shannon, Maximum, Neural represent shannon entropy loss, maximum square loss, and neural cross-entropy loss.

\begin{table*}[ht]
	\caption{The adaptation performance and comparison on SYNTHIA-to-Cityscapes 
	(mIoU: 16 classes; mIoU*: 13 classes)}
	\label{tab:synthia_comparison}
	\resizebox{\linewidth}{!}{
	\begin{tabular}{c|c|cccccccccccccccc|c|c}
		\hline
		Methods & Network & road & sw & build & wall$^{*}$ & fence$^{*}$ & pole$^{*}$ & light & sign & vege & sky & person & rider & car & bus & motor & bike & mIoU & mIoU$^{*}$ \\ \hline
		AdaptSeg\cite{leanringoutput} & \multirow{13}{*}{\begin{tabular}{c} DeeplabV2\\(ResNet101) \end{tabular}} & 84.3 & 42.7 &77.5 &-&-&-&4.7 &7.0 &77.9 &82.5 &54.3 &21.0 &72.3 &32.2 &18.9 &32.3& - & 46.7\\
		Shannon\cite{advent} &  & 85.6 & 42.2 & 79.7 & 8.7 & 0.4 & 25.9 & 5.4 & 8.1 & 80.4 & 84.1 & 57.9 & 23.8 & 73.3 & 36.4 & 14.2 & 33.0 & 41.2 & 48.0 \\ 
		Maximum\cite{maximum} &  & 82.9 & 40.7 & 80.3 & 10.2 & 0.8 & 25.8 & 12.8 & 18.2 & 82.5 & 82.2 & 53.1 & 18.0 & 79.0 & 31.4 & 10.4 & 35.6 & 41.4 & 48.2 \\ 
		BDL \cite{bidirectional} & & 86.0 & 46.7 & 80.3 & - & - & - & 14.1 & 11.6 & 79.2 & 81.3 & 54.1 & 27.9 & 73.7 & 42.2 & 25.7 & 45.3 & - & 51.4 \\ 
		FDA\cite{fda} & & 79.3 & 35.0 & 73.2 & - & - & - & 19.9 & 24.0 & 61.7 & 82.6 & 61.4 & 31.1 & 83.9 & {40.8} & {38.4} & 51.1 & - & 52.5 \\ 
		Confidence\cite{confidence} & & 87.6 & 46.1 & 82.0 & 10.0 & 0.4 & 33.6 & 21.4 & 14.9 & 81.2 & 85.2 & 57.2 & 26.4 & 83.0 & 33.3 & 24.0 & 46.8 & 45.8 & 53.0 \\
		DACS\cite{dacs} &  & 80.6 & 25.1 & 81.9 & {24.5} & 2.9 & 37.2 & 22.7 & 24.0 & 83.7 & {90.8} & {67.6} & {38.3} & 82.9 & {38.9} & 28.5 & 47.6 & 48.3 & 54.8 \\ 
		Rectifying\cite{rectifying} & & 87.6 & 41.9 & 83.1 & 14.7 & 1.7 & 36.2 & 31.3 & 19.9 & 81.6 & 80.6 & 63.0 & 21.8 & 86.2 & {40.7} & 23.6 & {53.1} & 47.9 & 54.9 \\ 
		IAST\cite{iast} &  & 81.9 & 41.5 & 83.3 & 17.7 & {4.6} & 32.3 & 30.9 & 28.8 & 83.4 & 85.0 & 65.5 & 30.8 & {86.5} & 38.2 & {33.1} & 52.7 & 49.8 & 57.0 \\ 
		SAC\cite{sac} & & {89.3} & 47.2 & \bfseries{85.5} & {26.5} & 1.3 & 43.0 & {45.5} & 32.0 & 87.1 & 89.3 & 63.6 & 25.4 & {86.9} & 35.6 & 30.4 & {53.0} & {52.6} & {59.3} \\ 
		CorDA\cite{corda} & & \bfseries{93.3} & \bfseries{61.6} & 85.3 & 19.6 & \bfseries{5.1} & 37.8 & 36.6 & 42.8 & 84.9 & 90.4 & 69.7 & \bfseries{41.8} & 85.6 & 38.4 & 32.6 & \bfseries{53.9} & 55.0 & 62.8 \\ 
		ProDA\cite{prototypical} & & 87.8 & 45.7 & 84.6 & {37.1} & 0.6 & 44.0 & \bfseries{54.6} & 37.0 & 88.1 & 84.4 & {74.2} & 24.3 & {88.2} & {51.1} & {40.5} & 45.6 & {55.5} & {62.0}\\
		Ours & & 88.6 & 52.4 & \bfseries{85.5} & \bfseries{39.4} & 0.3 & 44.9 & 51.4 & \bfseries{60.3} & 88.1 & 88.1 & \bfseries{75.5} & 28.6 & \bfseries{88.7} & \bfseries{52.5} & \bfseries{42.2} & 48.2 & \bfseries{58.4} & \bfseries{65.4} \\ \hline
		Oracle & \multirow{8}{*}{\begin{tabular}{c} BiSeNet\\(ResNet18) \end{tabular} } & 85.1 & 74.0 & 80.2 & 46.4 & 49.7 & 50.7 & 55.1 & 64.8 & 89.4 & 66.0 & 70.0 & 47.2 & 83.5 & 72.6 & 45.2 & 66.5 & 65.4 & 69.2 \\ \cline{1-1} \cline{3-20}
		Baseline (IST) &  & 64.0 & 34.8 & 65.3 & 10.2 & 0.7 & 37.5 & 24.7 & 37.5 & 85.1 & 87.9 & 53.2 & 19.1 & 70.5 & 14.3 & 9.2 & 38.7 & 40.8 & 46.5 \\ 
		Shannon \cite{advent} &  & 84.9 & 41.9 & 80.0 & 5.1 & 0.7 & 37.1 & 25.4 & 37.5 & 85.2 & 87.4 & 52.8 & 19.7 & 71.9 & 17.9 & 9.4 & 34.9 & 43.2 & 49.9 \\
		Maximum \cite{maximum} &  & 51.2	& 31.2 & 57.8 & 8.6	& 0.3 & 42.0 & 28.2 & 35.4 & 86.5 & 88.9 & 56.9 & 16.8 & 72.4 & 15.8 & 9.5 & 34.3 & 39.7 & 45.0 \\ 
		Confidence\cite{confidence} & & 85.9  & 44.0  & 78.2  & 6.0  & 0.6  & 36.8  & 22.1  & 32.5  & 85.7  & 87.1  & 54.9  & 19.7  & 81.5  & 23.0  & 9.7  & 33.4  & 43.8  & 50.6  \\
		Neutral\cite{neutral} & & 85.7 & 46.6 & 81.6 & 10.7 & 0.7 & 38.1 & 23.9 & 37.2 & 86.1 & 88.1 & 54.7 & 19.5 & 74.1 & 25.9 & 9.5 & 36.5 & 44.9 & 51.5\\ 
		Ours(stage 1) & & {89.8} & 52.0 & 83.9 & 14.9 & 0.5 & 44.2 & 32.1 & 42.3 & 87.6 & 89.5 & 58.6 & 19.0 & 84.4 & 34.3 & 11.8 & 39.5 & 49.0 & 55.7	\\
		Ours(stage 2) & & 87.3 & {53.4} & 82.9 & 20.5 & 0.4 & \bfseries{45.6} & {45.5} & {58.7} & \bfseries{88.6} & \bfseries{91.7} & 66.4 & 25.3 & 79.2 & 29.3 & 13.5 & 46.3 & 52.2 & 59.1 \\ \hline
	\end{tabular}}
\end{table*}

\begin{figure*}[htbp]
	\centering
	\captionsetup[subfloat]{font=scriptsize,labelfont=scriptsize}
	\subfloat{\includegraphics[width=0.125\linewidth]{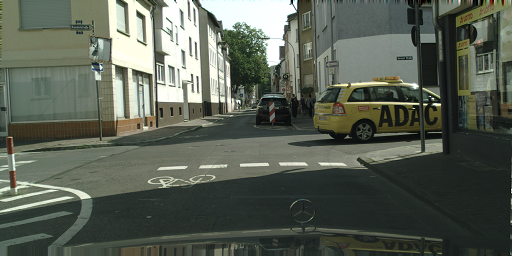}}
	\subfloat{\includegraphics[width=0.125\linewidth]{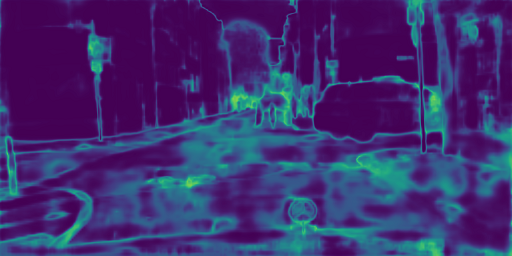}}
	\subfloat{\includegraphics[width=0.125\linewidth]{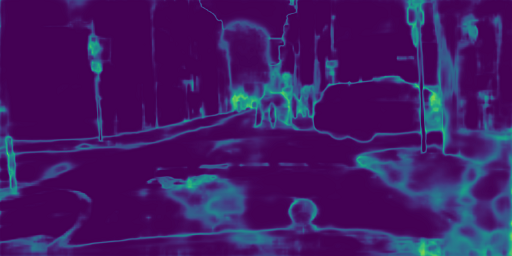}}
	\subfloat{\includegraphics[width=0.125\linewidth]{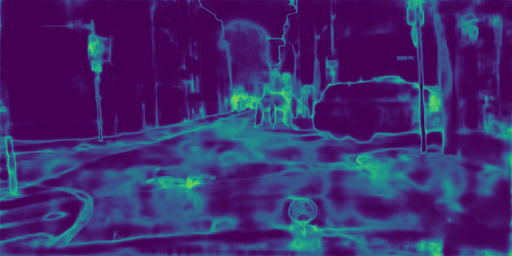}}
	\subfloat{\includegraphics[width=0.125\linewidth]{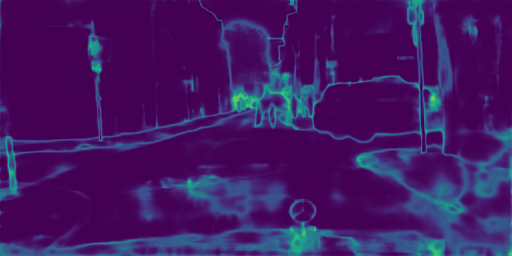}}
	\subfloat{\includegraphics[width=0.125\linewidth]{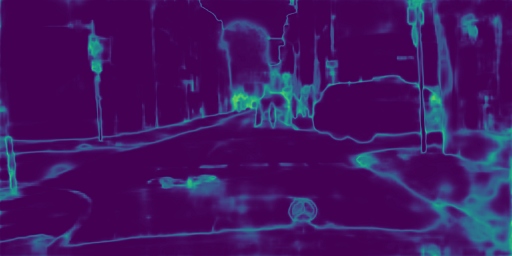}}
	\subfloat{\includegraphics[width=0.125\linewidth]{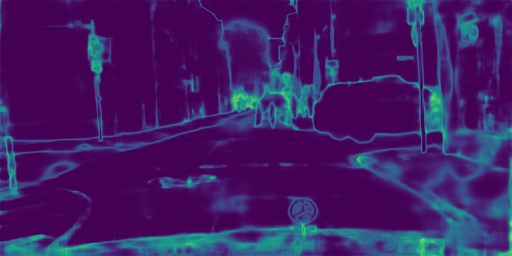}}
	\subfloat{\includegraphics[width=0.125\linewidth]{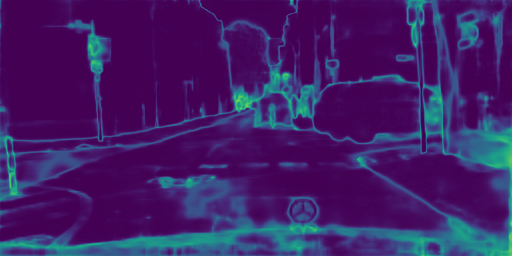}} \\ \vspace{-0.30cm}
	\subfloat{\includegraphics[width=0.125\linewidth]{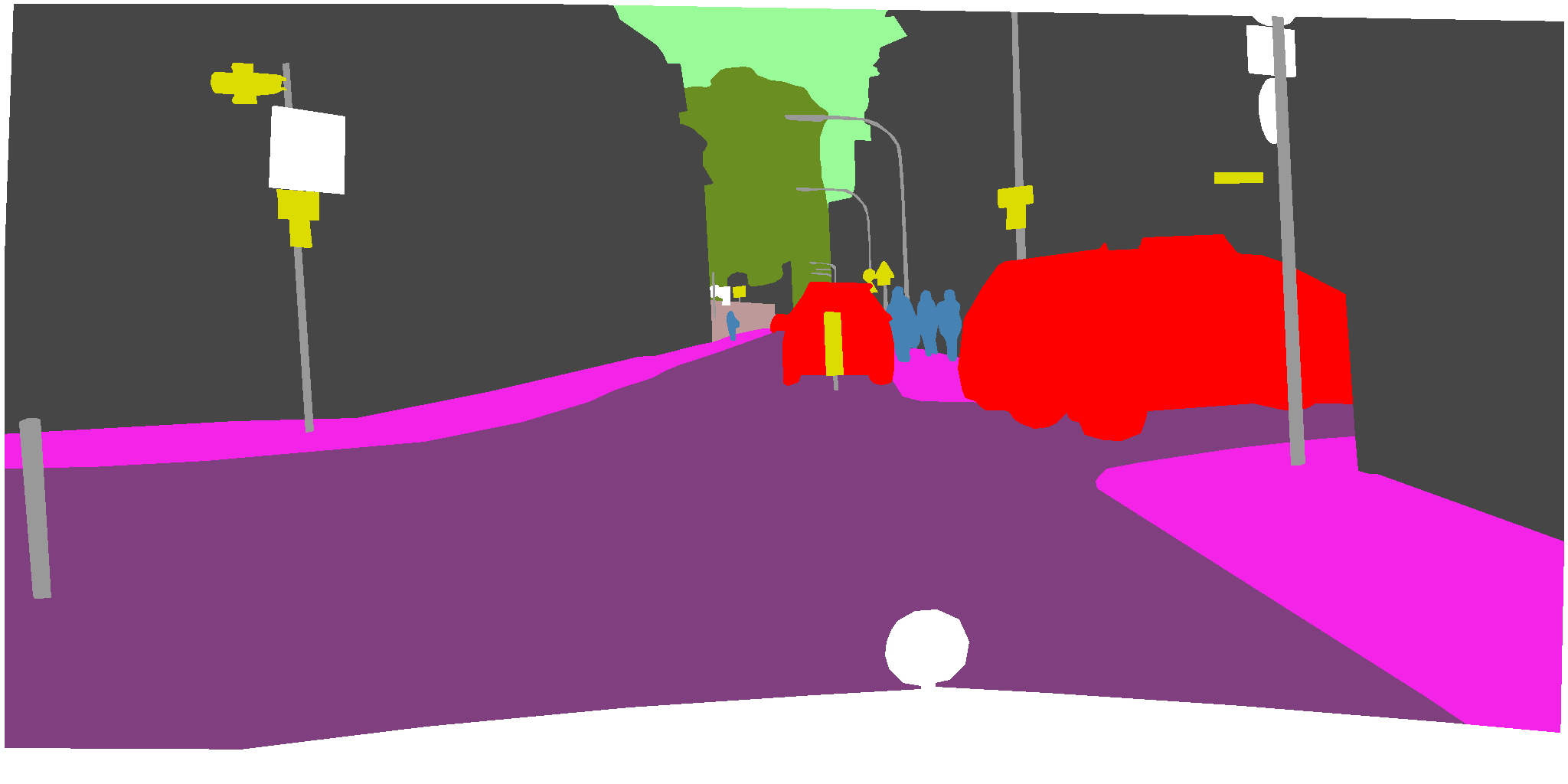}}
	\subfloat{\includegraphics[width=0.125\linewidth]{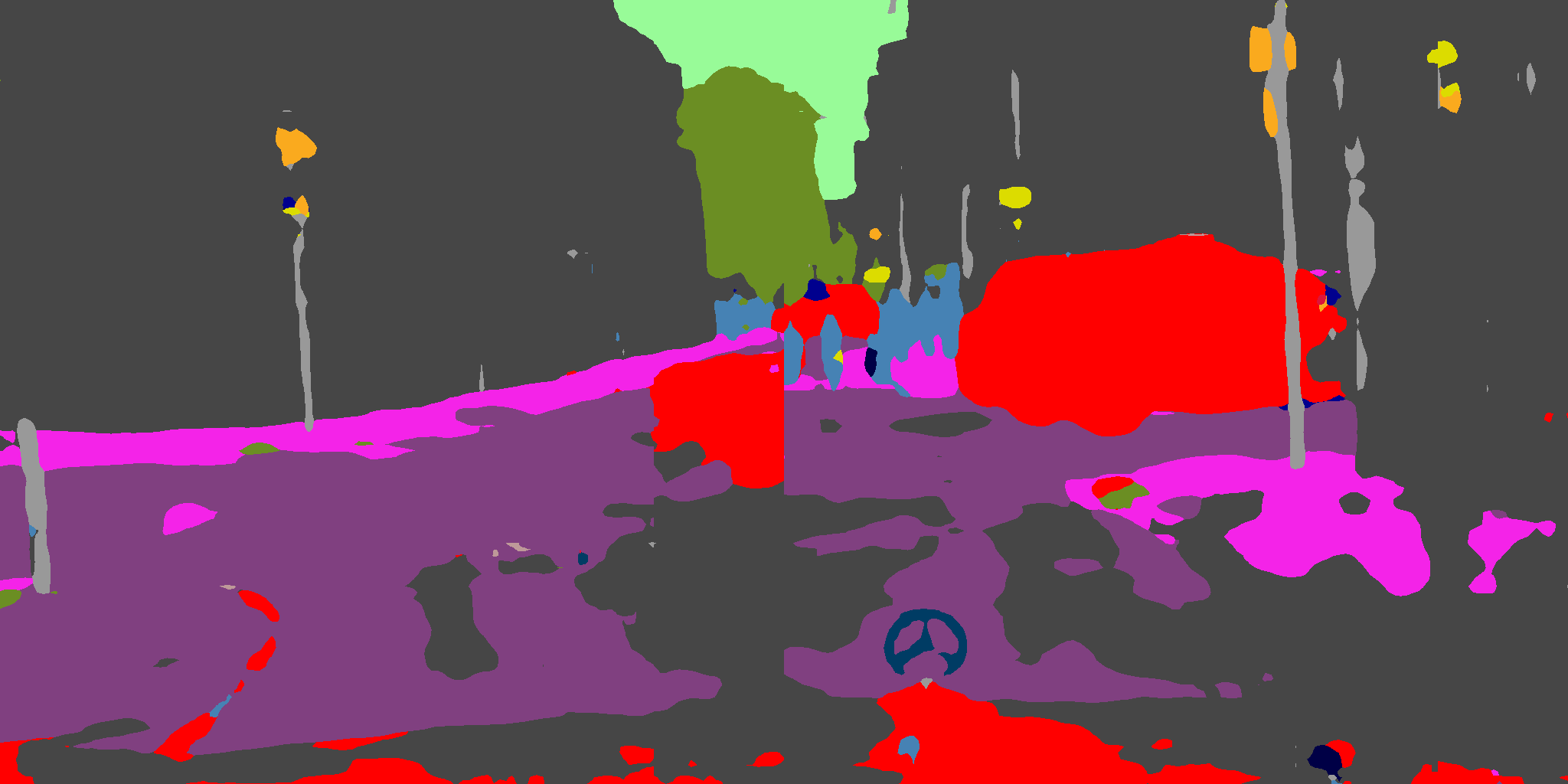}}
	\subfloat{\includegraphics[width=0.125\linewidth]{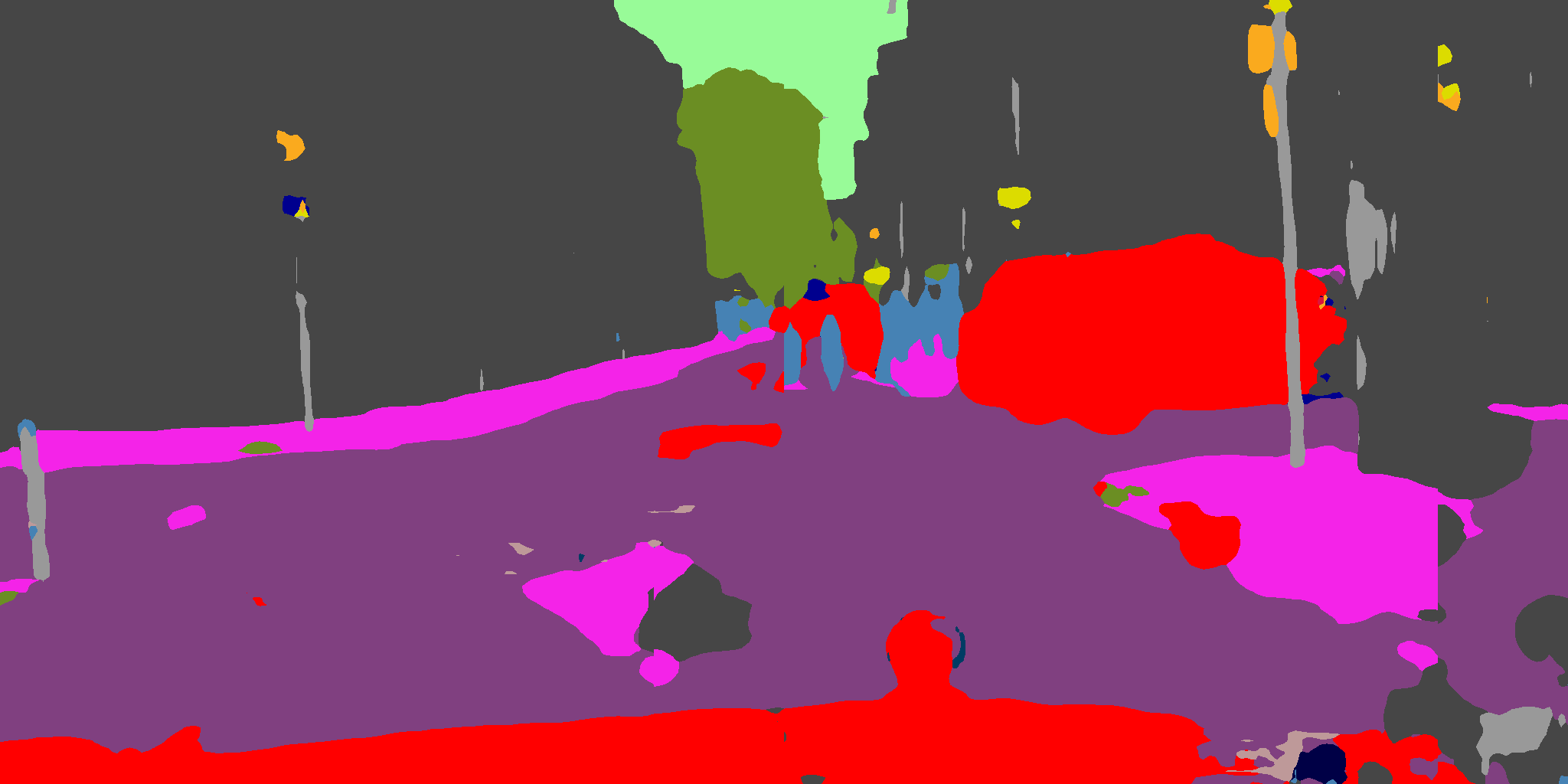}}
	\subfloat{\includegraphics[width=0.125\linewidth]{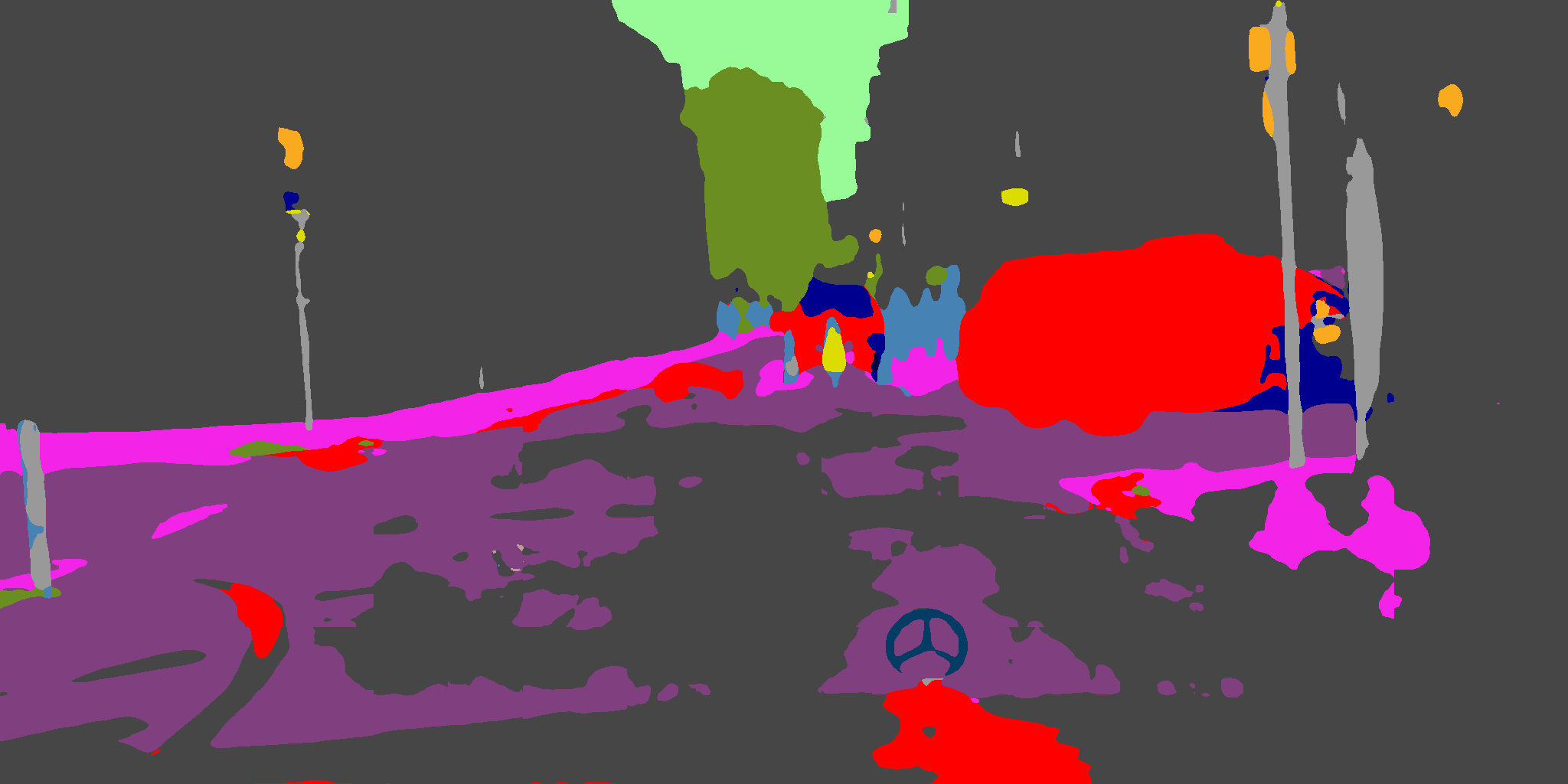}}
	\subfloat{\includegraphics[width=0.125\linewidth]{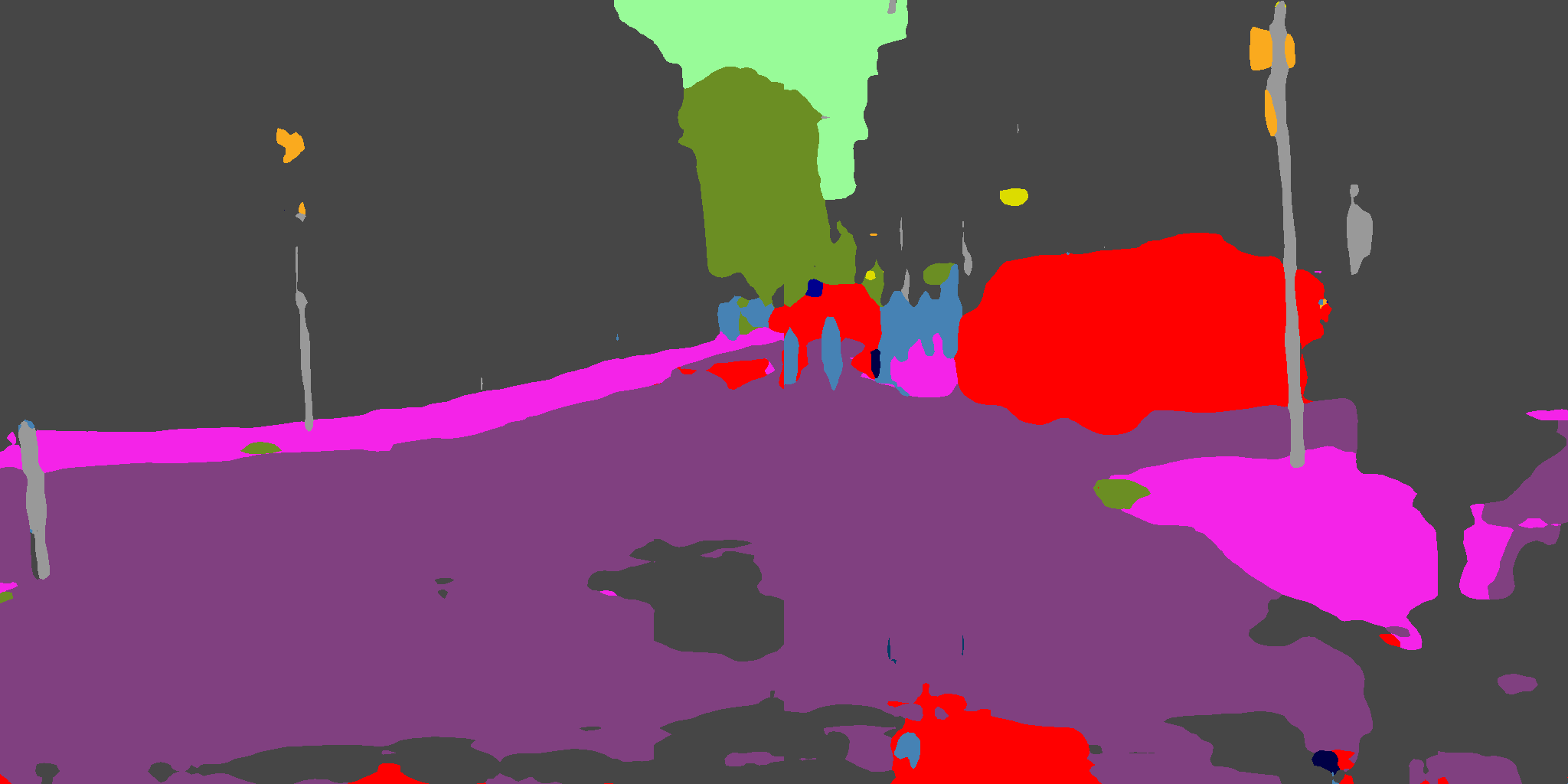}}
	\subfloat{\includegraphics[width=0.125\linewidth]{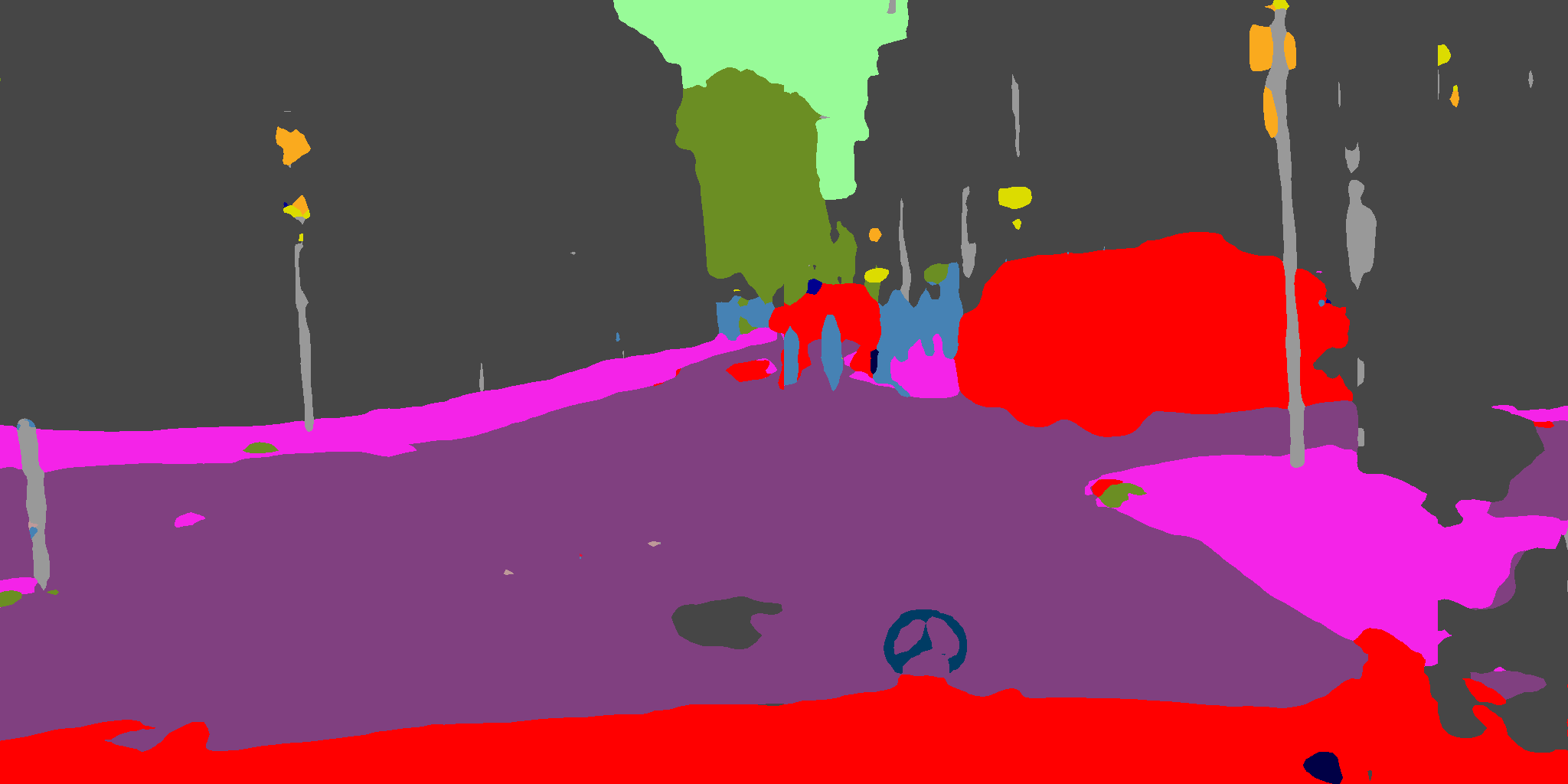}}
	\subfloat{\includegraphics[width=0.125\linewidth]{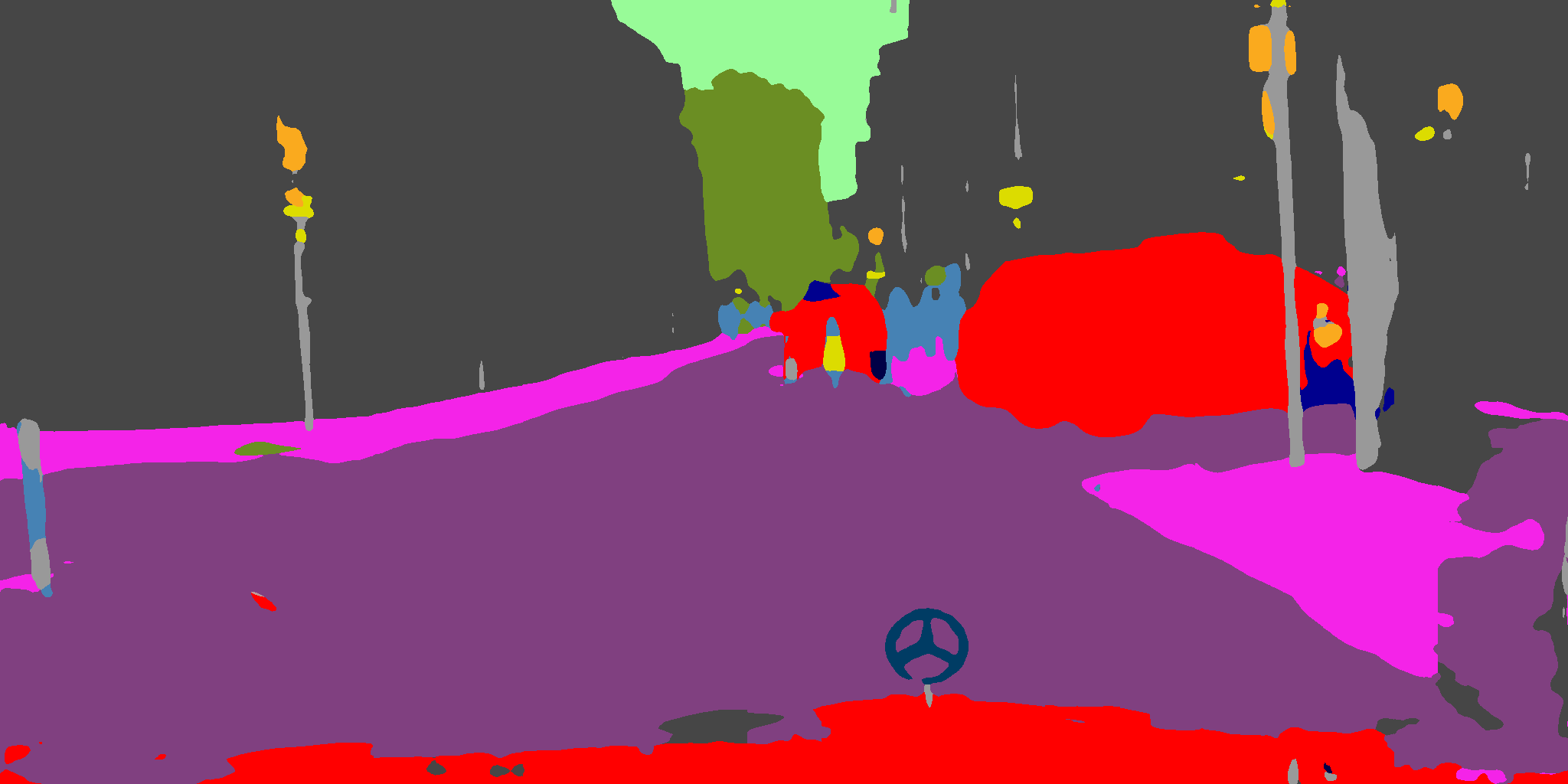}}
	\subfloat{\includegraphics[width=0.125\linewidth]{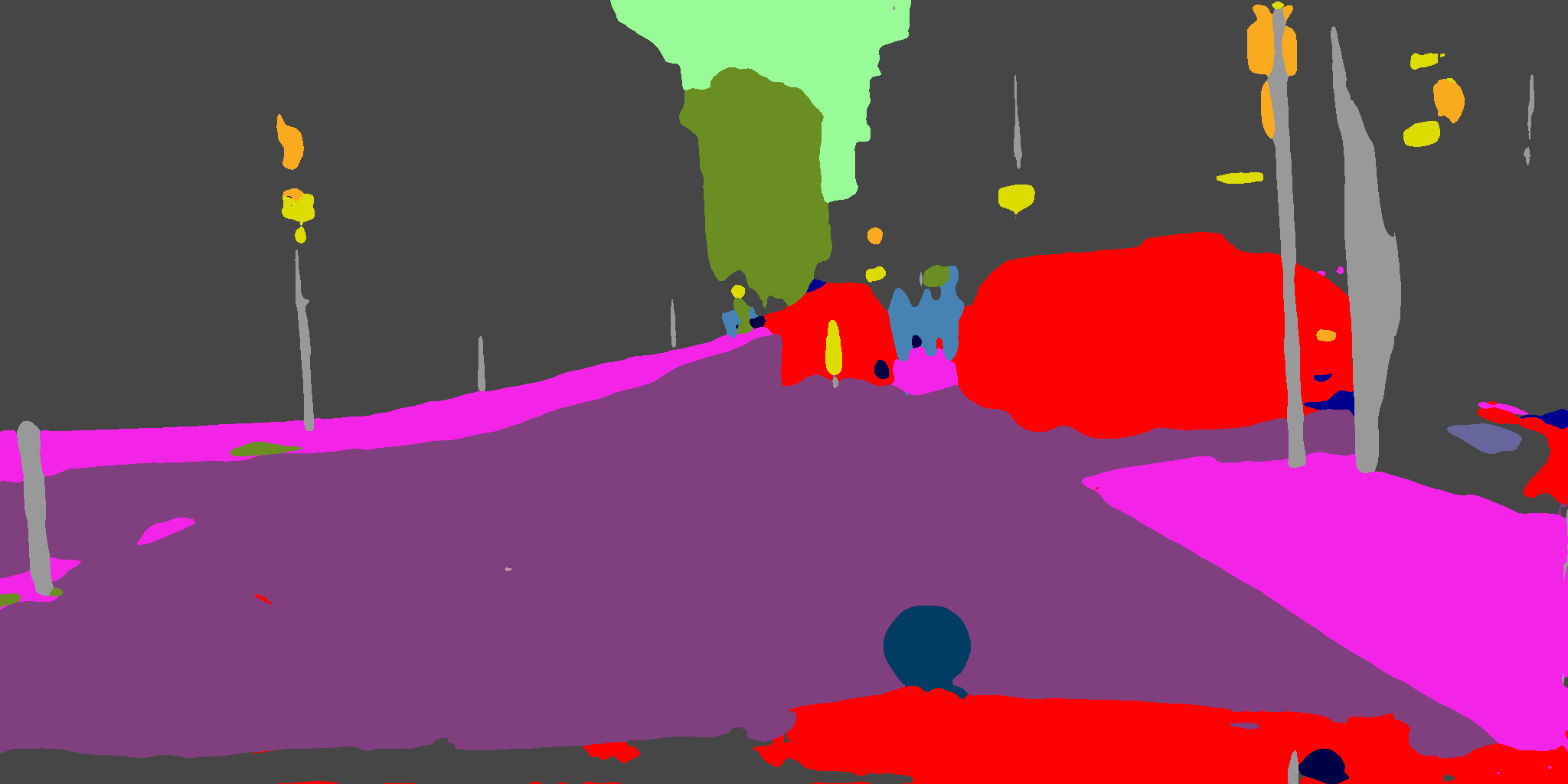}} \\ \vspace{-0.30cm}
	\subfloat{\includegraphics[width=0.125\linewidth]{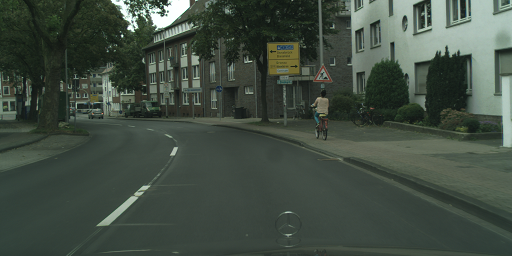}}
	\subfloat{\includegraphics[width=0.125\linewidth]{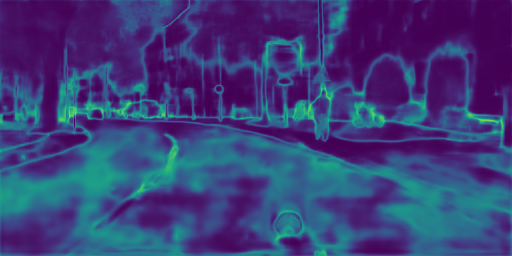}}
	\subfloat{\includegraphics[width=0.125\linewidth]{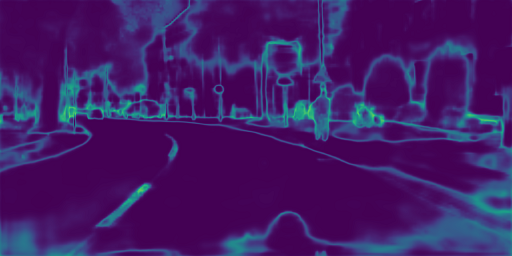}}
	\subfloat{\includegraphics[width=0.125\linewidth]{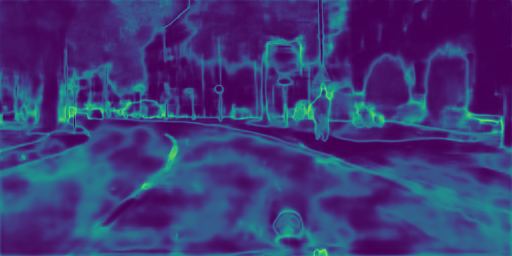}}
	\subfloat{\includegraphics[width=0.125\linewidth]{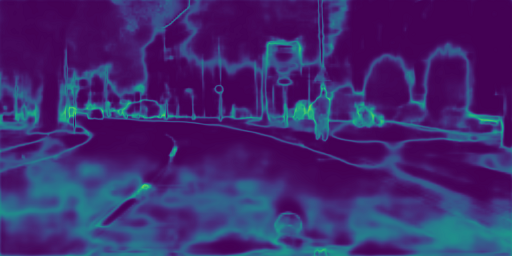}}
	\subfloat{\includegraphics[width=0.125\linewidth]{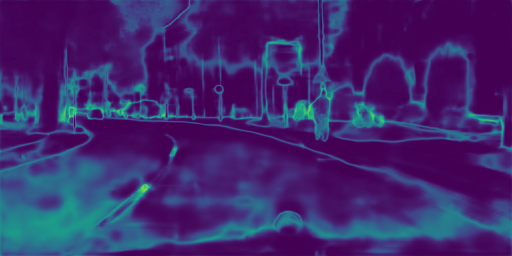}}
	\subfloat{\includegraphics[width=0.125\linewidth]{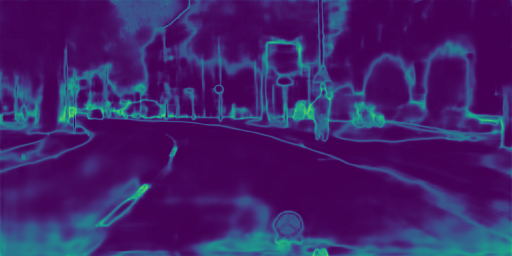}}
	\subfloat{\includegraphics[width=0.125\linewidth]{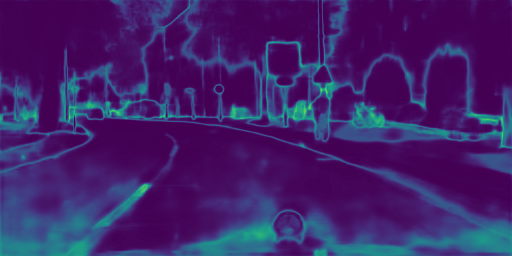}} \\ \vspace{-0.30cm}
	\subfloat[Image/Ground truth]{\includegraphics[width=0.125\linewidth]{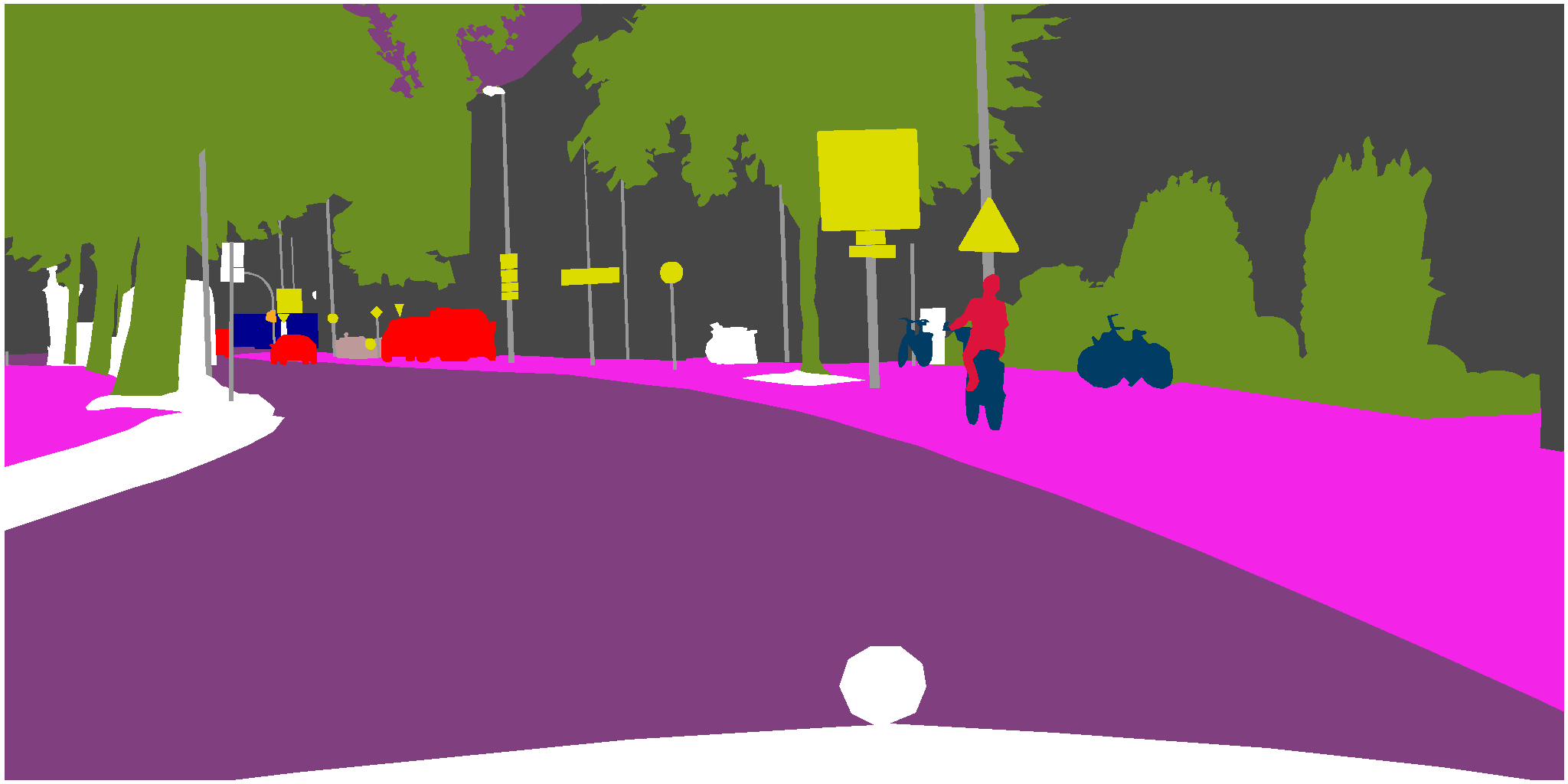}}
	\subfloat[Baseline(IST)]{\includegraphics[width=0.125\linewidth]{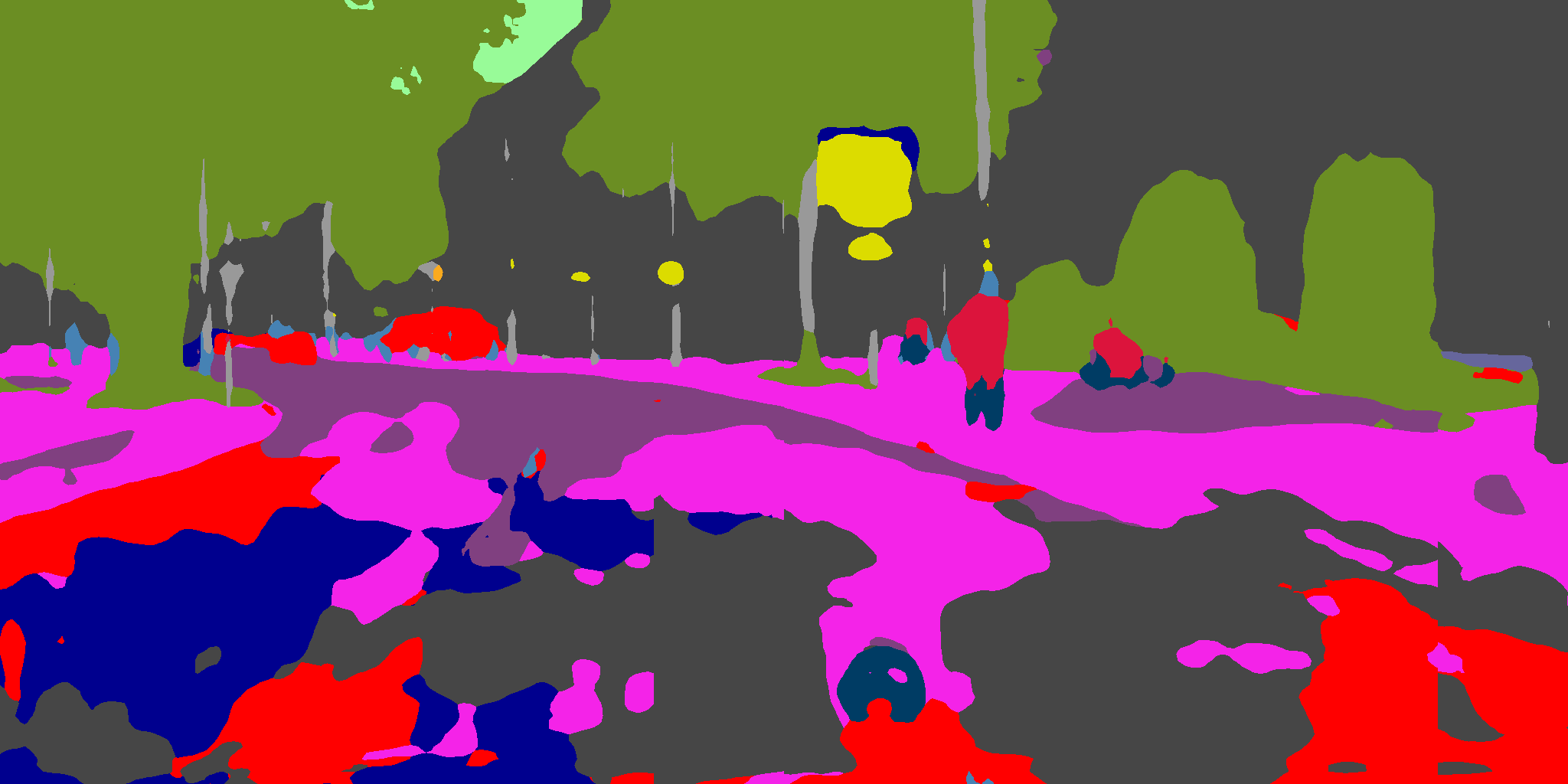}}
	\subfloat[Shannon]{\includegraphics[width=0.125\linewidth]{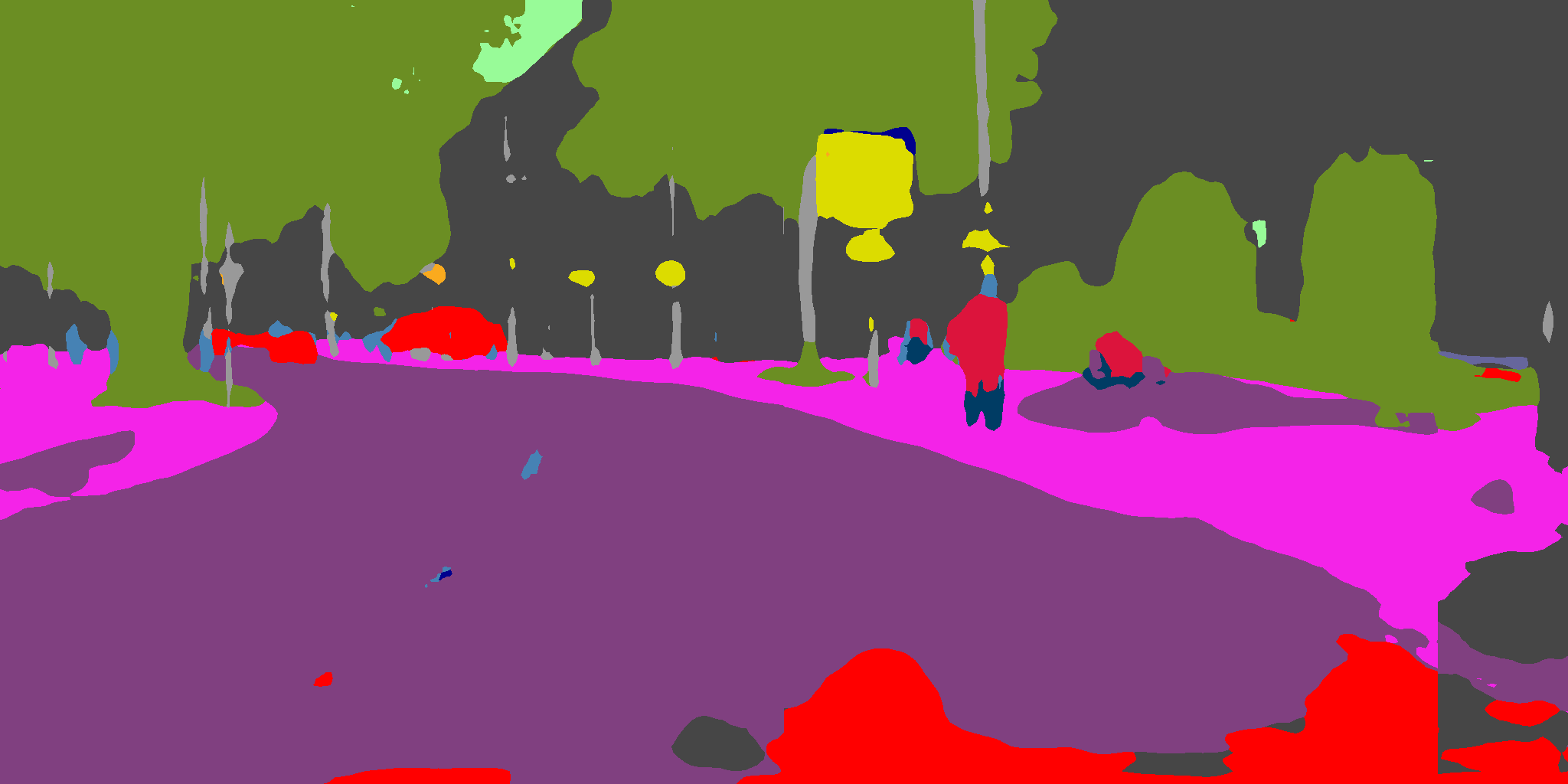}}
	\subfloat[Maximum]{\includegraphics[width=0.125\linewidth]{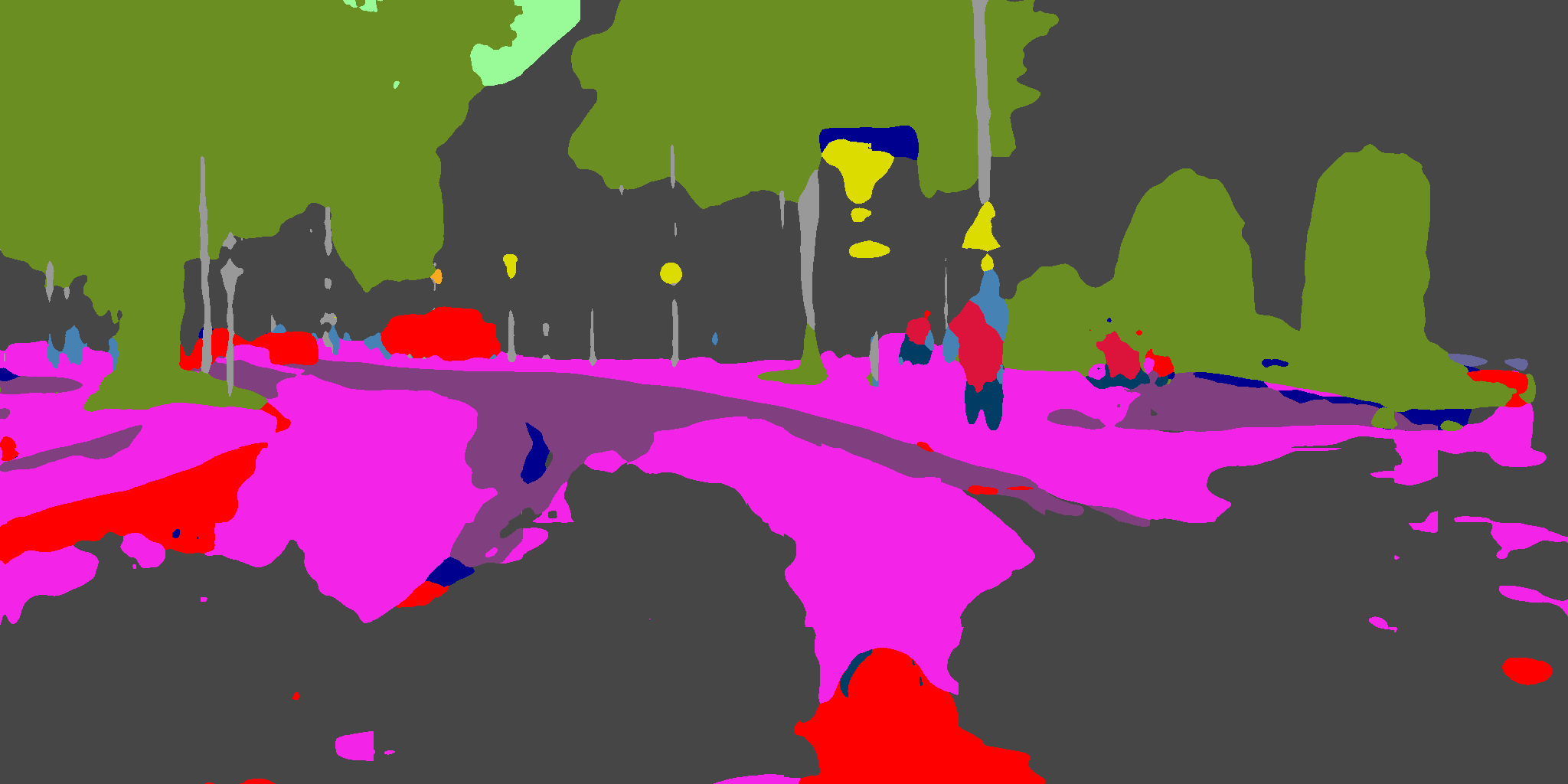}}
	\subfloat[Confidence]{\includegraphics[width=0.125\linewidth]{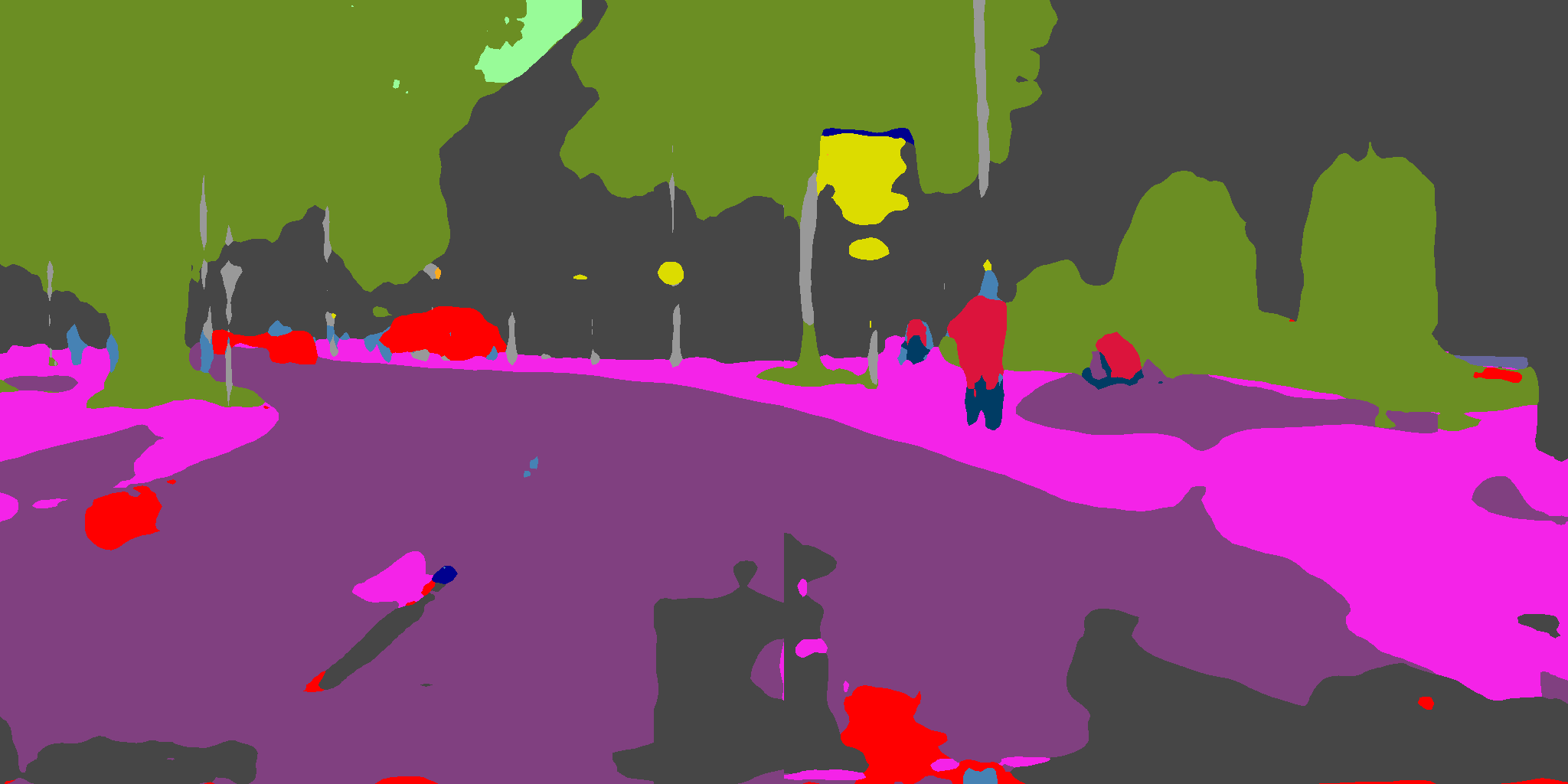}}
	\subfloat[Neutral]{\includegraphics[width=0.125\linewidth]{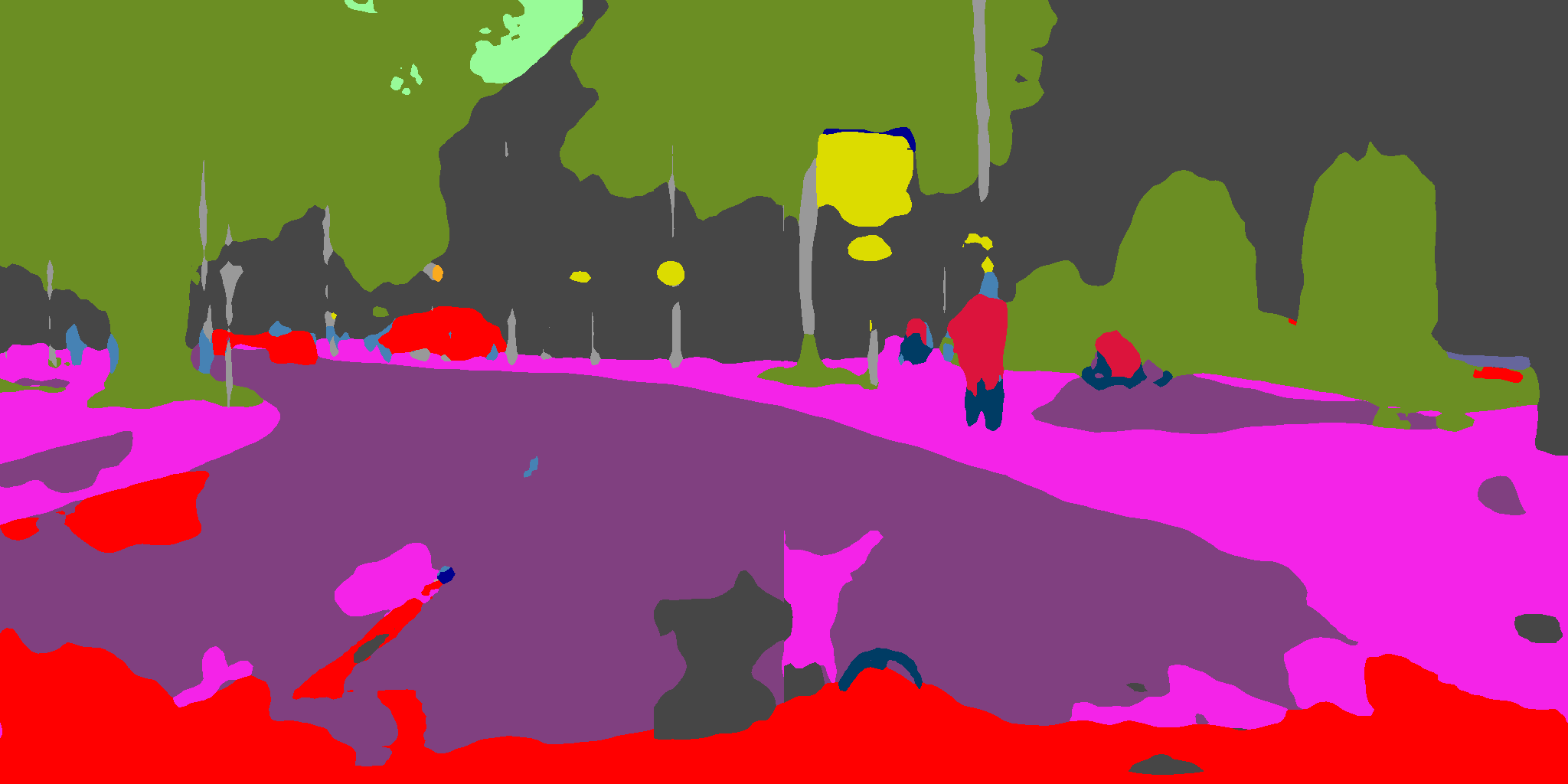}}
	\subfloat[Ours(stage 1)]{\includegraphics[width=0.125\linewidth]{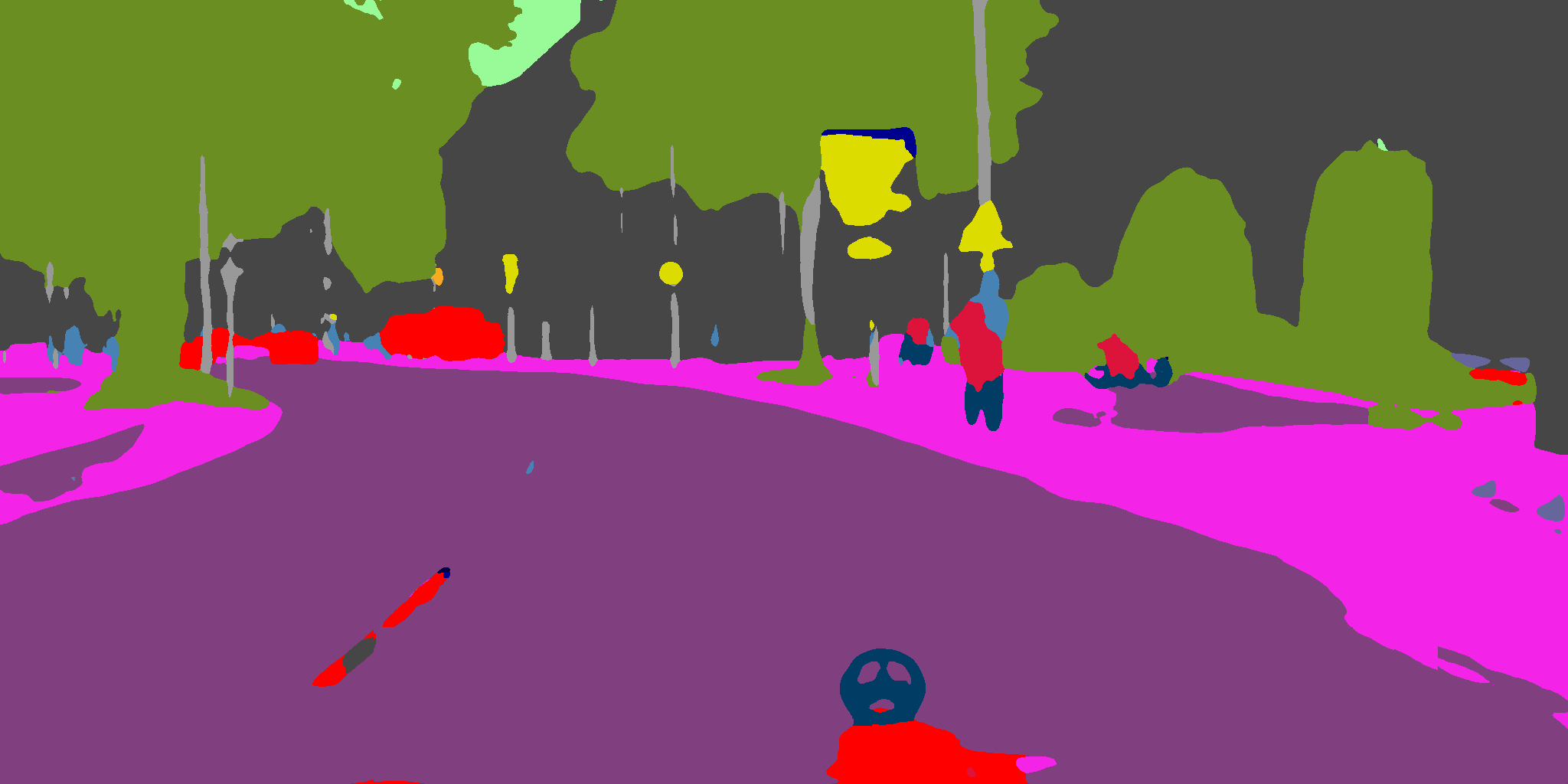}}
	\subfloat[Ours(stage 2)]{\includegraphics[width=0.125\linewidth]{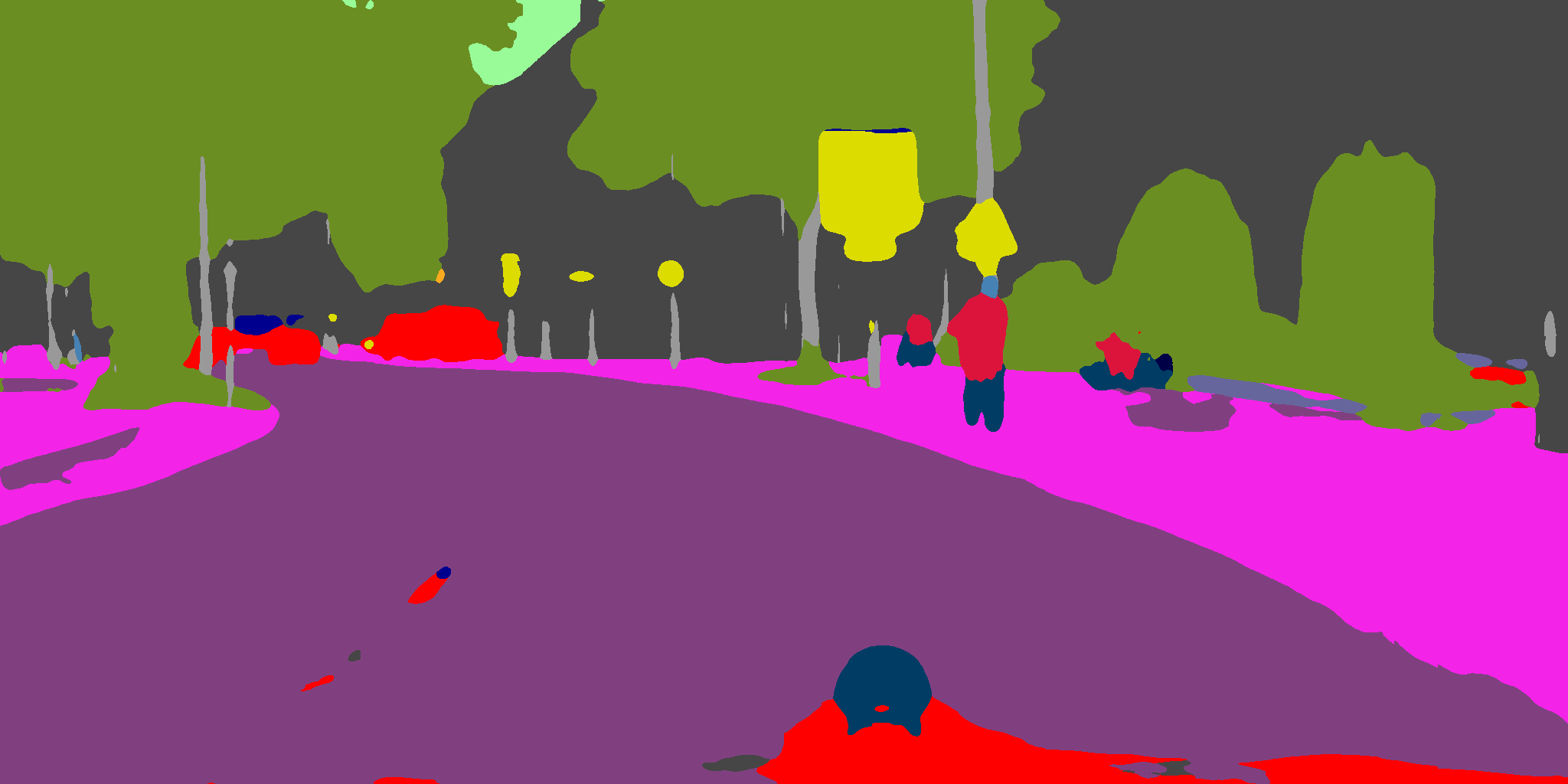}} \\
	\caption{Qualitative adaptation results of different entropy-based UDA methods using BiSeNet on SYNTHIA-to-Cityscapes.
	From left to right are images/ground truth, the predictions of baseline, shannon entropy loss, maximum square loss, neutral cross-entropy loss, and our two-stage UDA method.
	The second to seventh columns of the first and third rows are the entropy maps of the prediction results of corresponding methods.}
	\label{fig:pics_synthia}
\end{figure*}

Our two-stage UDA method using BiSeNet reaches 52.2\% mIoU of 16 classes and 59.1\% mIoU* of 13 classes in SYNTHIA-to-Cityscapes,
which are 11.4\% and 12.6\% higher than the baseline 
and comparable with the performance of some self-training methods\cite{rectifying,iast} using DeeplabV2.
Threshold-adaptative unsupervised focal loss helps the model achieve 49.0\% mIoU and 55.7\% mIoU*, 
outperforming previously advanced neutral cross-entropy loss with 4.1\% and 4.2\%, and all other entropy-based UDA methods.
It is supervised that maximum square loss degrades the performance in our experiments, 
implying that it relies on the image-wise weighting and multi-level self-guided approach \cite{maximum}.
The model has steady improvement on some difficult classes, 
such as sidewalk, wall, pole, traffic light, traffic sign, bus, and bike, 
indicating our method help to optimize the ``hard-to-transfer'' samples.
When using the DeepLabV2, our method reaches 58.4\% mIoU and 65.4\% mIoU*,
which are state-of-the-art and exceed previous adversarial-based\cite{leanringoutput,bidirectional} and self-training\cite{rectifying,iast,prototypical} methods with healthy margins.

The qualitative adaptation results of different entropy-based UDA methods using BiSeNet are shown in Fig.~\ref{fig:pics_synthia}, 
which is consistent with the quantitative results in Table~\ref{tab:synthia_comparison}. 
Our two-stage UDA method helps the model make more explicit predictions in the target domain 
and performs better on categories like sidewalk and traffic sign.
Meanwhile, the entropy maps produced by our method are much clearer, 
implying that many hard-to-transfer samples got optimized.

\subsubsection{GTA5-to-Cityscapes}
The overall performance of our UDA method on the GTA5-to-Cityscapes is shown in Table~\ref{tab:gta5_comparison}.

Our two-stage UDA method achieves 55.4\% mIoU of 19 classes and is 8.9\% higher than the baseline. 
It brings steady improvement for low-IoU classes like wall, traffic light, traffic sign, truck, and bike.
Meanwhile, the proposed threshold-adaptative unsupervised focal loss brings 6.8\% mIoU improvement to the baseline, 
surpassing all previous entropy-based UDA methods. 
What's more, our method achieves state-of-the-art 59.6\% mIoU when adopting DeepLabV2.

\begin{figure*}[htbp]
	\centering
	\captionsetup[subfloat]{font=scriptsize,labelfont=scriptsize}
	\subfloat{\includegraphics[width=0.125\linewidth]{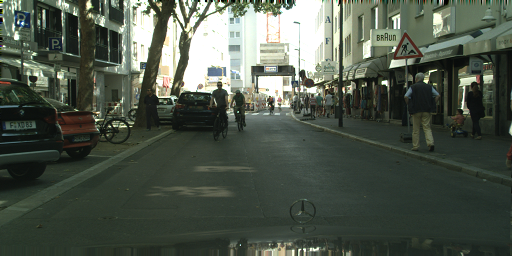}}
	\subfloat{\includegraphics[width=0.125\linewidth]{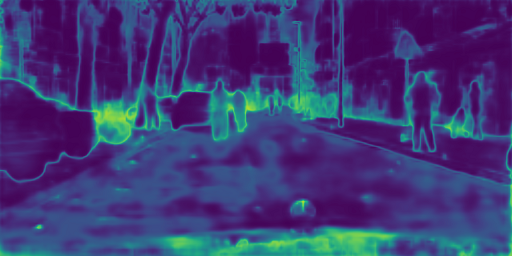}}
	\subfloat{\includegraphics[width=0.125\linewidth]{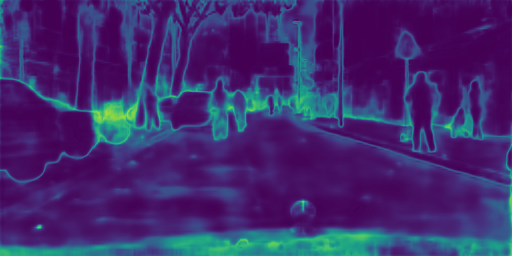}}
	\subfloat{\includegraphics[width=0.125\linewidth]{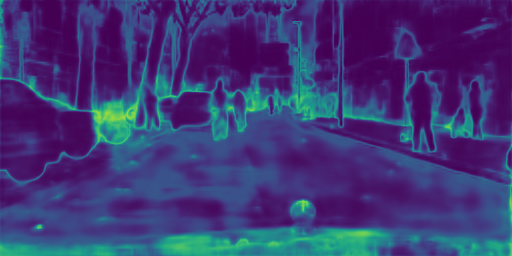}}
	\subfloat{\includegraphics[width=0.125\linewidth]{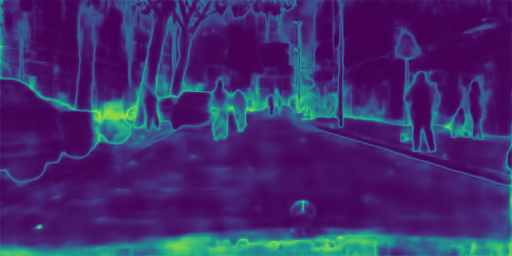}}
	\subfloat{\includegraphics[width=0.125\linewidth]{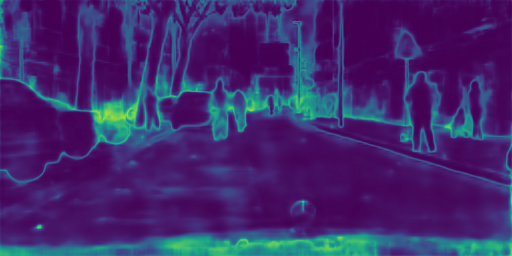}}
	\subfloat{\includegraphics[width=0.125\linewidth]{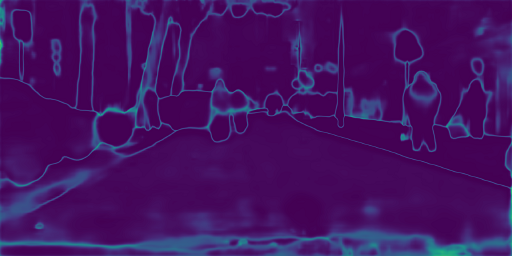}}
	\subfloat{\includegraphics[width=0.125\linewidth]{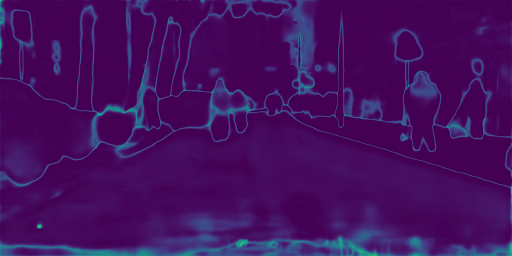}} \\ \vspace{-0.30cm}
	\subfloat{\includegraphics[width=0.125\linewidth]{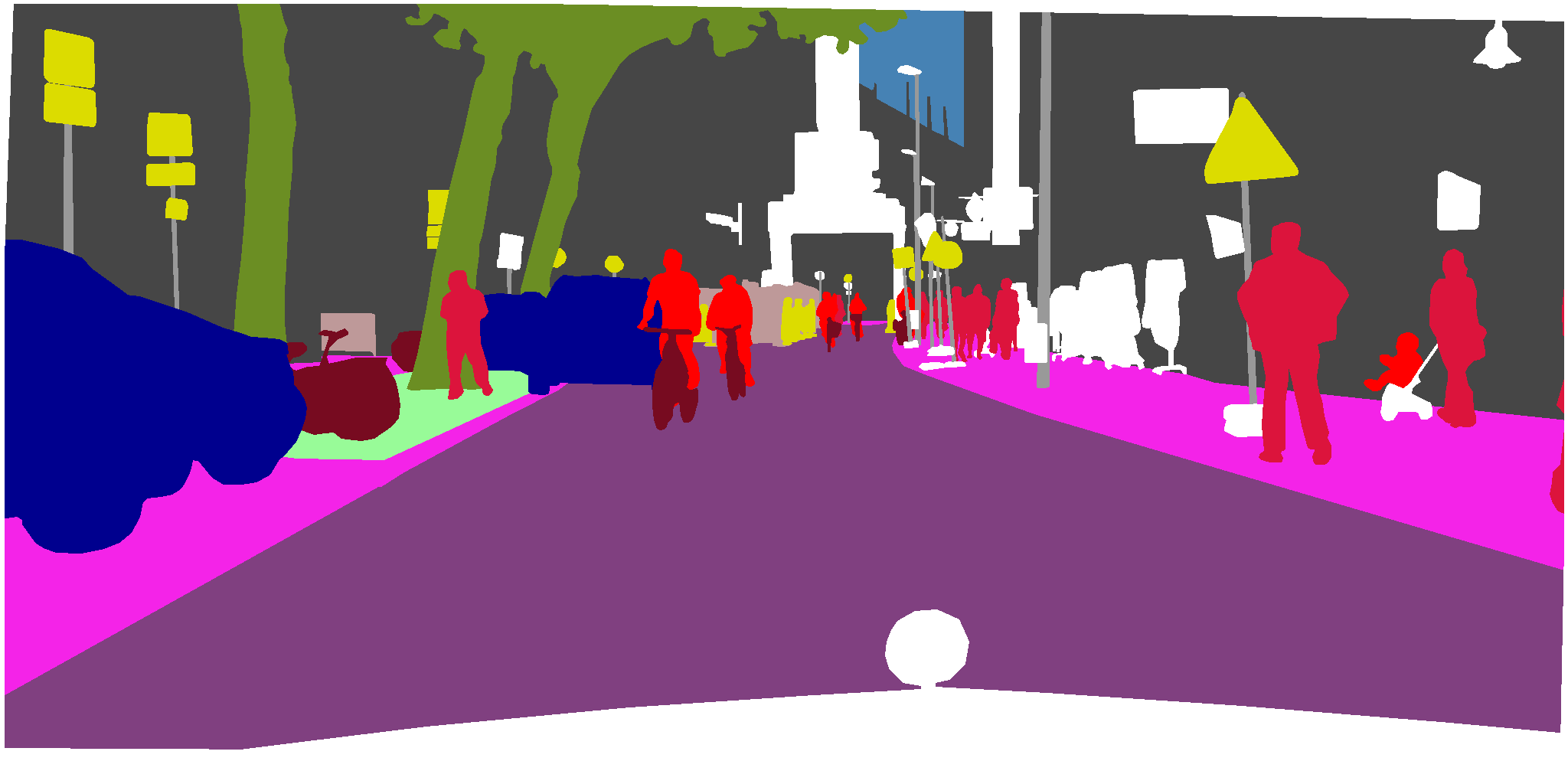}}
	\subfloat{\includegraphics[width=0.125\linewidth]{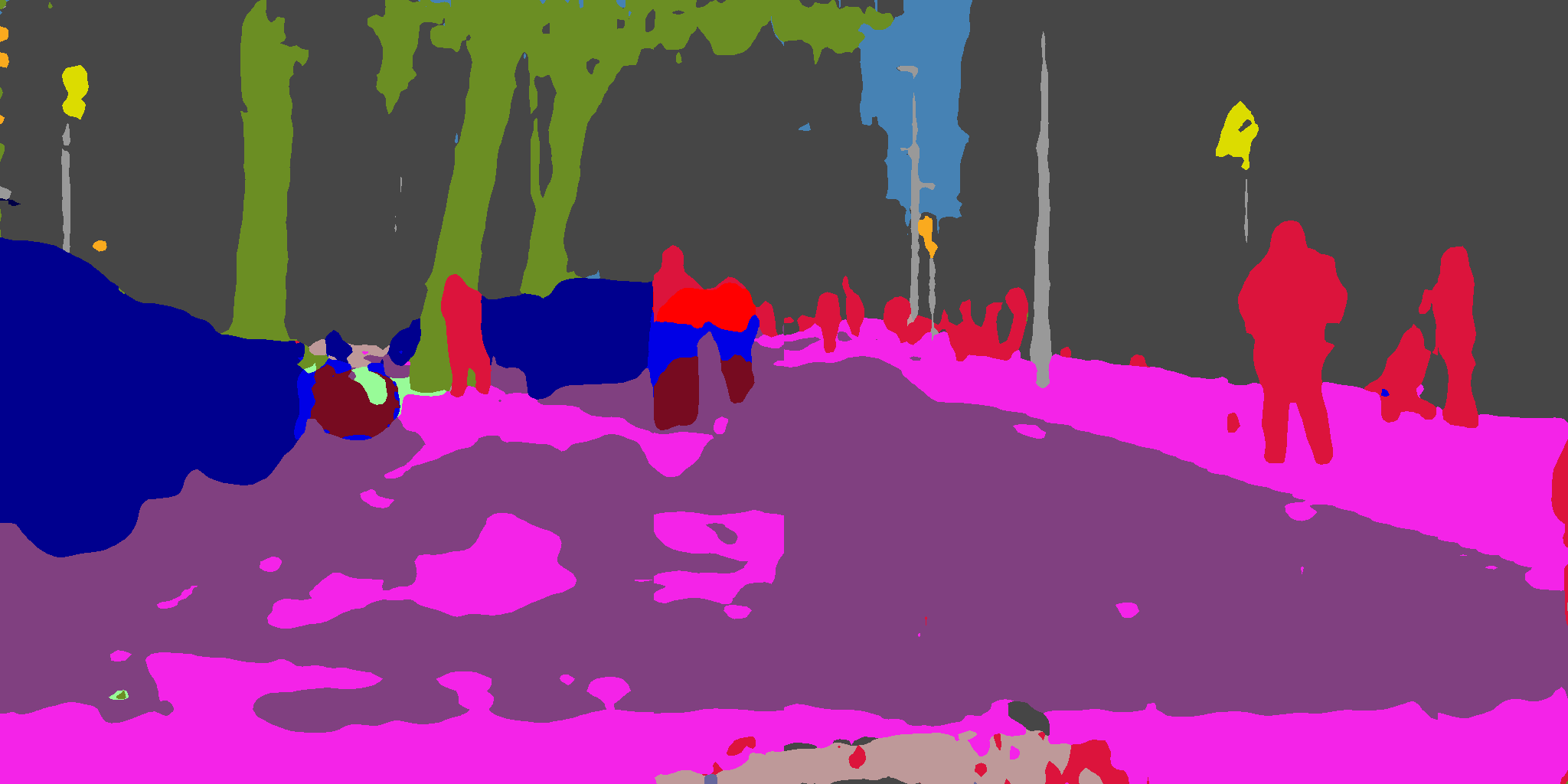}}
	\subfloat{\includegraphics[width=0.125\linewidth]{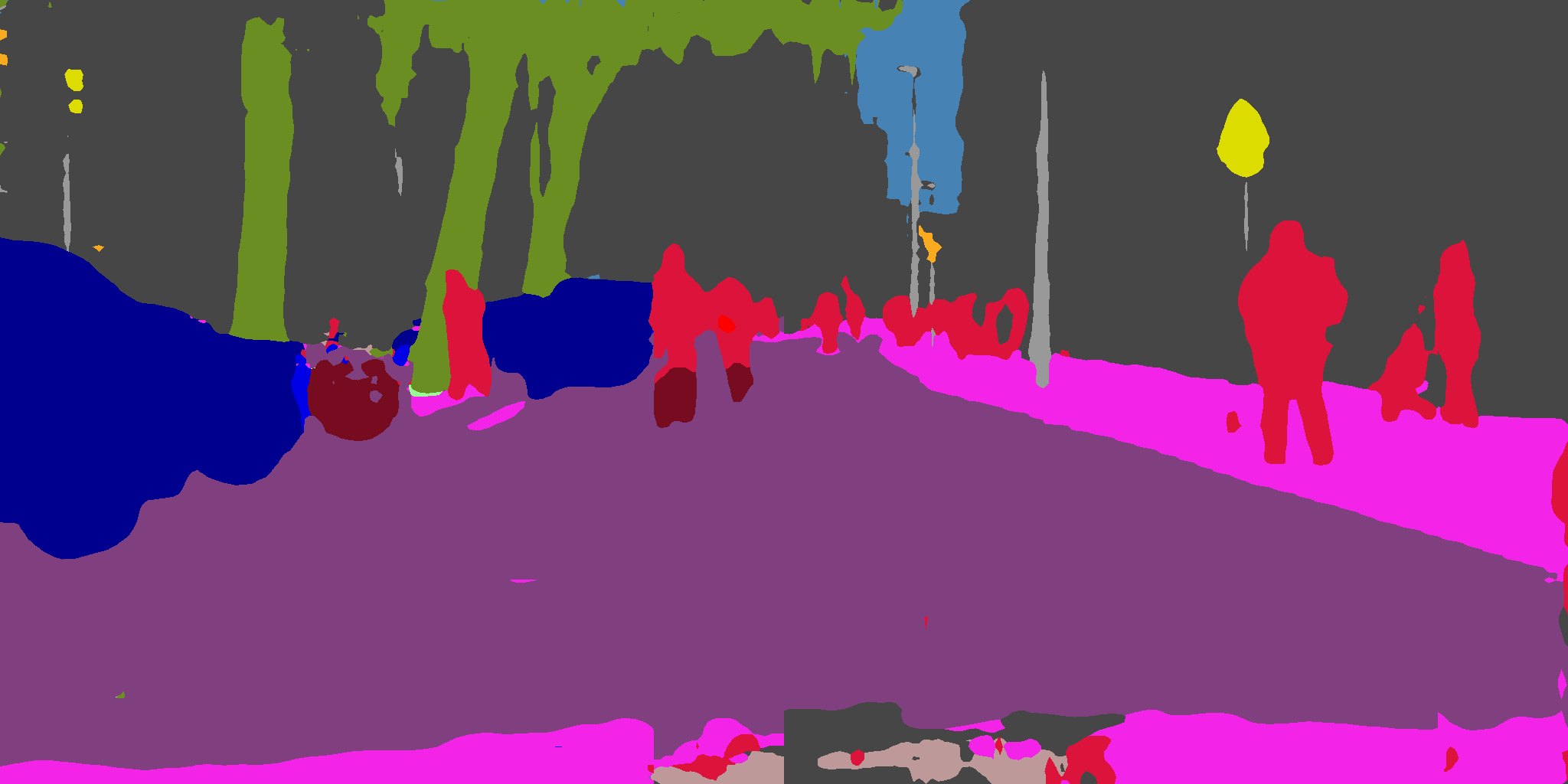}}
	\subfloat{\includegraphics[width=0.125\linewidth]{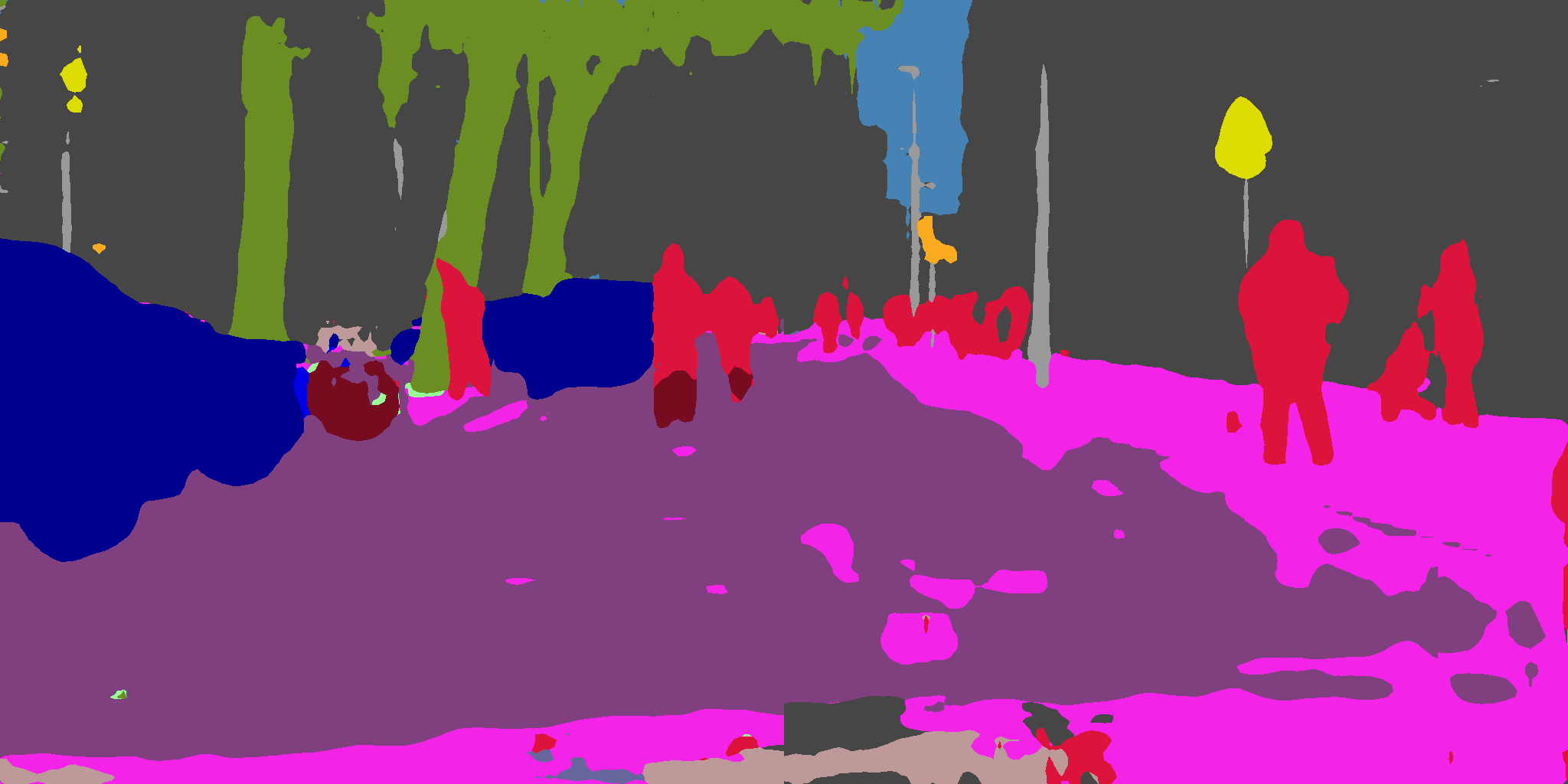}}
	\subfloat{\includegraphics[width=0.125\linewidth]{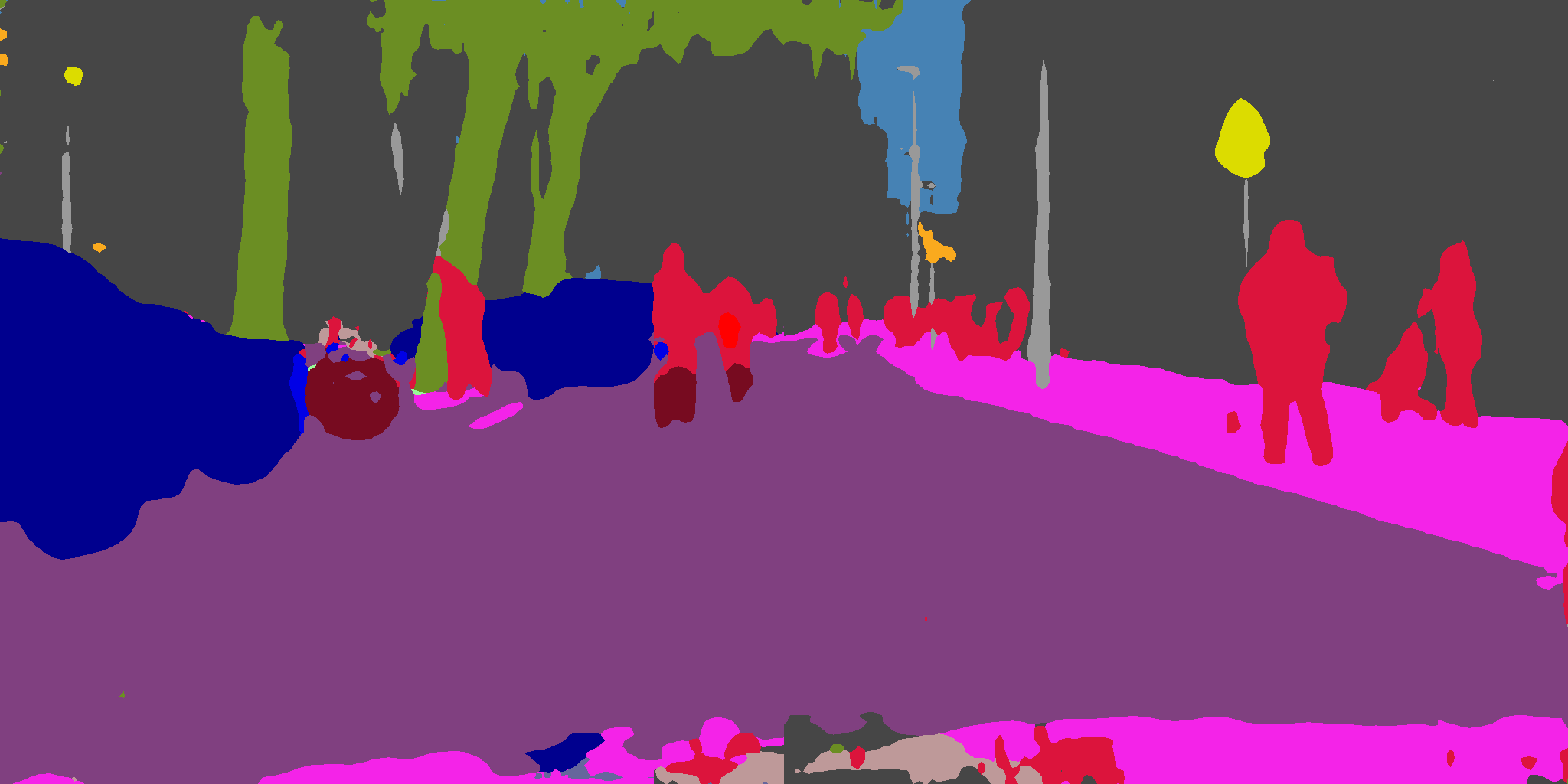}}
	\subfloat{\includegraphics[width=0.125\linewidth]{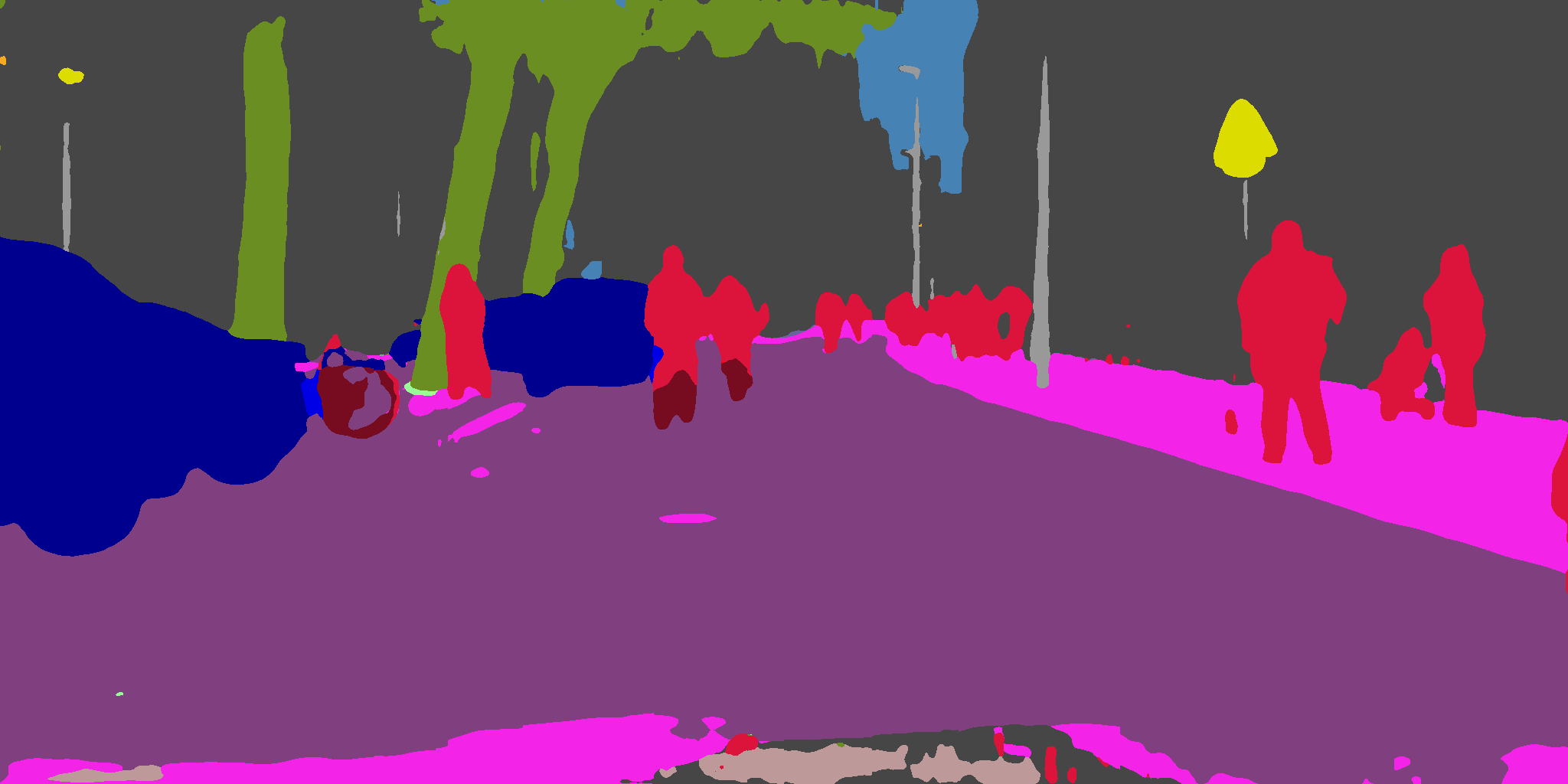}}
	\subfloat{\includegraphics[width=0.125\linewidth]{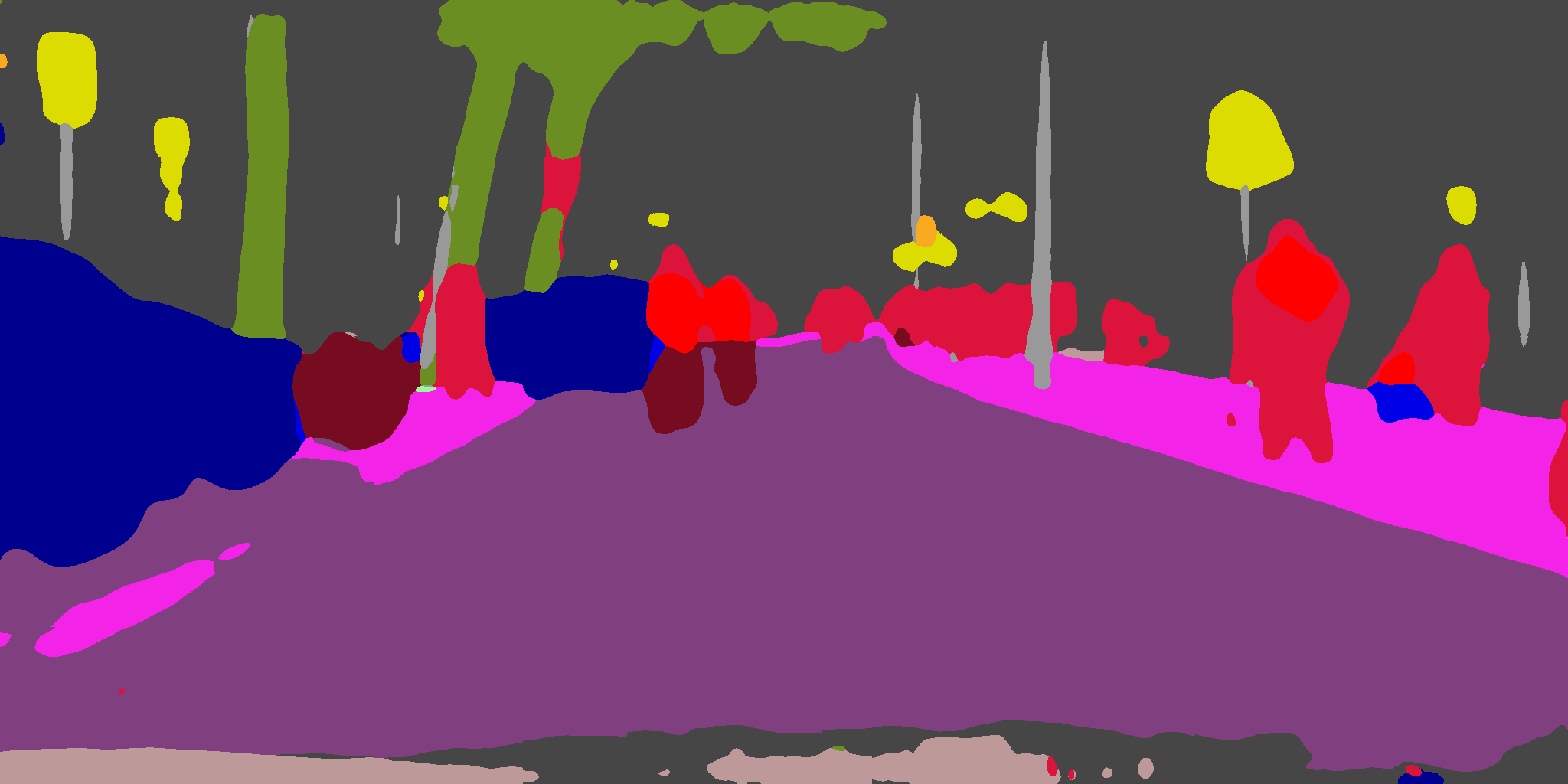}}
	\subfloat{\includegraphics[width=0.125\linewidth]{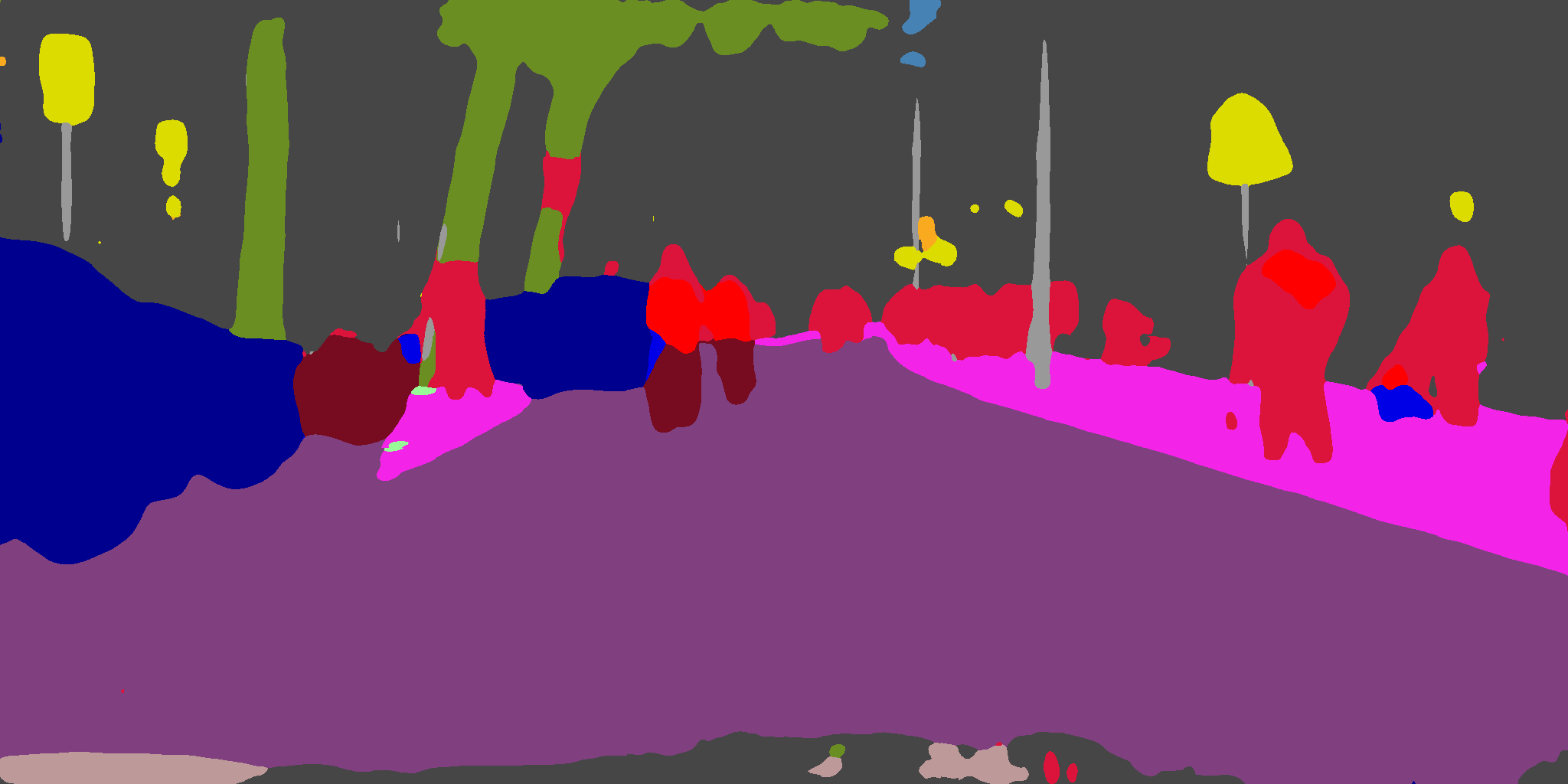}} \\ \vspace{-0.30cm}
	\subfloat{\includegraphics[width=0.125\linewidth]{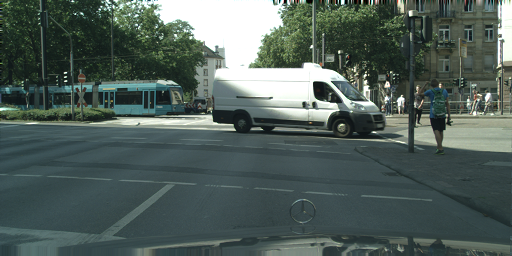}}
	\subfloat{\includegraphics[width=0.125\linewidth]{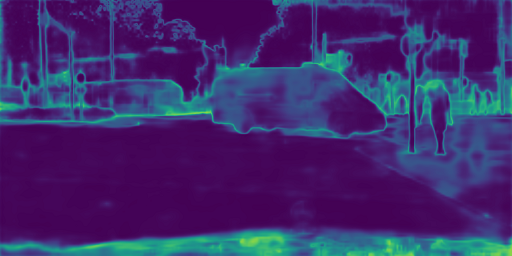}}
	\subfloat{\includegraphics[width=0.125\linewidth]{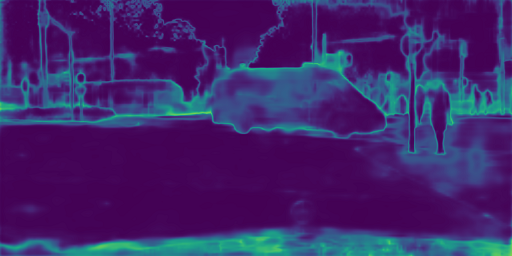}}
	\subfloat{\includegraphics[width=0.125\linewidth]{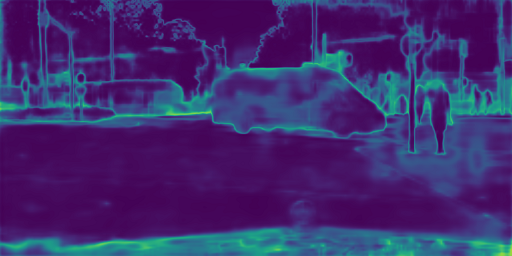}}
	\subfloat{\includegraphics[width=0.125\linewidth]{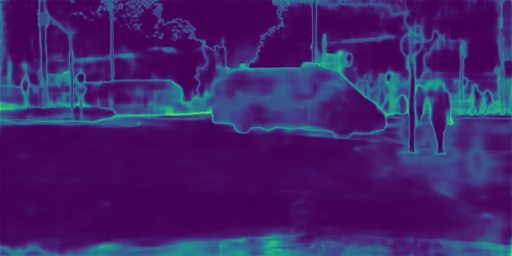}}
	\subfloat{\includegraphics[width=0.125\linewidth]{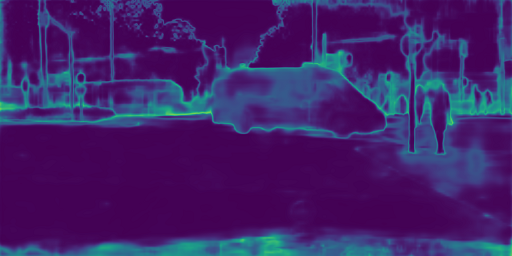}}
	\subfloat{\includegraphics[width=0.125\linewidth]{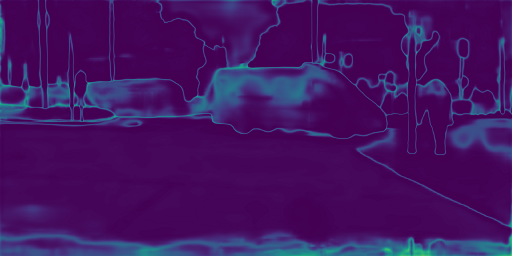}}
	\subfloat{\includegraphics[width=0.125\linewidth]{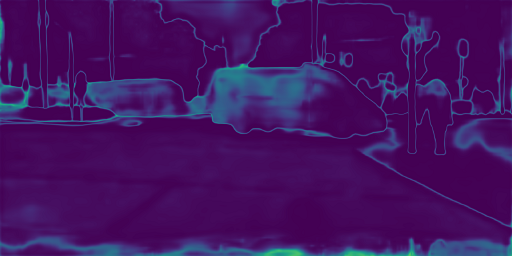}} \\ \vspace{-0.30cm}
	\subfloat[Image/Ground truth]{\includegraphics[width=0.125\linewidth]{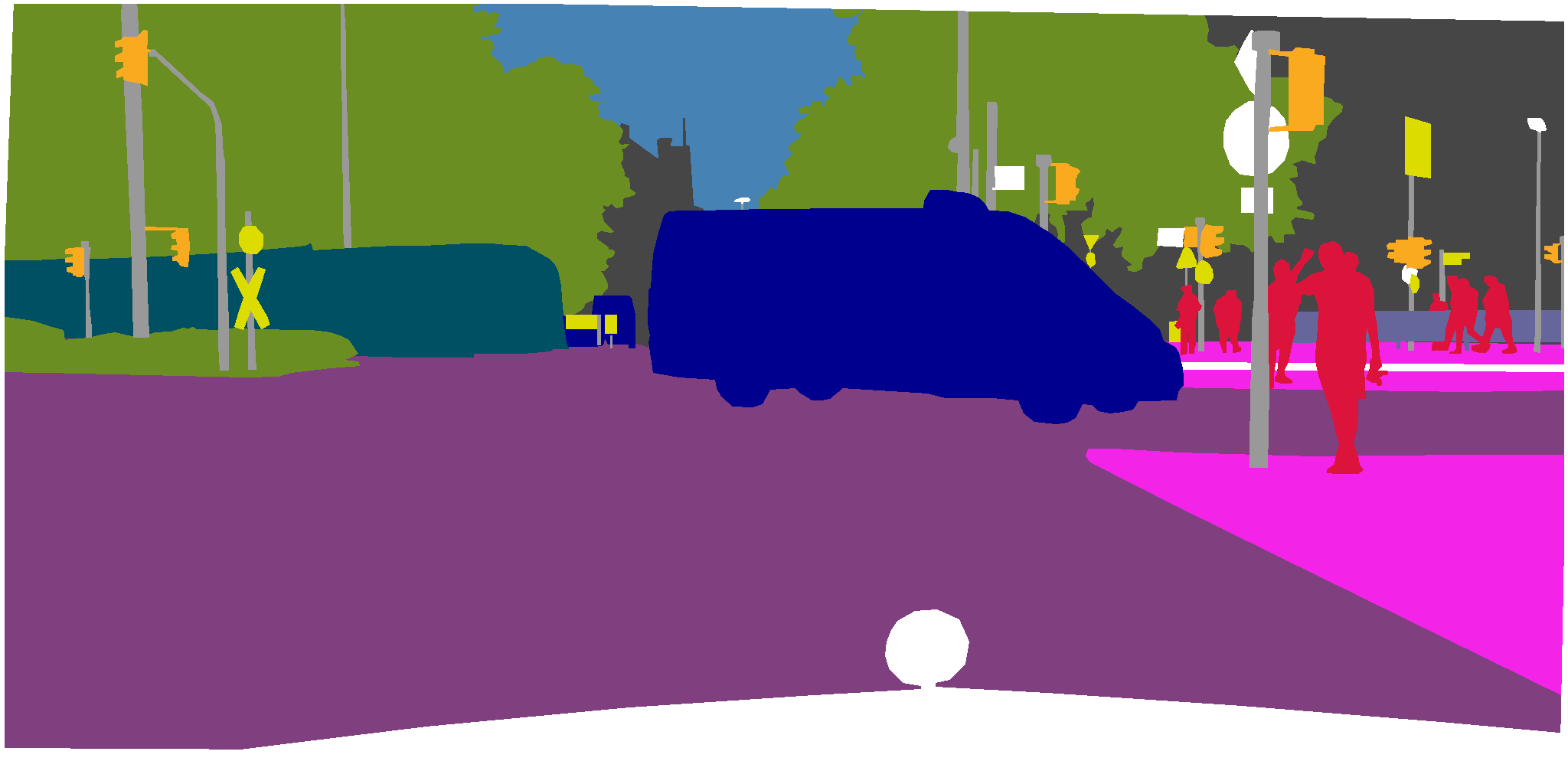}}
	\subfloat[Baseline(IST)]{\includegraphics[width=0.125\linewidth]{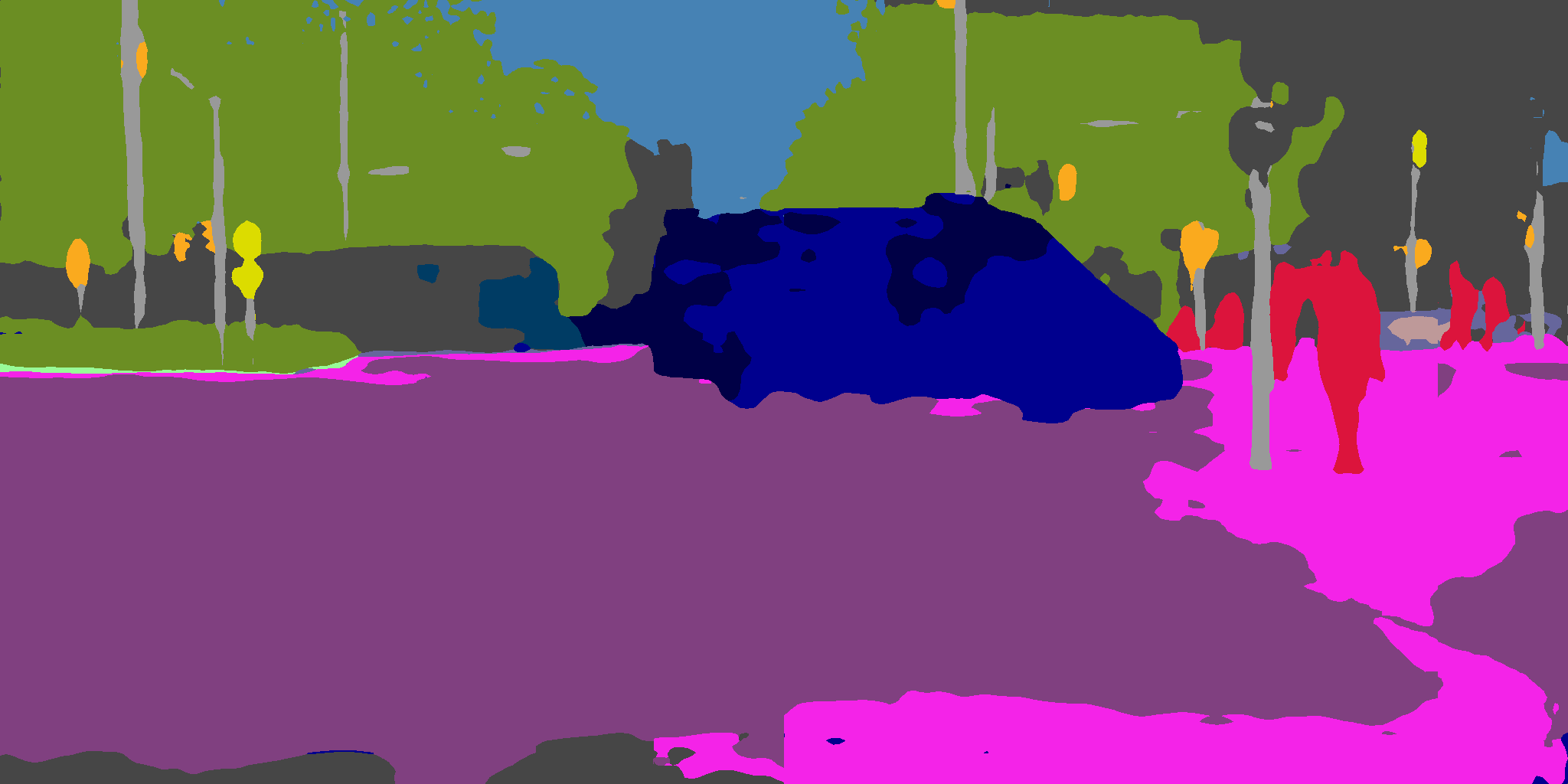}}
	\subfloat[Entropy]{\includegraphics[width=0.125\linewidth]{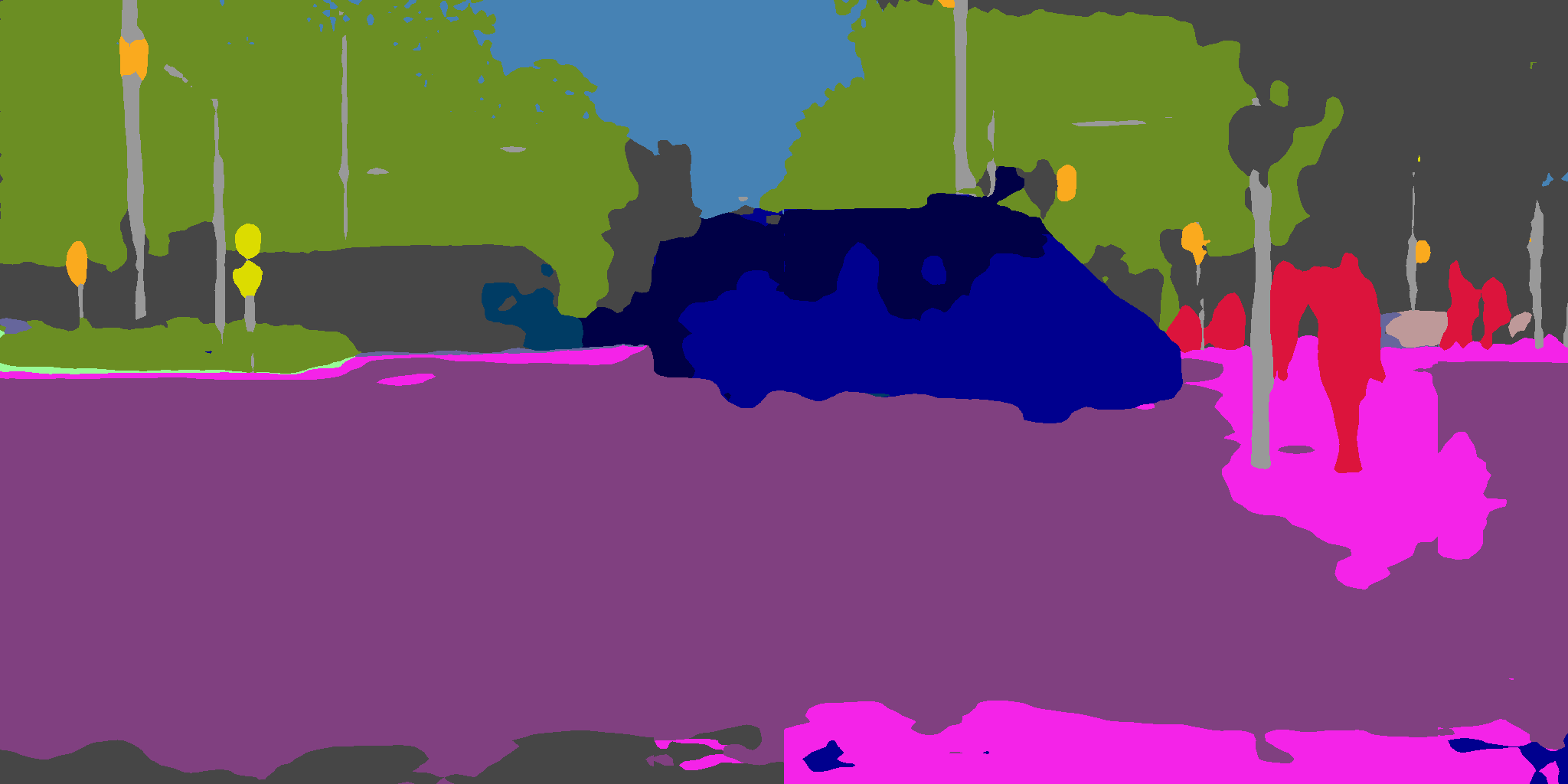}}
	\subfloat[Maximum]{\includegraphics[width=0.125\linewidth]{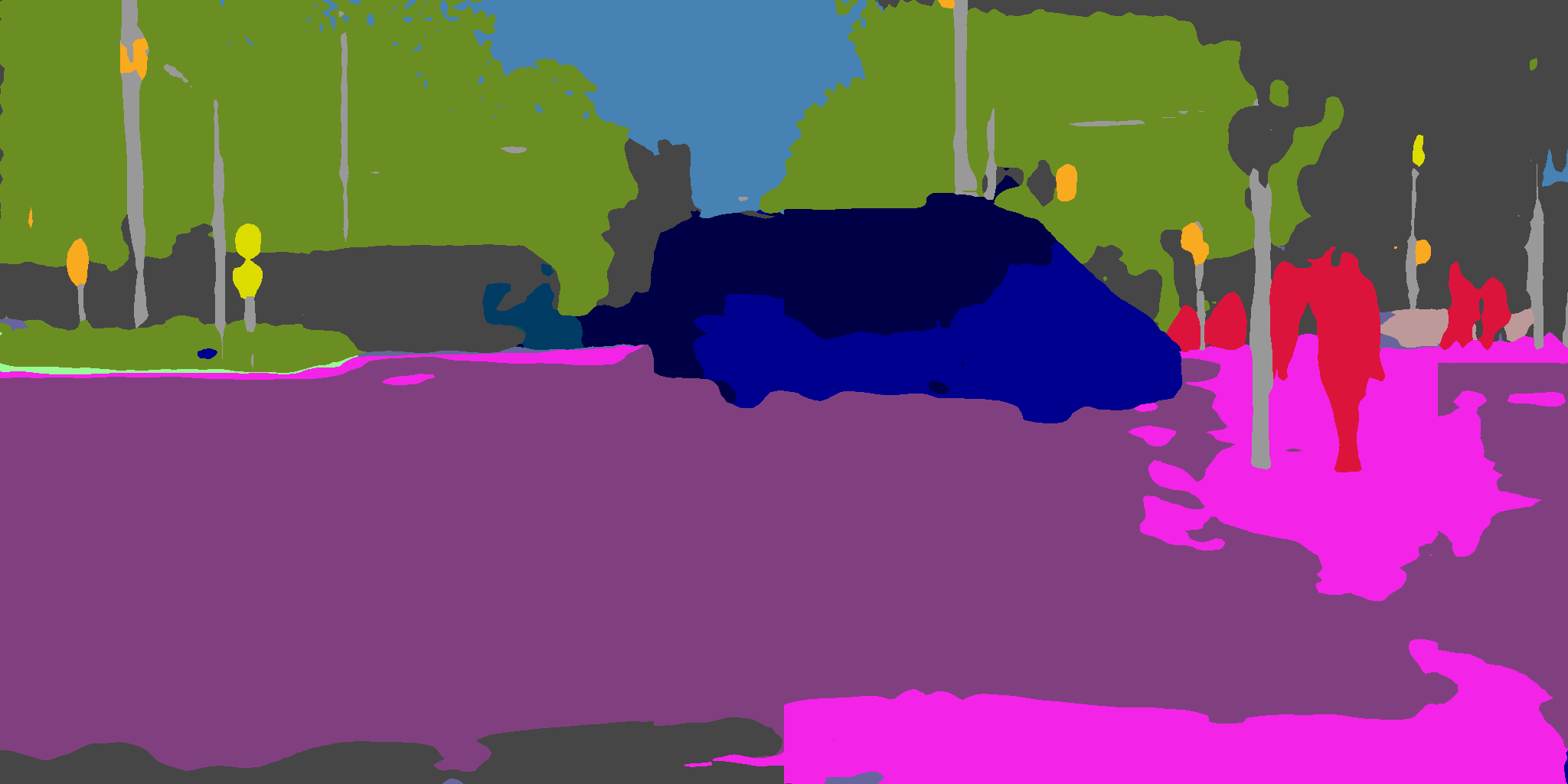}}
	\subfloat[Confidence]{\includegraphics[width=0.125\linewidth]{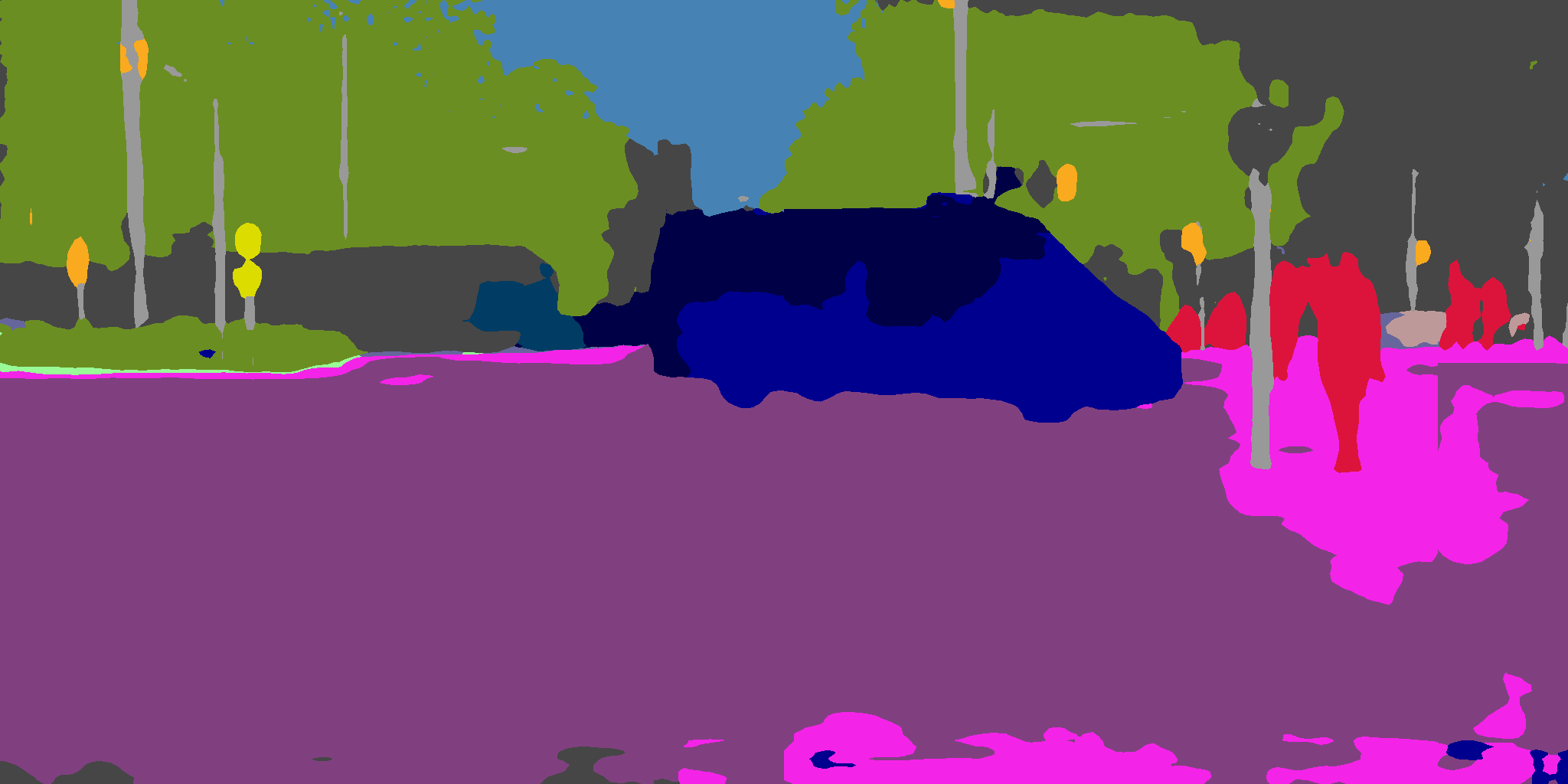}}
	\subfloat[Neutral]{\includegraphics[width=0.125\linewidth]{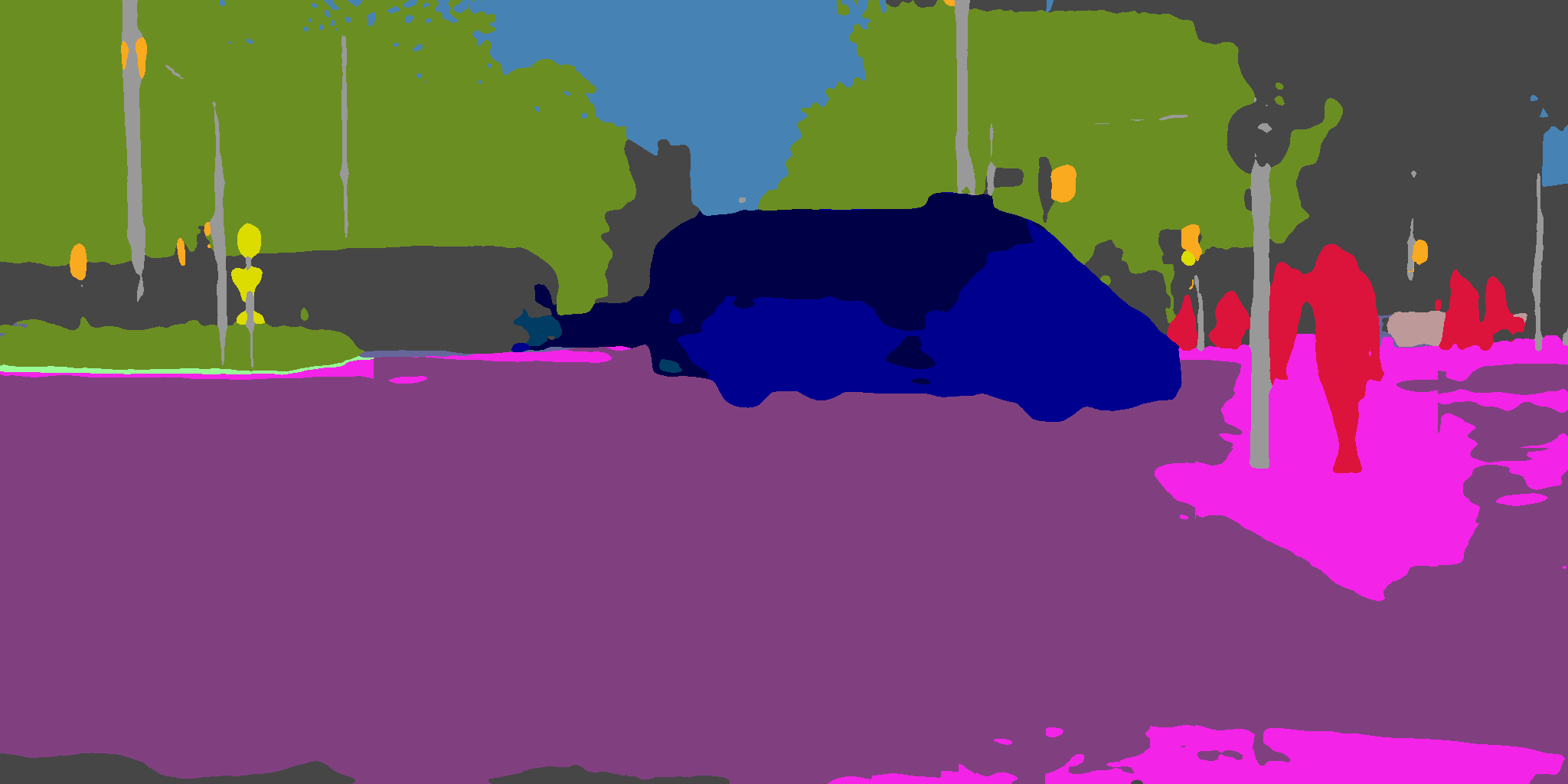}}
	\subfloat[Ours(stage 1)]{\includegraphics[width=0.125\linewidth]{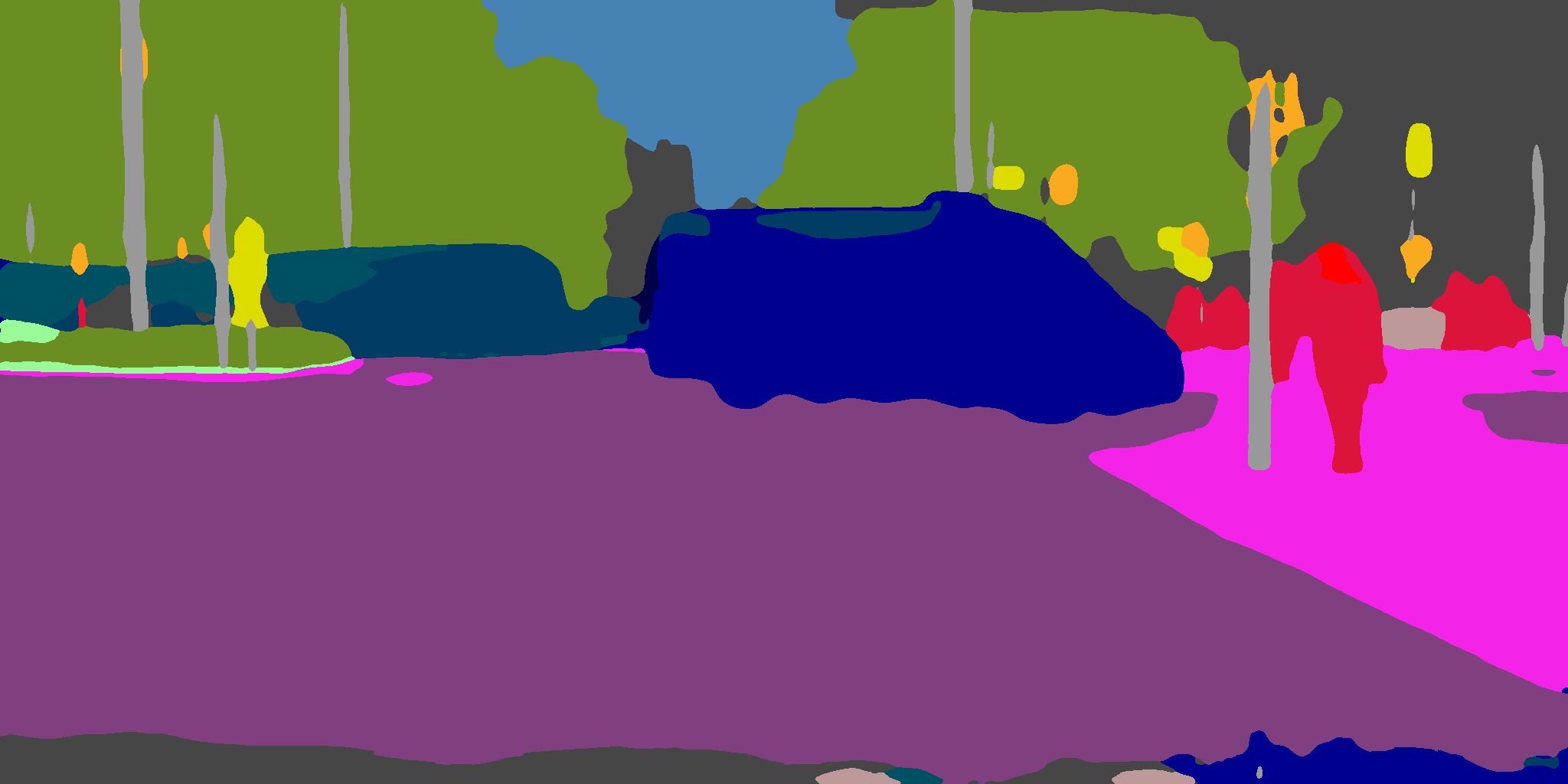}}
	\subfloat[Ours(stage 2)]{\includegraphics[width=0.125\linewidth]{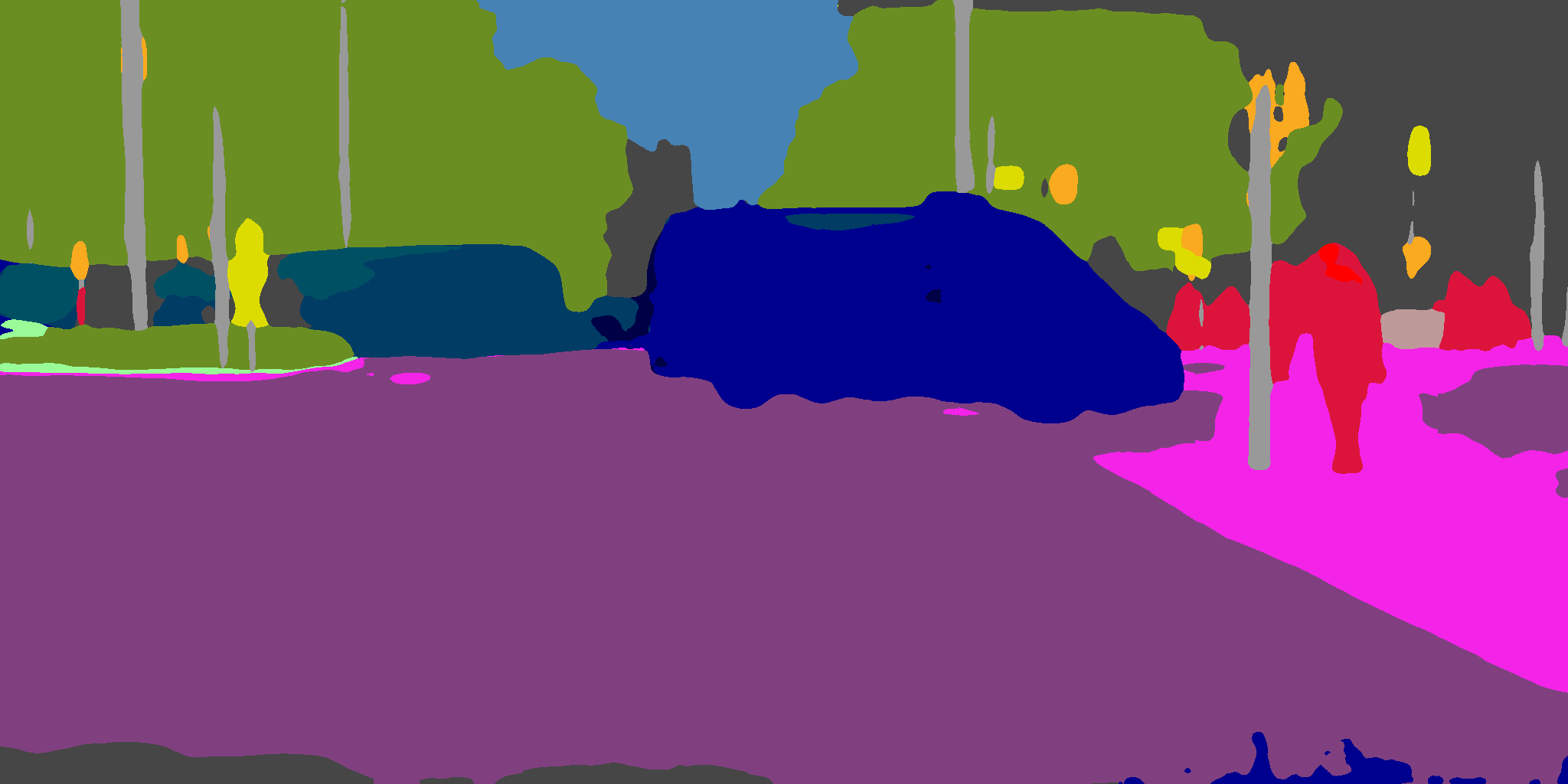}}\\
	\caption{Qualitative adaptation results of different entropy-based UDA methods using BiSeNet on GTA5-to-Cityscapes.
	From left to right are images/ground truth, the predictions of baseline, shannon entropy loss, maximum square loss, neutral cross-entropy loss, and our two-stage UDA method.
	The second to seventh columns of the first and third rows are the entropy maps of the prediction results of corresponding methods.}
	\label{fig:pics_gta5}
\end{figure*}

Fig. \ref{fig:pics_gta5} shows the qualitative adaptation results, 
where our method performs better than other entropy-based UDA methods and demonstrates its effectiveness.

\begin{table*}[ht]
	\caption{The adaptation performance and comparison on GTA5-to-Cityscapes}
	\label{tab:gta5_comparison}
	\resizebox{\linewidth}{!}{
	\begin{tabular}{c|c|ccccccccccccccccccc|c}
		\hline
		Methods & Network & road & sw & build & wall & fence & pole & light & sign & vege & terrain & sky & person & rider & car & truck & bus & train & motor & bike & mIoU \\ \hline
		AdaptSeg\cite{leanringoutput} & \multirow{13}{*}{\begin{tabular}{c} DeeplabV2\\(ResNet101) \end{tabular}} & 86.5 & 36.0 & 79.9 & 23.4 & 23.3 & 23.9 & 35.2 & 14.8 & 83.4 & 33.3 & 75.6 & 58.5 & 27.6 & 73.7 & 32.5 & 35.4 & 3.9 & 30.1 & 28.1 & 42.4\\
		Shannon\cite{advent} &   & 89.4 & 33.1 & 81.0 &26.6 & 26.8 & 27.2 & 33.5 & 24.7 & 83.9 & 36.7 & 78.8 & 58.7 & 30.5 & 84.8 & 38.5 & 44.5 & 1.7 & 31.6 & 32.4 & 45.5 \\ 
		Maximum\cite{maximum} &  & 89.4 & 43.0 & 82.1 & 30.5 & 21.3 & 30.3 & 34.7 & 24.0 & 85.3 & 39.4 & 78.2 & 63.0 & 22.9 & 84.6 & 36.4 & 43.0 & 5.5 & 34.7 & 33.5 & 46.4 \\ 
		BDL \cite{bidirectional} &  & 91.0 & 44.7 & 84.2 & 34.6 & 27.6 & 30.2 & 36.0 & 36.0 & 85.0 & 43.6 & 83.0 & 58.6 & 31.6 & 83.3 & 35.3 & 49.7 & 3.3 & 28.8 & 35.6 & 48.5 \\ 
		Confidence\cite{confidence} & & 90.7 & 45.9 & 84.5 & 34.7 & 29.2 & 31.9 & 37.6 & 33.1 & 84.4 & 42.6 & 85.2 & 58.1 & 32.5 & 83.0 & 34.7 & 50.1 & 4.4 & 29.5 & 30.7 & 48.6 \\
		Rectifying\cite{rectifying} & &  90.4 & 31.2 & 85.1 & 36.9 & 25.6 & 37.5 & 48.8 & 48.5 & 85.3 & 34.8 & 81.1 & 64.4 & 36.8 & 86.3 & 34.9 & 52.2 & 1.7 & 29.0 & 44.6 & 50.3 \\ 
		FDA\cite{fda} & & 92.5 & 53.3 & 82.4 & 26.5 & 27.6 & 36.4 & 40.6 & 38.9 & 82.3 & 39.8 & 78.0 & 62.6 & 34.4 & 84.9 & 34.1 & 53.1 & 16.9 & 27.7 & 46.4 & 50.5 \\ 
		DACS\cite{dacs} &  & 89.9 & 39.7 & \bfseries{87.9} & 30.7 & {39.5} & 38.5 & 46.4 & 52.8 & {88.0} & {44.0} & 88.8 & 67.2 & 35.8 & 84.5 & 45.7 & 50.2 & 0.0 & 27.3 & 34.0 & 52.1 \\ 
		IAST\cite{iast} &  & {94.1} & {58.8} & 85.4 & 39.7  & 29.2 & 25.1 & 43.1 & 34.2 & 84.8 & 34.6 & 88.7 & 62.7 & 30.3 & 87.6 & 42.3 & 50.3 & \bfseries{24.7} & {35.2} & 40.2 & 52.2 \\  
		SAC\cite{sac} & & 90.4 & 53.9& 86.6 & 42.4 & 27.3 & {45.1} & 48.5 & 42.7 & 87.4 & 40.1 & 86.1 & {67.5} & 29.7 & {88.5} & 49.1 & {54.6} & 9.8 & 26.6 & 45.3 & 53.8 \\ 
		CorDA\cite{corda} & & \bfseries{94.7} & \bfseries{63.1} & 87.6 & 30.7 & 40.6 & 40.2 & 47.8 & 51.6 & 87.6 & 47.0 & 89.7 & 66.7 & 35.9 & \bfseries{90.2} & 48.9 & 57.5 &  0 & 39.8 & 56.0 & 56.6 \\
		ProDA\cite{prototypical} & & 87.8 & 56.0 & 79.7 & {46.3} & \bfseries{44.8} & {45.6} & {53.5} & 53.5 & {88.6} & {45.2} & 82.1 & {70.7} & {39.2} & {88.8} & 45.5 & \bfseries{59.4} & 1.0 & {48.9} & {56.4} & {57.5} \\ 
		Ours & & 92.0 & {59.1} & 84.6 & \bfseries{48.0} & 40.3 & \bfseries{48.0} & \bfseries{55.1} & \bfseries{61.7} & \bfseries{89.1} & \bfseries{51.2} & 84.0 & \bfseries{72.6} & \bfseries{43.4} & 87.5 & {50.9} & 51.2 & 6.4 & \bfseries{50.9} & 55.6 & \bfseries{59.6} \\\hline
		Oracle & \multirow{8}{*}{\begin{tabular}{c} BiSeNet\\(ResNet18) \end{tabular} } & 95.6 & 79.4 & 87.6 & 53.5 & 50.7 & 52.3 & 57.9 & 67.0 & 90.4 & 58.0 & 91.7 & 71.5 & 47.8 & 91.1 & 59.5 & 72.7 & 57.5 & 45.6 & 66.7 & 68.2\\ \cline{1-1} \cline{3-22}
		Baseline (IST)&  & 88.8 & 49.0 & 84.8 & 32.7 & 23.7 & 36.3 & 42.7 & 40.6 & 83.4 & 31.7 & 84.9 & 59.6 & 24.4 & 84.5 & 33.8 & 39.1 & 4.1 & 13.2 & 26.4 & 46.5 \\ 
		Shannon \cite{advent} &  & 89.8 & 49.2 & 84.8 & 34.6 & 22.3 & 36.0 & 42.1 & 37.5 & 84.7 & 33.8 & 87.1 & 60.0 & 24.5 & 86.1 & 32.2 & 41.8 & 4.0 & 18.6 & 33.8 & 47.5 \\
		Maximum \cite{maximum} &  & 87.1 & 44.7 & 85.0 & 33.3 & 25.3 & 37.6 & 42.9 & 41.0 & 84.1 & 30.9 & 86.0 & 60.9 & 21.6 & 85.7 & 31.8 & 38.8 & 4.7 & 18.1 & 31.0 & 46.9 \\ 
		Confidence\cite{confidence} & & 90.9  & 48.0  & 84.9  & 33.7  & 25.1  & 38.3  & 43.6  & 40.6  & 85.4  & 36.9  & 83.9  & 60.1  & 24.2  & 85.1  & 27.6  & 32.8  & 5.1  & 17.2  & 41.4  & 47.6 \\
		Neutral\cite{neutral} & & 90.9 & 52.9 & 86.1 & 38.5 & 24.5 & 41.5 & 46.7 & 42.1 & 86.7 & 37.0 & 86.4 & 63.3 & 24.2 & 88.1 & {36.8} & 42.4 & 2.3 & 19.0 & 30.4 & 49.5 \\
		Ours(stage 1) & & 91.7 & 57.2 & 87.4 & 39.3 & 29.9 & 42.7 & {51.7} & 57.3 & 87.3 & 39.7 & 89.9 & 63.2 & 26.3 & {88.5} & 46.8 & 53.1 & 2.1 & 18.5 & 41.0 & 53.3 \\
		Ours(stage 2) & & 89.8 & 54.9 & 87.1 & {47.0} & 34.0 & 41.8 & {53.0} & {58.6} & 86.6 & 38.4 & \bfseries{92.3} & 57.7 & 28.3 & 87.3 & \bfseries{51.0} & 50.2 & 9.3 & 26.7 & \bfseries{58.3} & 55.4 \\ \hline
	\end{tabular}}
\end{table*}

\subsection{Ablation Study}
\subsubsection{The improvement of each method}
We gradually add unsupervised focal loss (focal), class-level dynamic threshold adjustment strategy (threshold), 
and cross-domain image mixing (CIM) to the baseline (IST) to explore their performance. 
The results on SYNTHIA-to-Cityscapes are shown in Table \ref{tab:each_method_synthia}.
The unsupervised focal loss brings 5.1\% mIoU improvement, 
the dynamic threshold adjustment strategy further boosts 3.1\% mIoU,
while the CIM improves the mIoU to 52.2\%, demonstrating their effectiveness.

\begin{table}[ht]
	\centering
	\caption{Improvement of each method on SYNTHIA-to-Cityscapes}
	\label{tab:each_method_synthia}
	\begin{tabular}{ccc|c|c}
		\hline
		focal & threshold & mixing & mIoU & mIoU$^{*}$ \\ \hline
		& & & 40.8 & 46.5 \\
		$\surd$ & & & 45.9 & 52.6 \\
		$\surd$ & $\surd$ & & 49.0 & 55.7 \\
		$\surd$ & $\surd$ & $\surd$ & \bfseries{52.2} & \bfseries{59.1} \\ \hline
	\end{tabular}
\end{table}

\subsubsection{The coefficient $\gamma$ of unsupervised focal loss}
The $\gamma$ in unsupervised focal loss regularizes the loss contribution of easy-to-transfer samples.
We fix the confidence threshold to 0.8 and set $\gamma$ to 1.0, 2.0, and 3.0.
Table \ref{tab:gamma_focal} presents the experimental results, 
where $\gamma=2$ performs best and is selected in our experiments.

\begin{table}[ht]
	\centering
	\caption{The performance of different $\gamma$ on SYNTHIA-to-Cityscapes}
	\label{tab:gamma_focal}
	\begin{tabular}{cccc}
		\hline
		$\gamma$ & 1.0  & 2.0 & 3.0 \\ \hline
		mIoU  & 45.5  & \bfseries{45.9} & 44.5 \\
		mIoU$^{*}$ & 52.2  & \bfseries{52.6} & 51.1\\ \hline
	\end{tabular} 
\end{table}

\subsubsection{The parameters of class-level dynamic threshold adjustment strategy}

We set $a=0.9$ and gradually tune the $b$ and $d$, 
which is shown in Table \ref{tab:parameters_threshold}.
The first row where $a=1.0$ denotes the unsupervised focal loss with a fixed confidence threshold of 0.8.
When increasing the $d$ from 4 to 10, 
the mIoU first increases then decreases, 
which indicates that we need to regularize the thresholds of the ``hard" classes to some extent, 
but not excessively.
Meanwhile, bigger $b$ (0.9) may bring more noise, 
while lower $b$ (0.5) makes fewer pixels be included in loss calculation,
and the middle one (0.8) is finally adopted.
Nevertheless, the class-level dynamic threshold adjustment strategy can improve the adaptation performance within the appropriate parameter range, 
only needing fine-tuning to the best.
Finally, $a$, $b$, and $d$ are set as 0.9, 0.8, and 8.
 
\begin{table}[ht]
	\centering
	\caption{The performance of different parameters in dynamic threshold adjustment strategy on SYNTHIA-to-Cityscapes}
	\label{tab:parameters_threshold}
	\begin{tabular}{ccc|c|c}
		\hline
		$a$ & $b$ & $d$ & mIoU & mIoU$^{*}$ \\ \hline
		1.0 & - & - & 45.9 & 52.6 \\
		0.9 & 0.8 & 4 & 47.5 & 54.0 \\
		0.9 & 0.8 & 8 & \bfseries{48.4} & \bfseries{55.7} \\
		0.9 & 0.8 & 10 & 46.9 & 53.4 \\
		0.9 & 0.9 & 8 & 46.8 & 53.2 \\
		0.9 & 0.5 & 8 & 45.2 & 51.7 \\
		\hline
	\end{tabular}
\end{table}

\subsubsection{Fixed threshold vs. dynamic threshold}
We set the confidence threshold to 0.2, 0.4, 0.6, 0.8, 
and the results are presented in Table \ref{tab:solid_threshold}.
It can be observed that 
a fixed threshold is hard to balance noise suppression (high threshold) and incorporating more pixels into the loss calculation (low threshold), 
making it hard for entropy-based UDA methods to achieve advanced performance.
Our unsupervised focal loss performs better at all fixed thresholds  
and further increases the mIoU$^{*}$ to 55.7 with the dynamic threshold adjustment strategy.

\begin{table}[ht]
	\centering
	\caption{The performance of fixed threshold and dynamic threshold on SYNTHIA-to-Cityscapes}
	\label{tab:solid_threshold}
	\begin{tabular}{c|ccccc}
	\hline
	\multirow{2}{*}{Method} & \multicolumn{5}{c}{mIoU$^{*}$} \\ \cline{2-6} 
			& 0.2 & 0.4 & 0.6 & 0.8 & dynamic \\ \hline
	Shannon	& 50.2 & 50.6 & 49.4 & 49.9 & - \\ 
	Maximum & 41.8 & 43.5 & 42.7 & 45.0 & - \\
	Confidence & 47.8 & 49.4 & 50.5 & 50.9 & - \\
	Neutral	& 51.5 & 50.5 & 51.6 & 51.5 & - \\
	Focal   & \bfseries{51.9} & \bfseries{51.7} & \bfseries{52.8} & \bfseries{52.6} & \bfseries{55.7} \\\hline
	\end{tabular}
\end{table}

\subsubsection{The coefficients of loss function}
\begin{figure}
	\centering
	\includegraphics[width=0.95\linewidth]{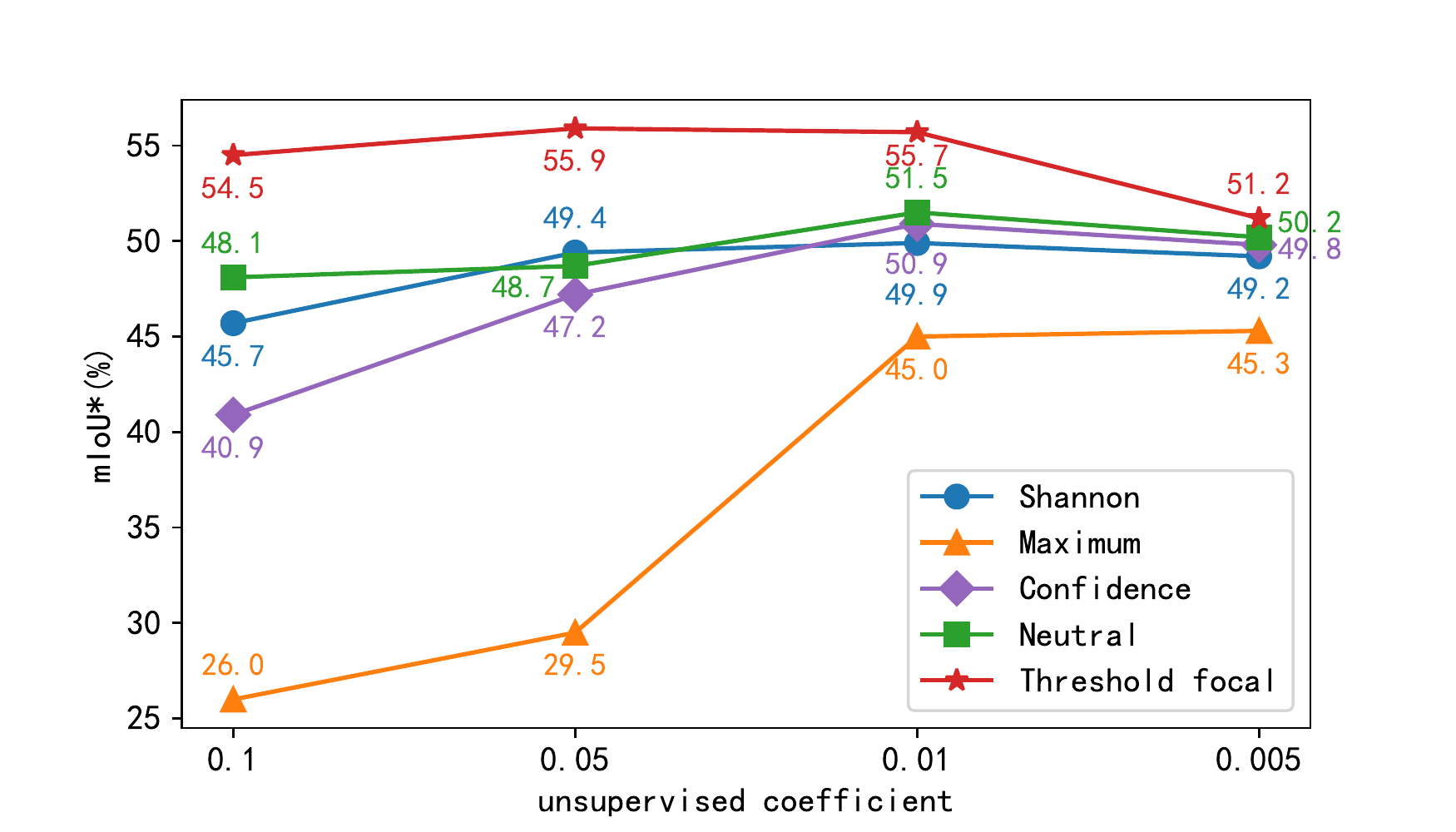}
	\caption{The performance of different $\lambda_{u}$ on SYNTHIA-to-Cityscapes.}
	\label{fig:coefficient_u}
\end{figure}

\begin{figure}
	\centering
	\includegraphics[width=0.92\linewidth]{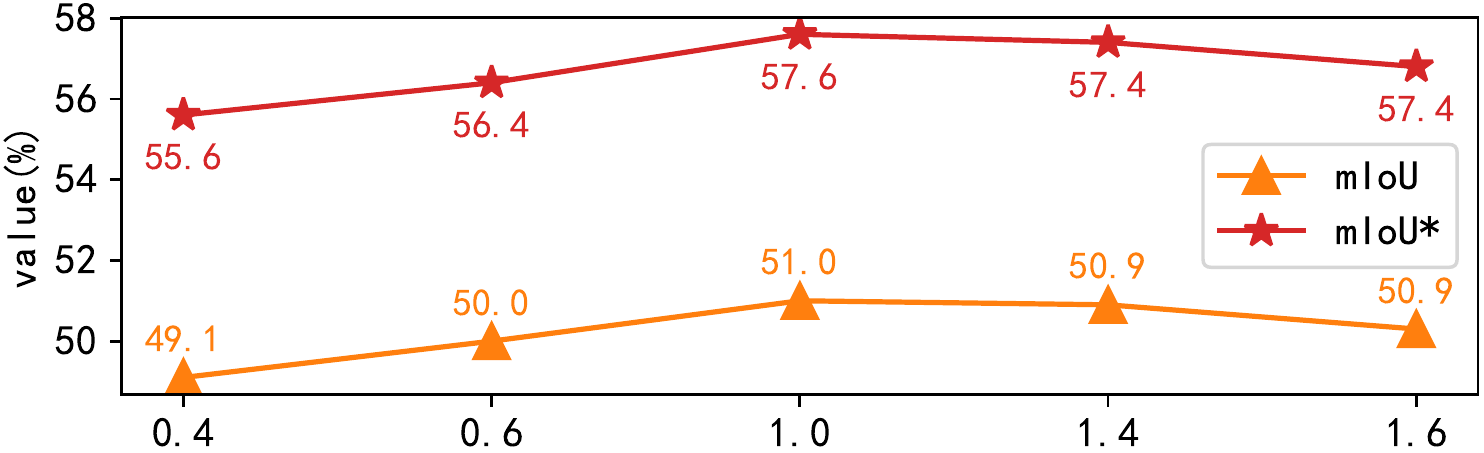}
	\caption{The performance of different $\lambda_{m}$ on SYNTHIA-to-Cityscapes.}
	\label{fig:coefficient_m}
\end{figure}

In stage one, we set the coefficient $\lambda_u$ of unsupervised loss $L_u$ to 0.1, 0.05, 0.01, 0.005, 
and the results are shown in Fig.~\ref{fig:coefficient_u}.
Shannon entropy loss, maximum square loss, neutral cross-entropy loss, and our threshold-adaptative unsupervised focal loss 
peak at 0.01, 0.005, 0.01, and 0.05, respectively.
It indicates that entropy-based losses essentially act as the regularization term. 
In stage two, we set the coefficient $\lambda_{m}$ of the mixed loss to 0.4, 0.6, 1.0, 1.4, and 1.6, 
and Fig.~\ref{fig:coefficient_m} presents the results.
With the increase of $\lambda_{m}$, the performance of the model first improves and then degrades, 
and finally, 1.0 is selected in our experiments.

\subsection{Comparison of UDA method details}
Table~\ref{tab:model_detail} exhibits some details of our UDA method 
and the state-of-the-art ProDA\cite{prototypical}.
It can be seen that DeeplabV2 adapts better than the BiSeNet, 
but will bring more computation, parameters and memory usage, 
and take more time for training and inference.
Due to the optimization of hard samples, 
our method converges faster than ProDA and achieves higher mIoUs.
Moreover, our two-stage method only needs about 7 hours for training 
on two 3090 GPUs and 28 ms for inference on a single one when using BiSeNet, 
dramatically reducing the training time and enabling real-time inference.

\begin{table}[!t]
	\centering 
	\caption{The details of our UDA method and ProDA}
	\label{tab:model_detail}
	\begin{tabular}{c|c|c|c}
		\hline
		 Method & \multicolumn{2}{c|}{Our} & ProDA\cite{prototypical} \\ \hline
		 Model & BiSeNet & DeepLabV2 & DeepLabV2 \\ \hline
		 MACs (G) & \bfseries{121} & 2172 & 2172 \\
		 Params (M) & \bfseries{13} & 65 & 65\\
		 Disk usage (MB) & \bfseries{54} & 261 & 261 \\ \hline
		 Training time (h) & \bfseries{7} & 87 & 104 \\
		 Inference time (ms) & \bfseries{28} &  94 & 94 \\ \hline
		 mIoU from GTA5 (\%) & 55.4 & \bfseries{59.6} & 57.5 \\ 
		 mIoU from SYNTHIA (\%) & 52.2 & \bfseries{58.4} & 55.5 \\ \hline
	\end{tabular}
\end{table}

\subsection{Discussion}
Experiments on two synthetic-to-real settings verify the effectiveness of our methods. 
Here we discuss the relationship of our methods with the self-training methods.

The shannon entropy loss can be seen as a soft-assignment version of the pseudo label cross-entropy\cite{advent},
which tends to make prediction probabilities at 0 and 1, making $p_t$ $(\hat{p}_t)$ be like pseudo labels in the target domain.
Then KL divergence measures the prediction differences $(\hat{p}_t, p_{t^*})$ between perturbed image pairs,
which is similar to making predictions of target samples to be like the pseudo labels, but in a soft manner.
Threshold-adaptative unsupervised focal loss adopts loss regularization and dynamic threshold adjustment strategy to optimize hard samples.
In stage two, pseudo labels are generated for CIM.
What we focus on is bridging the semantic knowledge between two domains 
while not using pseudo labels to re-train the model in the target domain.
Meanwhile, our method only contains two-stage of training, 
unlike self-training methods that usually require iterative training.
Moreover, self-training methods can be used to further adapt the model from stage two and will be our future work.


\section{Conclusion}
In this paper, we propose a two-stage UDA framework for domain adaptation of semantic segmentation.
In stage one, we design the threshold-adaptative unsupervised focal loss 
with a class-level dynamic threshold adjustment strategy, 
which helps optimize hard samples and outperforms all previous entropy-based methods.
In stage two, we introduce CIM with long-tail class pasting to bridge the semantic knowledge between two domains,
further boosting the adaptation performance.
Extensive experiments on two synthetic-to-real benchmarks demonstrate that our method achieves state-of-the-art.
In the future, we will explore the adaptation performance of our method on more semantic segmentation networks.


\bibliographystyle{IEEEtran}
\bibliography{IEEEabrv,reference.bib}

\begin{IEEEbiography}[{\includegraphics[width=1in,height=1.25in,clip,keepaspectratio]{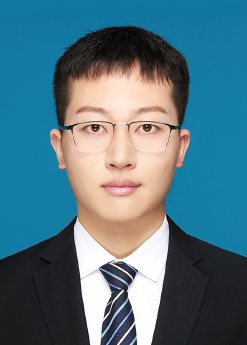}}]{\bfseries{Weihao Yan}}
received the B.S. degree in automation from Shanghai Jiao Tong University, Shanghai, China, in 2020. He is currently working toward the Ph.D. degree in Control Science and Engineering with Shanghai Jiao Tong University.\\
His main research interests include computer vision, image processing, deep learning and their applications in intelligent transportation systems.
\end{IEEEbiography}
\vspace{-10 mm}
	
\begin{IEEEbiography}[{\includegraphics[width=1in,height=1.25in,clip,keepaspectratio]{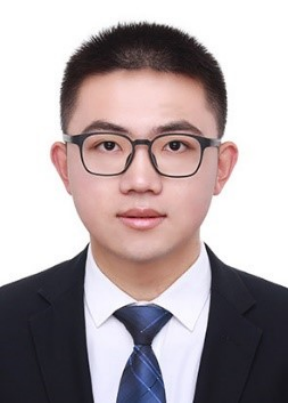}}]{\bfseries{Yeqiang Qian}}
received the Ph.D. degree in Control Science and Engineering from Shanghai Jiao Tong University, Shanghai, China, in 2020. He is currently a postdoctoral fellow with University of Michigan-Shanghai Jiao Tong University Joint Institute in Shanghai Jiao Tong University.\\ 
His main research interests include computer vision, pattern recognition, machine learning and their applications in intelligent transportation systems.
\end{IEEEbiography}
\vspace{-10 mm}

\begin{IEEEbiography}[{\includegraphics[width=1in,height=1.25in,clip,keepaspectratio]{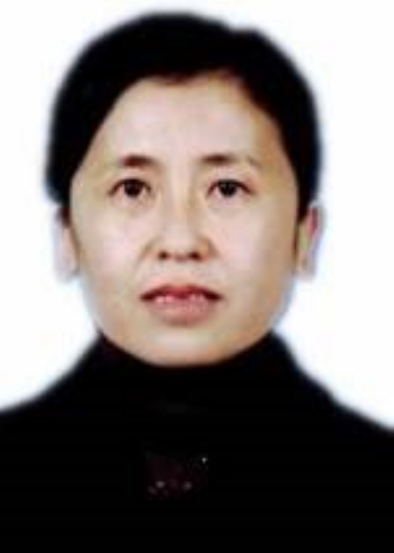}}]{\bfseries{Chunxiang Wang}}
received the Ph.D. degree in mechanical engineering from the Harbin Institute of Technology, Harbin, China, in 1999.\\
She is currently an Associate Professor with the Department of Automation at Shanghai Jiao Tong University, Shanghai, China. Her research interests include robotic technology and electromechanical integration.
\end{IEEEbiography}
\vspace{-10 mm}

\begin{IEEEbiography}[{\includegraphics[width=1in,height=1.25in,clip,keepaspectratio]{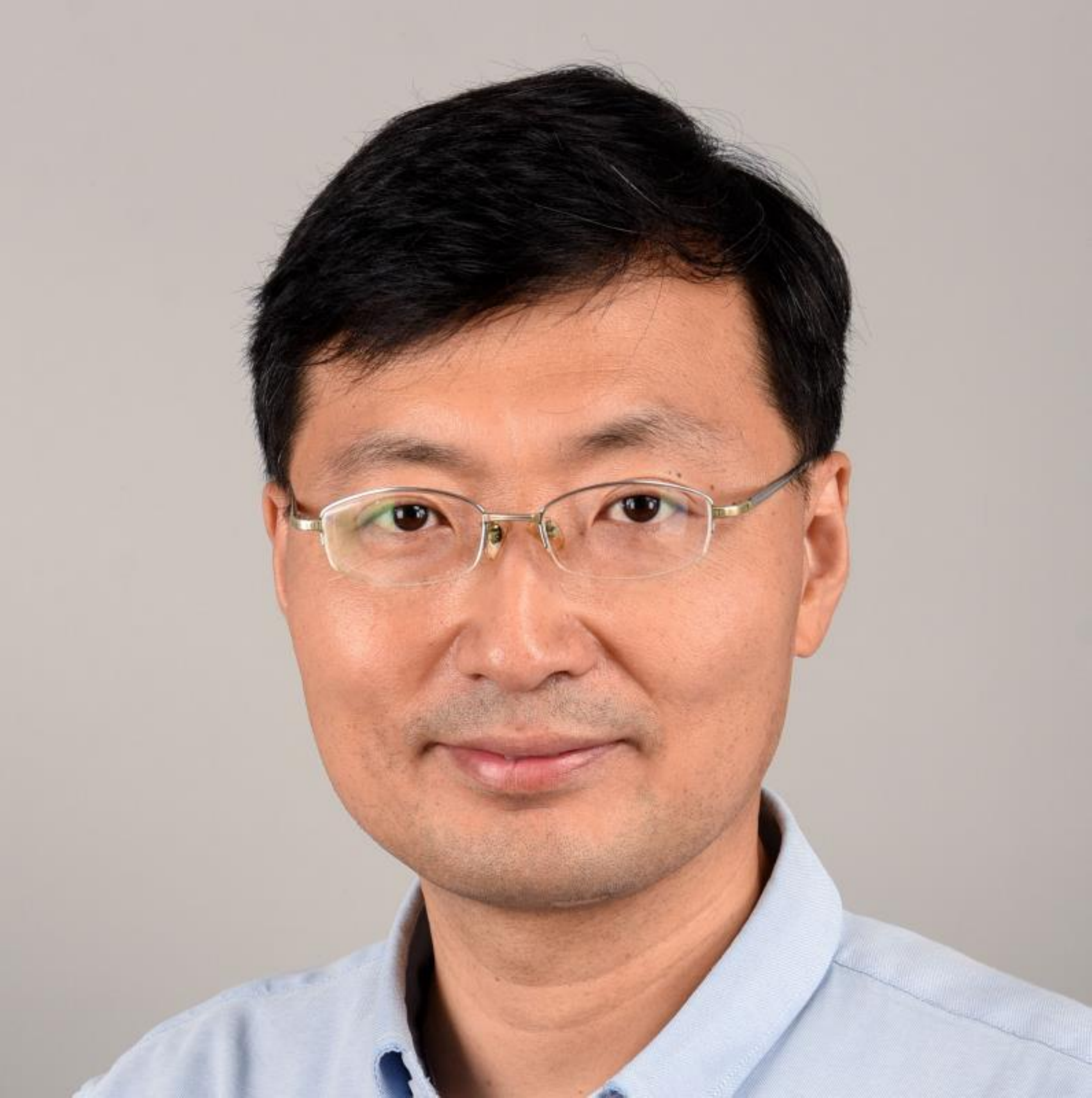}}]{\bfseries{Ming Yang}}
received the Master and Ph.D. degrees from Tsinghua University, Beijing, China, in 1999 and 2003, respectively.\\
He is currently the Full Tenure Professor at Shanghai Jiao Tong University, the deputy director of the Innovation Center of Intelligent Connected Vehicles. He has been working in the field of intelligent vehicles for more than 20 years. He participated in several related research projects, such as the THMR-V project (first intelligent vehicle in China), European CyberCars and CyberMove projects, CyberC3 project, CyberCars-2 project, ITER transfer cask project, AGV, etc.
\end{IEEEbiography}


\end{document}